\documentclass[acmtog]{acmart}
\acmSubmissionID{253}

\usepackage{booktabs}

\usepackage[ruled]{algorithm2e}

\SetAlFnt{\small}
\SetAlCapFnt{\small}
\SetAlCapNameFnt{\small}
\SetAlCapHSkip{0pt}

\newcommand{\ie}{\emph{i.e.}, }
\newcommand{\eg}{\emph{e.g.}, }

\newcommand{\datasetlink}{\href{https://syntec-research.github.io/Cafca}{https://syntec-research.github.io/Cafca}}

\makeatletter
\newcommand*{\addFileDependency}[1]{
  \typeout{(#1)}
  \@addtofilelist{#1}
  \IfFileExists{#1}{}{\typeout{No file #1.}}
}
\makeatother

\DeclareMathOperator*{\argmin}{arg\,min} 

\copyrightyear{2024} 
\acmYear{2024} 
\setcopyright{rightsretained} 
\acmConference[SA Conference Papers '24]{SIGGRAPH Asia 2024 Conference Papers}{December 3--6, 2024}{Tokyo, Japan}
\acmBooktitle{SIGGRAPH Asia 2024 Conference Papers (SA Conference Papers '24), December 3--6, 2024, Tokyo, Japan}\acmDOI{10.1145/3680528.3687580}
\acmISBN{979-8-4007-1131-2/24/12}

\begin{document}
\title{Cafca: High-quality Novel View Synthesis of Expressive Faces from Casual Few-shot Captures}
\subtitle{\datasetlink}
\author{Marcel C. Buehler}
\orcid{0000-0001-8104-9313}
\affiliation{
 \institution{ETH Zurich}
 \country{Switzerland}}
\author{Gengyan Li}
\orcid{0000-0002-1427-7612}
\affiliation{
 \institution{ETH Zurich and Google}
 \country{Switzerland}}
\author{Erroll Wood}
\orcid{0009-0006-2033-4704}
\affiliation{
 \institution{Google}
 \country{UK}}
\author{Leonhard Helminger}
\orcid{0000-0001-8528-2059}
\affiliation{
 \institution{Google}
 \country{Switzerland}}
\author{Xu Chen}
\orcid{0000-0003-1715-2820}
\affiliation{
 \institution{Google}
 \country{Switzerland}}
\author{Tanmay Shah}
\orcid{0009-0007-6492-8413}
\affiliation{
 \institution{Google}
 \country{USA}}
\author{Daoye Wang}
\orcid{0000-0002-2879-6114}
\affiliation{
 \institution{Google}
 \country{Switzerland}}
\author{Stephan Garbin}
\orcid{0009-0000-5005-8110}
\affiliation{
 \institution{Google}
 \country{UK}}
\author{Sergio Orts-Escolano}
\orcid{0000-0001-6817-6326}
\affiliation{
 \institution{Google}
 \country{Switzerland}}
\author{Otmar Hilliges}
\orcid{0000-0002-5068-3474}
\affiliation{
 \institution{ETH Zurich}
 \country{Switzerland}}
\author{Dmitry Lagun}
\orcid{0009-0002-5077-3469}
\affiliation{
 \institution{Google}
 \country{USA}}
\author{Jérémy Riviere}
\orcid{0000-0002-5249-5135}
\affiliation{
 \institution{Google}
 \country{Switzerland}}
\author{Paulo Gotardo}
\orcid{0000-0001-8217-5848}
\affiliation{
 \institution{Google}
 \country{Switzerland}}
\author{Thabo Beeler}
\orcid{0000-0002-8077-1205}
\affiliation{
 \institution{Google}
 \country{Switzerland}}
\author{Abhimitra Meka}
\orcid{0000-0001-7906-4004}
\affiliation{
 \institution{Google}
 \country{USA}}
\author{Kripasindhu Sarkar}
\orcid{0000-0002-0220-0853}
\affiliation{
 \institution{Google}
 \country{Switzerland}}
\renewcommand\shortauthors{Buehler, M. C. et al.}

 \begin{teaserfigure}
\centering  
\includegraphics[width=\linewidth]{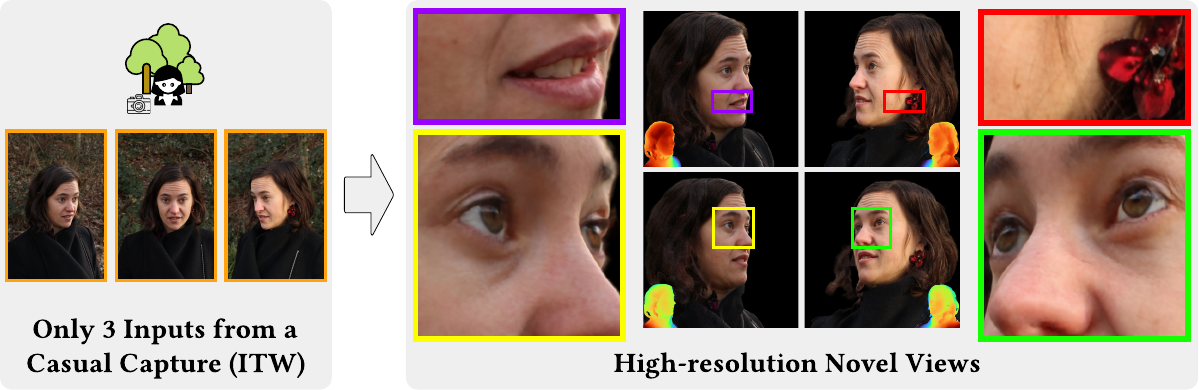} 
\caption{\label{fig:teaser}We propose a method for high-quality novel view synthesis of expressive faces captured in-the-wild. From a casual capture with only three images, our method synthesizes novel views at an unprecedented level of detail showing wrinkles, hair strands, skin pores, and eyelashes.}  
\end{teaserfigure}

\begin{abstract}
 Volumetric modeling and neural radiance field representations have revolutionized 3D face capture and photorealistic novel view synthesis. However, these methods often require hundreds of multi-view input images and are thus inapplicable to cases with less than a handful of inputs.
We present a novel volumetric prior on human faces that allows for high-fidelity expressive face modeling from as few as three input views captured in the wild. Our key insight is that an implicit prior trained on synthetic data alone can generalize to extremely challenging real-world identities and expressions and render novel views with fine idiosyncratic details like wrinkles and eyelashes.
We leverage a 3D Morphable Face Model to synthesize a large training set, rendering each identity with different expressions, hair, clothing, and other assets. We then train a conditional Neural Radiance Field prior on this synthetic dataset and, at inference time, fine-tune the model on a very sparse set of real images of a single subject. On average, the fine-tuning requires only three inputs to cross the synthetic-to-real domain gap. The resulting personalized 3D model reconstructs strong idiosyncratic facial expressions and outperforms the state-of-the-art in high-quality novel view synthesis of faces from sparse inputs in terms of perceptual and photo-metric quality.

\end{abstract}

%
%


%
%

\maketitle

\section{Introduction}
``Who sees the human face correctly: the photographer, the mirror, or the painter?” - Pablo Picasso. Our visual acuity is remarkably hypertuned to perceive the details of human faces due to evolutionary design~\cite{faceprocess}. Individual-specific expressions play a particularly significant role in perception tasks such as identification or estimation of emotion and intent~\cite{sinha06,eyeexpressions,lee2017reading}. Highly personalizable 3D representations that model the idiosyncracies of face shape and deformation at high quality are crucial to truly immersive and photorealistic 3D portrait photography. Perhaps, if trained well, an AI could join Picasso’s list. 

Wide-scale adoption and democratization of 3D portrait photography demands casual captures in uncontrolled environments -- \ie only a few shots taken with a handheld camera. 
Volumetric representations \cite{mildenhall2020nerf,mipnerf360,park2021nerfies,park2021hypernerf}, have demonstrated impressive quality and photorealism in synthesizing novel views of a large variety of 3D scenes using densely captured 2D images. Extended works relax the requirement for dense capture through various forms of regularization -- for instance, entropy minimization \citep{lolnerf}, spatial smoothness \cite{regnerf}, depth regularization \cite{diner,wang2023sparsenerf} and volume bounds \cite{sarkar2023litnerf}. These methods target general scenes and still require dozens of views captured simultaneously. As such, they cannot be applied to expressive 3D portrait photography in the wild, due to ambiguities that arise from the limited input and uncontrolled conditions.

Lifting very sparse 2D views onto a 3D reconstruction requires a strong face prior, crafted using diverse data captured across the human population. However, large real datasets are expensive and challenging to collect and typically suffer from low resolution, sampling limitations, and biases in their coverage of diversity and detail of facial geometry and appearance, under varied viewpoints and lighting. Here, our key insight is that such prior can be built from synthetic data alone and fine-tuned on a few real-world images to bridge the synthetic-real domain gap, generalizing robustly to challenging portraits captured in the wild, as shown in Figs. \ref{fig:teaser} and \ref{fig:inthewild}.

Recent works have shown that well-curated, calibrated, and diverse datasets can be built from synthetic graphical renderings. Such data has been successfully applied to various face perception tasks, including sparse problems like landmark localization and segmentation~\cite{wood2021fake,wood20223d} and dense problems like view synthesis~\cite{sun2021nelf} and relighting~\cite{yeh2022learning}. However, such synthetic data is unable to model the full light transport of complex interactions like sub-surface scattering, specular polarization, and global illumination, thus presenting a significant domain gap. Interestingly, sparse regression problems can robustly overcome such gaps at inference time. In contrast, denser synthesis problems cannot, requiring domain adaptation solutions that suffer from quality losses that prevent their use in building a 3D face prior.

We propose a novel, volumetric facial prior that is learned from a large dataset of synthetic images of different identities and expressions, rendered with accessories like hair and beard styles, glasses, clothing, and skin textures.
At inference time, given a few input images at arbitrary resolutions, we fine-tune this prior to reconstruct a high-quality, personalized volumetric model of the captured subject. This new model then allows for 3D-consistent novel view synthesis at high resolution, even when the original input shows exaggerated facial expressions and challenging lighting conditions. We overcome the domain gap between synthetic and real by using an MLP-based generator for volumetric rendering.
This MLP learns a strong prior over geometry while generalizing to unseen appearance domains, including colored lighting and shadows. We demonstrate the efficacy of our method through extensive evaluation, ablations, and comparison with the state of the art.

In summary, our key contributions are:
\begin{itemize}
\item We show that synthetic data alone can be successfully leveraged to learn a strong face prior for few-shot, personalized 3D face modeling in the wild.
\item Our new model and few-shot fine-tuning method can robustly reconstruct expressive faces under challenging in-the-wild lighting and synthesize photorealistic novel views with unprecedented quality and fine-scale idiosyncratic details.
\end{itemize}
Our synthetic dataset is available for research purposes at \\
\datasetlink.
\begin{figure}[t]

\begin{center}
\small
\setlength{\tabcolsep}{2pt}
\newcommand{\gtwidth}{2cm}
\newcommand{\width}{2cm}
\begin{tabular}{ccccccc}
  \includegraphics[width=\gtwidth]{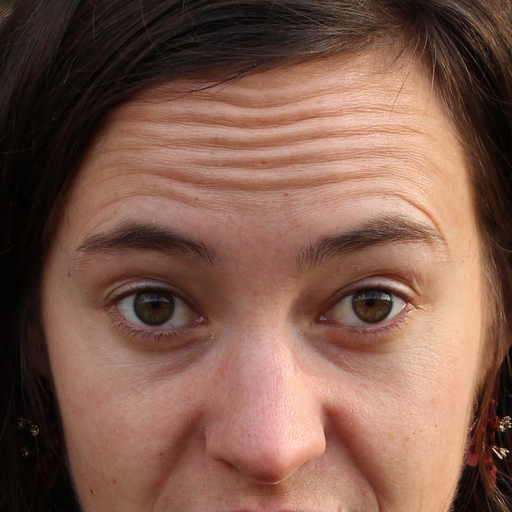}  
  &
  \includegraphics[width=\width]{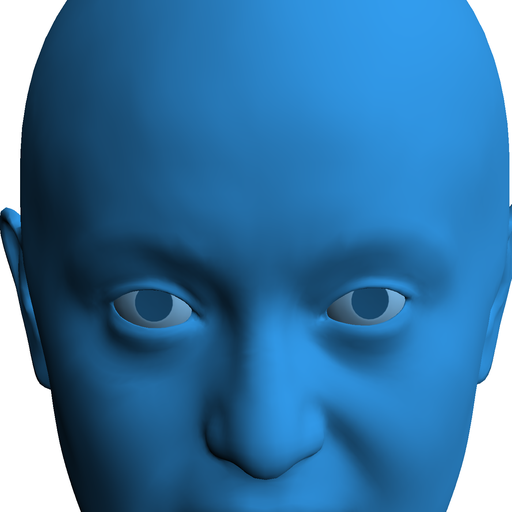} &
  \includegraphics[width=\width]{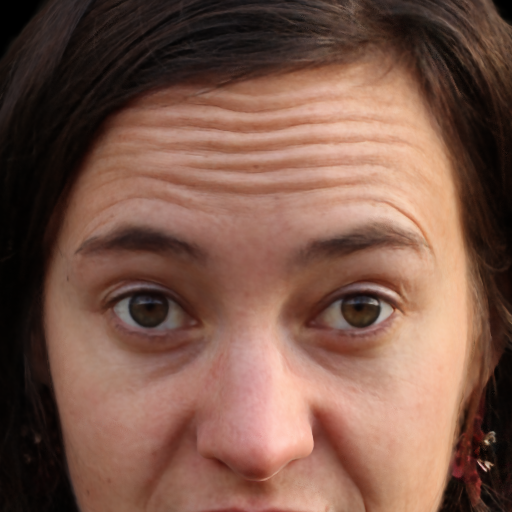}  
  \includegraphics[width=\width]{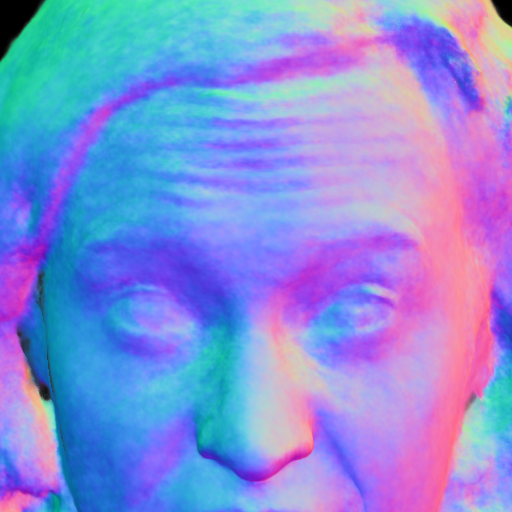}\\
  Input & 3DMM Fit & Reconstruction with Normals
\end{tabular}
\end{center}
\caption{\label{fig:details_in_3dmm}Our method reconstructs details that go far beyond the capabilities of our synthetic dataset. We show a crop of one of the input views from the teaser and its corresponding 3DMM fit. While the 3DMM lacks details, our reconstruction models the wrinkles on the forehead.}
\end{figure}

\begin{figure*}[ht]
    \centering
  \includegraphics[width=\textwidth]
                  {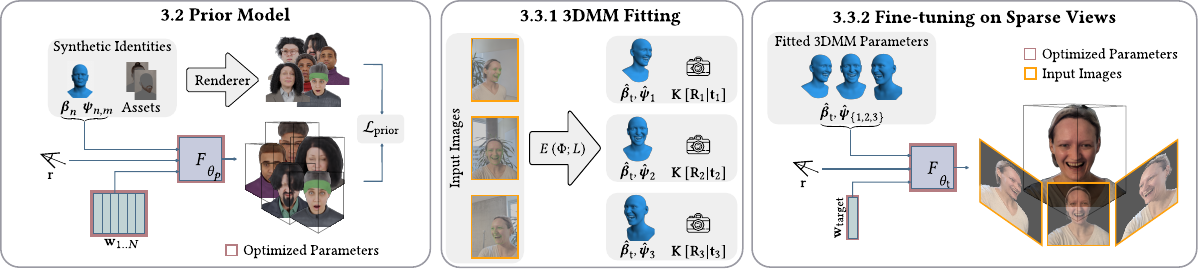}
    \caption{\label{fig:overview}Method Overview. We train an implicit prior model on renderings of a 3DMM combined with assets like hair, beard, clothing, and more (Sec. \ref{sec:prior}). For inference, we first estimate the camera, identity, and expression parameters (Sec. \ref{sec:3dmmfitting}) and then fine-tune the implicit model (Sec. \ref{sec:finetuning}). 
     The implicit prior trained on synthetic data alone can generalize to extremely difficult real-world expressions and render fine details like wrinkles and eyelashes (Figs. \ref{fig:teaser}, \ref{fig:inthewild}).
    While we show three input views for the average case, our method can also work with one (Fig. \ref{fig:singleimage}) or more input views (supplementary material).}
\end{figure*}
\section{Related Work}
Multiple works have explored ways to mitigate the data-intensive nature of volumetric reconstruction, where some main themes have emerged: regularization schemes~\cite{regnerf,Yang2023FreeNeRF,lolnerf,dietnerf,sparf2023}, carefully crafted initialization of model parameters~\cite{tancik2021learned,vora2021nesf,kundu2022panoptic},
depth signals or feature embeddings \cite{dietnerf,sparf2023,wang2023sparsenerf}, and data-driven, pre-trained priors~\cite{lolnerf,eg3d,gu2021stylenerf,h3dnet,voluxgan,pixelnerf,ibrnet,keypointnerf,chen2021mvsnerf,diner,sparf2023,dietnerf,zhang2022fdnerf,buhler2023preface,or2022stylesdf}.

\paragraph{Regularization}
RegNeRF~\cite{regnerf} employs a smoothness regularization term on the expected depth derived from the predicted density, in conjunction with a patch-based regularizer on appearance which together bias the model towards a 3D consistent solution. FreeNeRF~\cite{Yang2023FreeNeRF} proposes a gradual, coarse-to-fine training scheme to prevent overfitting caused by high-frequency, position encoding components early in the fit. Despite their promising results on in-the-wild images, these methods cannot yet provide high-quality reconstructions from few-shot inputs, as we show in Fig.~\ref{fig:main_comparison}.

\paragraph{Initialization}
A common strategy is to learn initial model parameters~\cite{finn2017modelmaml,nichol2018first,sitzmann2020metasdf,Zakharov_2019_ICCV,tancik2021learned} from a large collection of images. While this strategy has been shown to offer faster convergence, its applicability to high-resolution settings remains challenging due to computational requirements of large neural networks.

\paragraph{Data-driven Priors}
With the advent of large datasets of particular domains~\cite{choi2020stargan,liu2018large,karras2019style}, recent works have explored data-driven priors trained on a corpus of data specific to the reconstruction task at hand. Generative neural field models in particular ~\cite{eg3d,gu2021stylenerf,zhou2021CIPS3D,gram,lolnerf,pigan,niemeyer2021giraffe,schwarz2020graf,voluxgan,prao2022vorf,morf,deng2022gram} have shown promising results.
To tackle the heavy computational and memory requirement of volumetric rendering, EG3D~\cite{eg3d} proposes to use a lightweight tri-plane feature representation. Most of these models further rely on an extra 2D super-resolution step~\cite{eg3d,gu2021stylenerf,voluxgan,pigan,wang2023rodin,headnerf} as they can only be trained at low resolutions due to high memory footprint requirements.

Recent works~\cite{morf,cao2022authentic,buhler2023preface} have overcome the need for 2D super-resolution. MoRF~\cite{morf} learns a conditional neural reflectance field of faces from a small training dataset comprising of 12 views of 15 real subjects making neutral expressions, further augmented with synthetic renderings. ~\cite{cao2022authentic} trains an avatar prior with a Mixture of Volumetric Primitives (MVP)~\cite{lombardi2021mvp} from data captured in a controlled environment. They leverage their model to fine-tune to a specific target subject's identity based on a short sequence of a casually captured RGB-D video. Another line of work leverages 2D generative priors \cite{liu2023zero,melas2023realfusion,tang2023make,wu2023reconfusion,zeng2023avatarbooth,comas2024magicmirror}. These methods typically employ score distillation sampling \cite{poole2022dreamfusion,wang2024prolificdreamer} for injecting signals from unseen views given text prompts or sparse input images.

Multiple recent works have leveraged 3D Morphable Face Models (3DMM) \cite{blanz1999morphable} as strong priors. This has led to the development of avatars that can be driven by 3DMM expression coefficients \cite{xu2023latentavatar,zheng2023pointavatar,zheng2022imavatar,regnerf,zhao2023havatar,duan2023bakedavatar,varitex,li2024shellnerf}. Such avatars are typically trained on a sequence of monocular video frames showing various head poses and facial expressions. Other works have explored the reconstruction of faces and textures from a single input image \cite{gao2020portrait,vinod2024teglo,papantoniou2023relightify}. Another body of work has explored novel view synthesis from sparse inputs by training their model as an auto-encoder, coupled with image-based rendering. These methods~\cite{pixelnerf,ibrnet,keypointnerf,chen2021mvsnerf} typically train a convolutional encoder that maps input images to 2D feature maps in images-space to condition the scene's volume. This approach can typically be extended with additional priors such as keypoints~\cite{keypointnerf}, depth~\cite{diner,xu2022sinnerf,wang2023sparsenerf} or pixel-matches across input views~\cite{sparf2023}.

Preface~\cite{buhler2023preface} is a method for high-quality face avatar creation in which they train a low-resolution, generative prior on a dataset of facial identities captured in-studio. 
They could not reconstruct strong expressions due to the domain limitation of their prior model trained only on neutral faces. Our method tackles this challenging problem and we demonstrate compelling examples of novel view synthesis of expressive faces from sparse inputs.

\section{Method}

We present a method that takes as input as few as three images of a person's face -- of arbitrary identity, expression, and lighting condition -- and reconstructs a personalized 3D face model that can render high-quality, photorealistic novel views of that person, including fine details like freckles, wrinkles, eyelashes, and teeth. 

To overcome reconstruction ambiguities, our method uses a pre-trained volumetric face model as prior, trained on a large dataset of synthetic faces with a variety of identities, expressions, and viewpoints rendered in a single environment.
At inference time, our method first fits the coefficients of our prior model to a small set of real input images. It then further fine-tunes the model weights and effectively performs domain adaptation during the few-shot reconstruction process, see Fig.~\ref{fig:overview}.
While the prior model is trained only once on a large collection of synthetic face images, the inference-time optimization is performed on a per-subject basis from as few as three (\eg smartphone) images captured in-the-wild (see Fig. \ref{fig:inthewild}). 

This section starts with background information (Sec. \ref{sec:background}), details the training of synthetic prior (Sec. \ref{sec:prior}) and then finetuning from three input views (Sec.~\ref{sec:inference}).

\subsection{Background: NeRF and Preface}
\label{sec:background}

\paragraph{NeRF}
Our prior face model is built upon Neural Radiance Fields (NeRF) \cite{mildenhall2020nerf}. NeRFs represent 3D objects as density and (emissive) radiance fields parameterized by neural networks. Given a camera ray $\textbf{r}$, a NeRF samples 3D points $\textbf{x}$ along the ray that are fed together with the view direction $\textbf{d}$ into an MLP. The output is the corresponding density $\sigma$ and color value $\textbf{c}$ at $\textbf{x}$. A NeRF is rendered into any view via volumetric rendering. The color ${\bf c({\bf p})}$ of a pixel ${\bf p}$ is determined by compositing the density and color along the camera ray $\mathbf{r}$ within an interval between a near and a far camera plane $[t_n, t_f]$:

\begin{align}
    \mathbf{c}(\mathbf{p}) = \textbf{F}_\theta(\textbf{r}) = \int_{t_n}^{t_f}T(t)\sigma(\textbf{r}(t))\textbf{c}(\textbf{r}(t),\textbf{d}) dt \text{,} \label{eq:nerf}\\
    \text{ where } T(t) = \text{exp} \left( - \int_{t_n}^{t} \sigma(\textbf{r}(s)) ds \right).
\end{align}
The variable $\theta$ denotes the model parameters that are fit to the input data.

\paragraph{Preface} Our method also extends Preface~\cite{buhler2023preface}, a method for novel view synthesis of neutral faces given sparse inputs.
Besides the position $\textbf{x}$ and view direction $\textbf{d}$, Preface also takes a learned latent code $\textbf{w}$ as input. The latent code represents the identity and is optimized while training the model as an auto-decoder \cite{bojanowski2018optimizing}. During inference, Preface first projects the sparse input images into its latent space, by optimizing one identity code $\textbf{w}$. Then, it fine-tunes all model parameters under regularization constraints on the predicted normals and the view weights.
While Preface excels at high-resolution novel view synthesis of neutral faces, it struggles in the presence of strong, idiosyncratic expressions (Fig. \ref{fig:main_comparison}). In the following, we address this limitation while building an improved prior from synthetic images alone.

\subsection{Pretraining an Expressive Prior Model}
\label{sec:prior}

\begin{figure}[t]
    \centering
  \includegraphics[width=0.8\linewidth]
                  {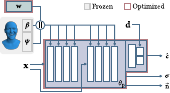}
    \caption{\label{fig:prior_model} We implement our prior model as a conditional NeRF \cite{mildenhall2020nerf,mipnerf360}. We condition by concatenating ($||$) three codes: a 3DMM identity code $\boldsymbol{\beta}$, a 3DMM expression code $\boldsymbol{\psi}$, and a learned latent code $\textbf{w}$ representing out-of-model characteristics like hair, clothing, etc. The outputs include the color $\boldsymbol{\hat c}$, the density $\sigma$, and a normal vector $\vec{\text{n}}$. Please see Sec. 2 in the supp. PDF for details.}
\end{figure}

We train a prior model to capture the distribution of human heads with arbitrary facial expressions. 
As Preface \cite{buhler2023preface}, our prior model is implemented as a conditional Neural Radiance Field (NeRF) \cite{mildenhall2020nerf,lolnerf} with a Mip-NeRF 360 backbone \cite{mipnerf360}, Fig.~\ref{fig:prior_model}.
Yet, we observe that the simple Preface auto-decoder \cite{lolnerf,bojanowski2018optimizing} cannot achieve high-quality fitting results on expressive faces (see the ablation in Sec.~\ref{sec:ablation}).
We hypothesize that the distribution of expressive faces is much more difficult to model and disentangle than the distribution of neutral faces. 
To address this limitation, we decompose the latent space of our prior model into three latent spaces: a predefined identity space $\mathcal{B}$, a predefined expression space $\Psi$, and a learned latent space $\mathcal{W}$.
The identity and expression spaces come from a linear 3D Morphable Model (3DMM) in the style of \cite{wood2021fake}.
The latent codes in these two spaces are known {\em a priori} and represent the face shape for the arbitrary identity and expression in each synthetic training image.
These codes are also frozen and do not change during training.
The latent space $\mathcal W$ represents characteristics that are not modeled by the 3DMM like hair, beard, clothing, glasses, appearance, lighting, etc., and is learned while training the auto-decoder as in Preface \cite{buhler2023preface}.
Considering this model, we adapt Eq.~\ref{eq:nerf} to obtain: 
\begin{align}
    \textbf{F}_{\theta_\text{p}}(\textbf{r}, \boldsymbol{\beta}, \boldsymbol{\psi}, \mathbf{w}) = \int_{t_n}^{t_f}T(t)\sigma(\textbf{r}(t),\boldsymbol{\beta},\boldsymbol{\psi},  \mathbf{w})\textbf{c}(\textbf{r}(t),\textbf{d},\boldsymbol{\beta},\boldsymbol{\psi},  \mathbf{w}) dt \text{,} \label{eq:ournerf}\\
    \text{ where } T(t) = \text{exp} \left( - \int_{t_n}^{t} \sigma(\textbf{r}(s),\boldsymbol{\beta},\boldsymbol{\psi}, \mathbf{w}) ds \right),
\end{align}
where $\boldsymbol{\beta} \in \mathcal{B} \subset \mathbb{R}^{48}$ and $\boldsymbol{\psi} \in \Psi \subset \mathbb{R}^{157}$ are the 3DMM identity and expression parameters, and $\mathbf{w} \in \mathcal{W} \subset \mathbb{R}^{64}$ is a learned parameter encoding additional characteristics.

We train this prior on synthetic data alone -- it never sees a real face. While it would be feasible to train a prior model on real data (see our ablation in Tbl. \ref{tbl:ablation_dataset}), we chose synthetic over real for multiple reasons. Real datasets exhibit limited diversity.
Most face datasets feature monocular frontal views only, with few expressions other than smiles.
Some multi-view, multi-expression datasets exist \cite{wuu2022multiface,kirschstein2023nersemble,facescape}, but consist of relatively few individuals due to the complexity and expense of running a capture studio.
Further, subjects must adhere to wardrobe restrictions: glasses are forbidden and hair must be tucked away.
A prior trained on such data will not generalize well to expressive faces captured in the wild. Besides, the logistics of capturing large-scale real data is extremely expensive, time and energy-consuming, and cumbersome.
Instead, synthetics guarantee us a wide range of identity, expression, and appearance diversity, at orders of magnitude lower cost and effort. In addition, synthetics provide perfect ground truth annotations: each render is accompanied by 3DMM latent codes $\boldsymbol{\beta}, \boldsymbol{\psi}$.

\begin{figure}[t]
\centering
\includegraphics[width=\linewidth]{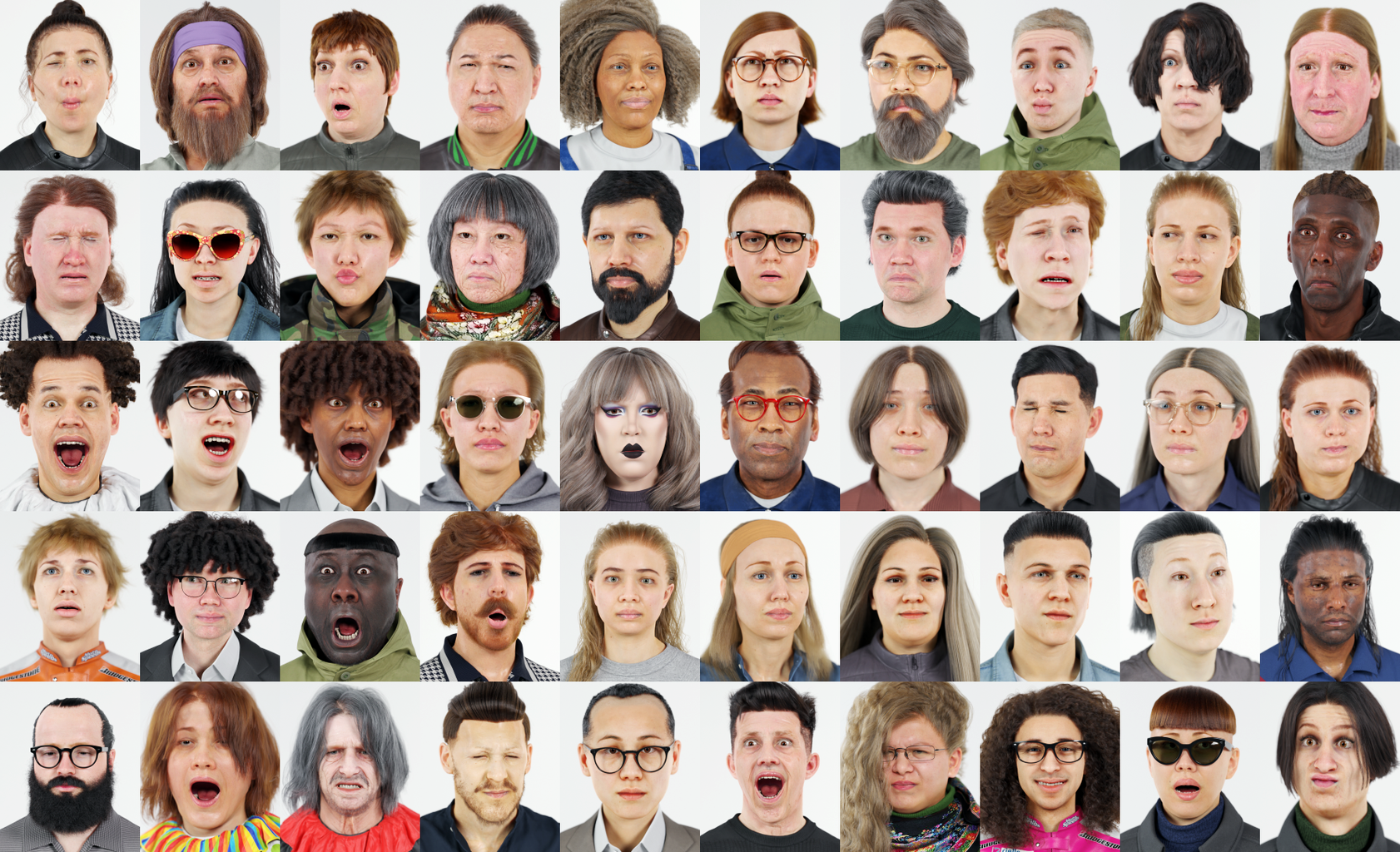}
    \caption{Our synthetic faces exhibit a wide range of geometric and expression diversity and can be rendered from any viewpoint.\label{fig:synthetics-examples}}
\end{figure}

We synthesize facial training data as in ~\citet{wood2021fake}.
We first generate the 3D face geometry by sampling the identity and expression spaces of the 3DMM.
We then make these faces look realistic by applying physically based skin materials, attaching strand-based hairstyles, and dressing them up with clothes and glasses from our digital wardrobe.
The scene is rendered with environment lighting using Cycles, a physically-based ray tracer (\texttt{www.cycles-renderer.org}).
Examples are shown in Fig. \ref{fig:synthetics-examples} and on the supplementary HTML page.
To help disentangle identity from expression, we sample 13 different random expressions for each random identity.
Each expression is then rendered from 30 random viewpoints around the head.
All faces are rendered under the same lighting condition, which was chosen to minimize shadows on the face.

Each training iteration randomly samples rays $\bf r$ from a subset of all identities and expressions. A ray is rendered into a pixel color as given by Eq.~\ref{eq:ournerf}. 
We optimize the network parameters $\bf \theta_\text{p}$ and $\text{N}$ per-identity latent codes $\mathbf{w}_{1..\text{N}}$ while keeping the 3DMM expression and identity codes $\boldsymbol{\beta}$ and $\boldsymbol{\psi}$ frozen:
\begin{align}
\label{eq:priorloss}
    \theta_\text{p}, \textbf{w}_\text{1..N} = \argmin_{\bf \theta, \textbf{w}_\text{1..N}} \mathcal{L}_\text{prior}\,,
    \qquad
    \mathcal{L}_\text{prior} = \mathcal{L}_\text{recon} + \lambda_\text{prop} \mathcal{L}_\text{prop}.
\end{align}
Here $\mathcal{L}_\text{recon}$ is the mean-absolute error between the predicted and ground-truth colors, and $\mathcal{L}_\text{prop}$ is the weight distribution matching loss from Mip-NeRF~\cite{barron2021mip}. Please see Sec. 2 in the supp. PDF for the spelled-out loss terms.

We find that, when training from scratch, the model quickly collapses and outputs zero densities everywhere. We solve this by first training on images with background for 50,000 steps and continuing without background.

\subsection{Inference from Sparse Views}
\label{sec:inference}
We use our low-resolution synthetic prior model to enable high-resolution novel view synthesis of real expressive faces from few input images. We first describe how we obtain the conditional inputs and camera parameters in Sec. \ref{sec:3dmmfitting} and the subsequent fine-tuning of our model in Sec. \ref{sec:finetuning}.

\subsubsection{3DMM Fitting and Camera Estimation}
\label{sec:3dmmfitting}

\begin{figure}[t]
    \centering
  \includegraphics[width=\linewidth]
                  {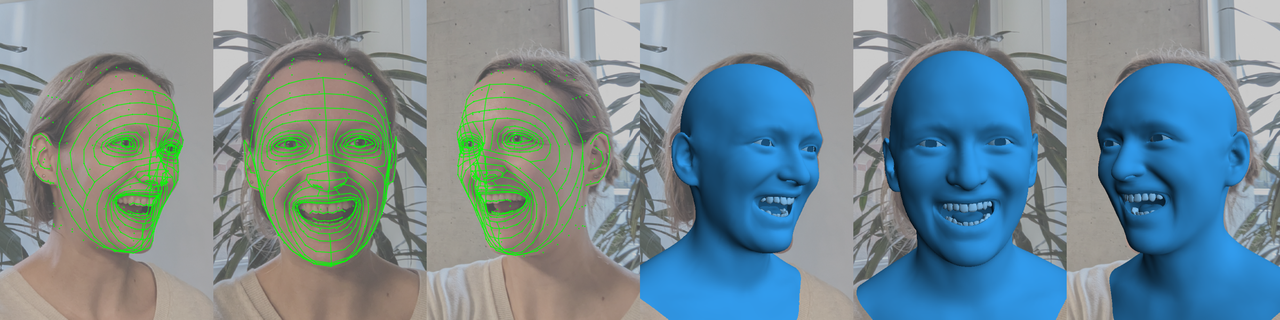}
    \caption{For inference, we first recover the 3DMM and camera parameters through model-fitting to probabilistic 2D landmarks (Sec.~\ref{sec:3dmmfitting}).\label{fig:landmarks}}
\end{figure}

Figure \ref{fig:overview} gives an overview of the 3DMM fitting.
During inference, the first step is to recover camera and 3DMM parameters from un-calibrated input images.
We follow the approach of previous work ~\cite{wood20223d} and fit to dense 2D landmarks (see Fig.~\ref{fig:landmarks}).
We first predict 599 2D probabilistic landmarks.
Each landmark corresponds to a vertex in our 3DMM and is predicted as a 2D isotropic Gaussian with expected 2D location $\mu$ and scalar uncertainty $\sigma$.
Next, we minimize an energy $E(\boldsymbol{\Phi}; L)$, where $L$ denotes the landmarks and $\boldsymbol{\Phi}$ all the optimized 3DMM parameters including identity, expressions, joint rotations, and global translation, and intrinsic and extrinsic camera parameters, if unknown.

Minimizing $E$ encourages the 3DMM to explain the observed landmarks with a probabilistic 2D landmark energy, and discourages unlikely faces using regularizers on 3DMM parameters and mesh self-intersection (additional detail is given in~\cite{wood20223d} and in Sec. 2 of the supp. mat.).

The benefit of the 3DMM fitting is two-fold. First, we get a good estimate of the world position of the camera and the head, so that later during inversion and finetuning of the model, camera parameters can be frozen. Second, thanks to the alignment of the 3DMM latent space and our prior model's latent space, we directly feed the 3DMM parameters into the model, which serves as a good initialization during the subsequent inversion stage. 

The outputs of this step are the camera parameters (shared intrinsics $\boldsymbol{\text{K}}$ and per-camera extrinsics $[\boldsymbol{\text{R}}_i|\boldsymbol{\text{t}}_i]$), a shared identity code $\boldsymbol{\beta}$, and per-image expression codes $\boldsymbol{\psi}_i$.
For casual in-the-wild captures, it can be challenging to hold the same expression while the data is being captured.
Therefore, we allow expression code to vary slightly between images to make the inversion robust to small, involuntary micro-changes in expression.
In the studio setting, however, the cameras are synchronized and hence it is sufficient to optimize for a single, shared expression code.

\subsubsection{Fine-tuning on Sparse Views}
\label{sec:finetuning}
This section describes how to fine-tune the low-resolution synthetic prior model to sparse, high-resolution real input images. Fine-tuning requires a short warm-up phase where only the latent code for the target $\mathbf{w}_\text{target}$ is optimized. After that, fine-tuning optimizes all model parameters under additional constraints on the geometry and the appearance weights. We randomly sample rays from all available inputs, typically three images, and mask them to the foreground by multiplying them by an estimated foreground mask \cite{pandey2021total}.

\paragraph{Warm-up by Latent Code Inversion}
While the 3DMM fitting provides the identity and expression codes $\boldsymbol{\beta}, \boldsymbol{\psi}_\text{i}$, our model also requires the conditional input $\mathbf{w}$, which models out-of-model characteristics like hair, clothing, and appearance. We follow \cite{buhler2023preface} and search the learned latent space $\mathcal{W}$ of the prior model for a latent code that roughly matches the geometry and appearance of the input images. We downscale the three input images to the resolution of the prior model, sample random patches, and optimize

\begin{equation}
    \label{eq:inversion}
    \mathbf{w}_\text{target} = \argmin_{\mathbf{w}} \mathcal{L}_\text{recon} + \lambda_\text{LPIPS} \mathcal{L}_\text{LPIPS}.
\end{equation}

The photo-metric reconstruction term $\mathcal{L}_\text{recon}$ is the mean absolute error between the rendered and the ground-truth patch and $\mathcal{\text{LPIPS}}$ is a perceptual loss in the feature space of a pre-trained image classifier \cite{zhang2018perceptual,simonyan2014veryvgg}. Note that the LPIPS loss is only employed during inversion, not during model fitting.
The camera, 3DMM identity, and expression parameters are frozen during inversion.

\paragraph{Model Fitting}
The output of the warm-up is a rough approximation of the input images in a \emph{low-resolution, synthetic} space. In model fitting, we cross the domain gap to enable detailed novel view synthesis at \emph{high resolution} for \emph{realistic} faces. The model fitting needs to cross a substantial domain gap so that the output can contain details that have never been seen during prior model training.

Model fitting optimizes all model parameters on sparse, usually two or three, input images. In a randomly initialized NeRF \cite{mildenhall2020nerf}, this optimization would overfit and would fail to produce correct novel views \cite{Yang2023FreeNeRF,sparf2023,buhler2023preface}. Thanks to our pretrained prior model, we can employ both \emph{implicit} and \emph{explicit} regularization, which yields high-quality results even in such sparse settings.

Implicit regularization comes from the fact that our prior model is trained on an aligned dataset of human faces. Initializing the weights of a NeRF with the correct latent code and weights of a prior model avoids total collapse. However, the optimization can still produce strong artifacts like duplicate ears and view-dependent color distortions. 
We follow \cite{buhler2023preface} and add explicit regularization on the consistency of predicted vs. analytical normals and an L2 regularization on the weight of the view branch to avoid view-dependent flickering. In addition, we add a distortion loss term \cite{mipnerf360} $\mathcal{L}_\text{dist}$ for a more compact geometry:

\begin{align}
\label{eq:modelfittingloss}
    \mathbf{\theta}_{\text{t}}, \mathbf{w}_{\text{target}} = \argmin_{\mathbf{\theta}, \mathbf{w}} \mathcal{L}_\text{fit} &= \mathcal{L}_\text{recon} + \lambda_\text{prop} \mathcal{L}_\text{prop}\\
    &+ \lambda_\text{normal} \mathcal{L}_\text{normal} + \lambda_{d} \mathcal{L}_{d} + \lambda_\text{dist} \mathcal{L}_\text{dist} \notag
\end{align}
where $\mathcal{L}_\text{normal}$ and $\mathcal{L}_{d}$ are the regularizers on predicted normals and view weights from Preface. We list and explain all loss terms in more detail in Sec. 2 of the supp. PDF.

\paragraph{Inference In-the-wild}
For in-the-wild (ITW) images, we capture three images sequentially with a hand-held camera. The captured face might inhibit small movements during the capture, called micromotions.
To mitigate these micromotions, we fine-tune with individual expression code for every input image. The 3DMM fitting yields a per-image expression code $\boldsymbol{\hat\psi}_{i}$.
During inference, we interpolate these expression codes based on their distance to the target camera. The weight is computed as the inverse squared distance between the target and all training cameras. The interpolated expression code  $\boldsymbol{\tilde{\psi}}_t$ for a target camera is computed as 
\begin{align}
    \boldsymbol{\tilde\psi}_t &= \sum_i \underbrace{\frac{Z}{\epsilon + || \boldsymbol{\tilde{l}}_t - \boldsymbol{\hat{l}}_i||^2_2}}_{\text{weight}}  \underbrace{\boldsymbol{\hat\psi}_i}_{\text{training expression}},
\end{align}
where $\boldsymbol{\hat\psi}_i$ are the expression codes of the training frames, $\boldsymbol{\tilde{l}_t}$ is the position of the target camera, $\boldsymbol{\hat{l}}_i$ are the positions of the training cameras, $\epsilon$ is a small constant, and $Z$ is a normalization factor to ensure that the weights sum up to 1.

\section{Experiments}
This section presents an extensive evaluation of our proposed method with both quantitative and qualitative comparisons to related work and additional ablations. For comparisons, we run publicly available code for SparseNeRF \cite{wang2023sparsenerf}, Sparf \cite{sparf2023}, and Diner \cite{diner}. For FreeNeRF \cite{Yang2023FreeNeRF} and Preface \cite{buhler2023preface}, we use our own implementation.
We compare renders at the resolution $1334\times2048$ pixels, except for Diner, where we render at a lower resolution ($160 \times 256$ pixels) due to its architecture and memory constraints.
Please refer to the supplementary video and HTML page for more results.

\subsection{Quantitative Evaluation}

\begin{table}
\caption{Quantitative evaluation on the Multiface dataset \cite{wuu2022multiface}.}
\label{tbl:main_comparison}
\small 
\begin{center}
\begin{tabular}{l | ccc}
\hline
\textbf{Method} &
    \multicolumn{1}{c}{\textbf{PSNR} $\uparrow$} &
    \multicolumn{1}{c}{\textbf{SSIM} $\uparrow$} & 
    \multicolumn{1}{c}{\textbf{LPIPS} $\downarrow$}\\
\hline
Sparse NeRF \cite{wang2023sparsenerf} & 16.29 & 0.6470 & 0.4024 \\
Diner \cite{diner} & 17.08 & 0.6608 & 0.3123\\
SPARF \cite{sparf2023} & 18.38 & 0.6401 & 0.4032 \\
FreeNeRF \cite{Yang2023FreeNeRF} & 21.42 & 0.7093 & 0.3612\\
Preface \cite{buhler2023preface} & 25.02 & 0.7539 & 0.3129 \\
\hline
\textbf{Ours} & \textbf{26.49} & \textbf{0.7721} & \textbf{0.2970} \\
\end{tabular}
\end{center}
\end{table}

\begin{table}[ht]
\caption{We ablate the performance for different variants of the prior model. We pre-train a) on a smaller training set with fewer subjects, b) for a shorter number of steps, c) with different configurations of our synthetic rendering pipeline, and d) include real data. The \emph{full} prior model is trained on 1,500 subjects with 13 expressions each for 1 Mio. steps on synthetic renderings with all available accessories in a single environment map. Please see the supp. mat. for details of the rendering pipeline.
\label{tbl:ablation_dataset}}
\small 
\begin{center}
\begin{tabular}{llccc}
\hline
& \textbf{Pre-training Variant} &
    \multicolumn{1}{c}{\textbf{PSNR} $\uparrow$} &
    \multicolumn{1}{c}{\textbf{SSIM} $\uparrow$} & 
    \multicolumn{1}{c}{\textbf{LPIPS} $\downarrow$}\\
\hline
a.i) & No Pre-training & 10.21 & 0.3448 & 0.4256 \\
a.ii) & On 1 subject & 24.00 & 0.7512 & 0.3358 \\
a.iii) & On 15 subjects & 25.47 & 0.7668 & 0.3248 \\
a.iv) & On 1,500 subjects & 26.54 & 0.7750 & 0.3144 \\
\hline
b.i) & For 105K steps & 26.64 & 0.7752 & 0.3208 \\
b.ii) & For 500K steps & 26.54 & 0.7750 & 0.3144 \\
b.iii) & For 1 Mio. steps & 26.49 & 0.7721 & 0.2970 \\
\hline
c.i) & On gray-scale renderings & 26.44 & 0.7740 & 0.3242 \\
c.ii) & On low-quality renderings  & 26.53 & 0.7727 & 0.3401 \\
c.iii) & In diverse environments & 26.58 & 0.7727 & 0.3415 \\
c.iv) & Without makeup & 26.72 & 0.7755 & 0.3210 \\
c.v) & Without accessories & 26.00 & 0.7731 & 0.3484 \\
c.vi) & Without hair & 26.14 & 0.7727 & 0.3303 \\
c.vii) & With all accessories & 26.54 & 0.7750 & 0.3144 \\
\hline
d.i) & On real images & 26.14 & 0.7708 & 0.3237 \\
d.ii) & On real \emph{and} synthetic images & 26.41 & 0.7726 & 0.3227 \\
d.iii) & On synthetic images & 26.54 & 0.7750 & 0.3144 \\
\hline
e) & Full & 26.49 & 0.7721 & 0.2970 \\
\end{tabular}
\end{center}
\end{table}

\begin{figure*}[ht]
\begin{center}
\small
\setlength{\tabcolsep}{2pt}
\newcommand{\width}{2cm}
\newcommand{\inputwidth}{2.2cm}
\begin{tabular}{c ccccccc}
  \includegraphics[width=\inputwidth]{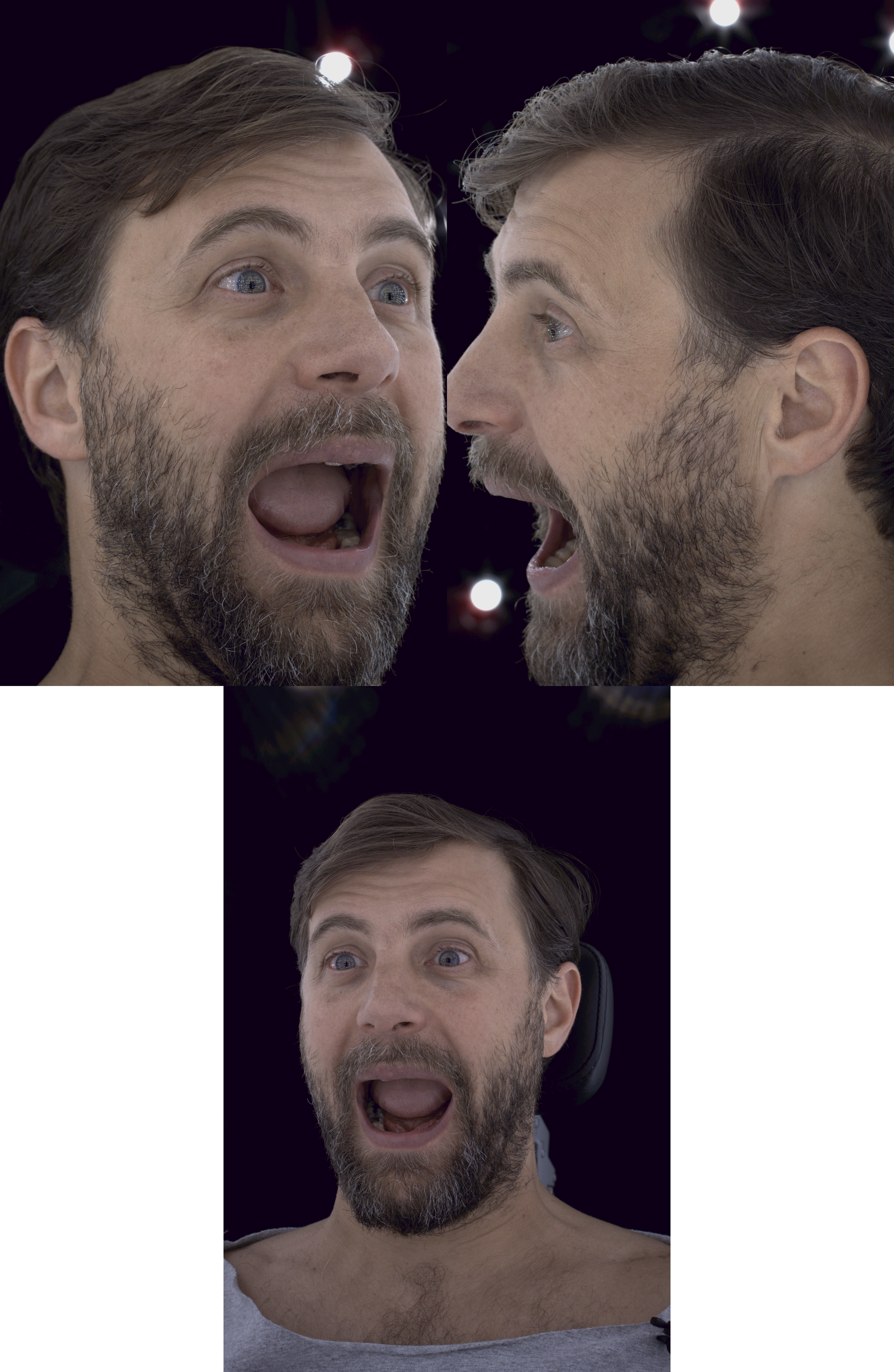}  &
      \includegraphics[width=\width]{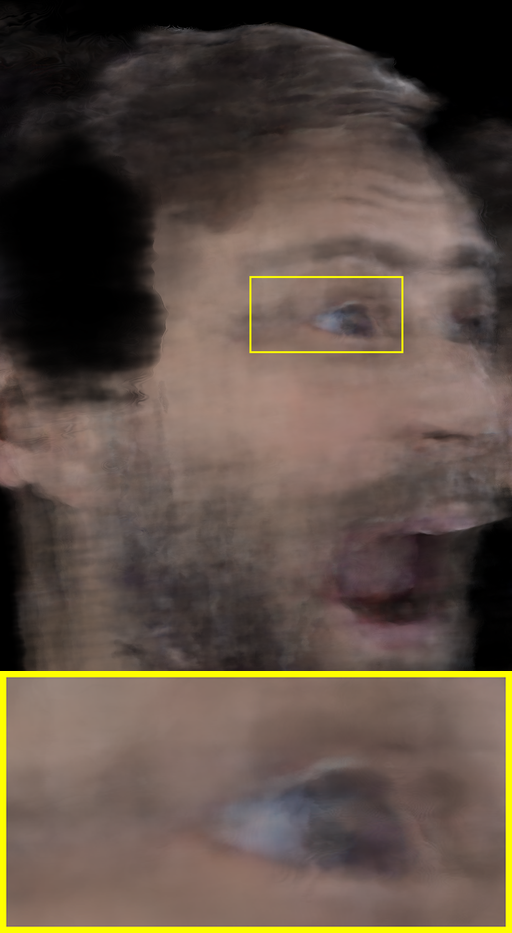}  &
      \includegraphics[width=\width]{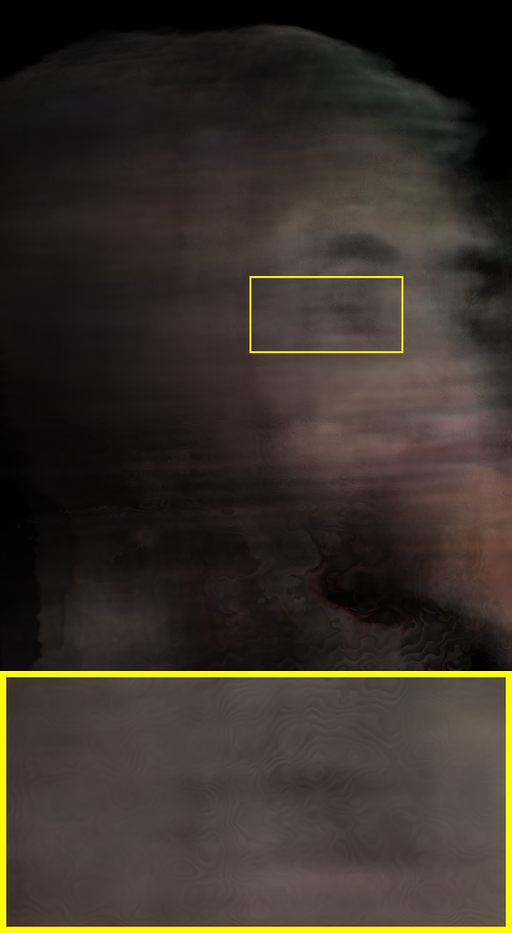}  &
      \includegraphics[width=\width]{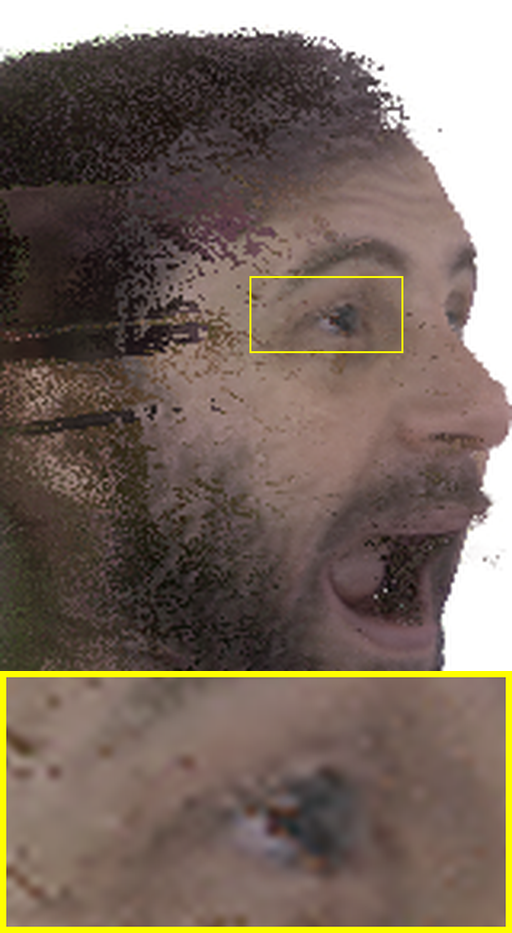}  &
      \includegraphics[width=\width]{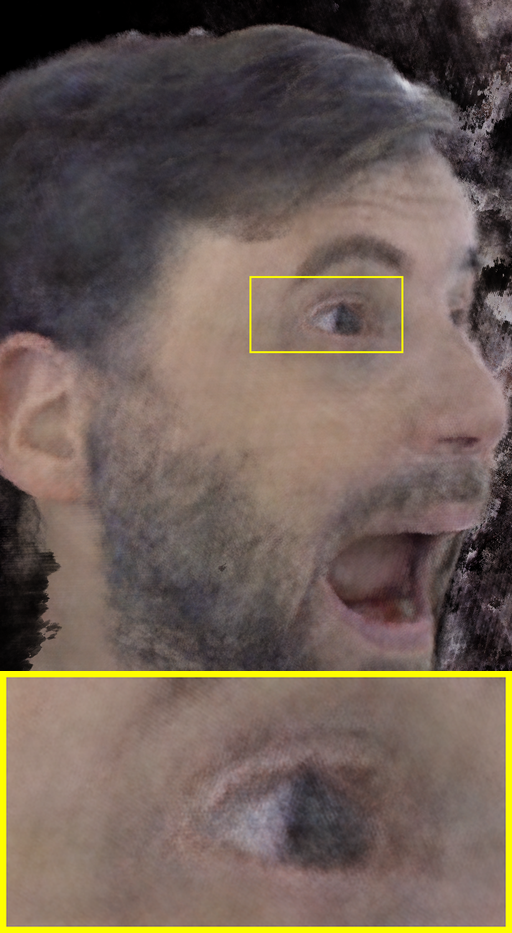} &
      \includegraphics[width=\width]{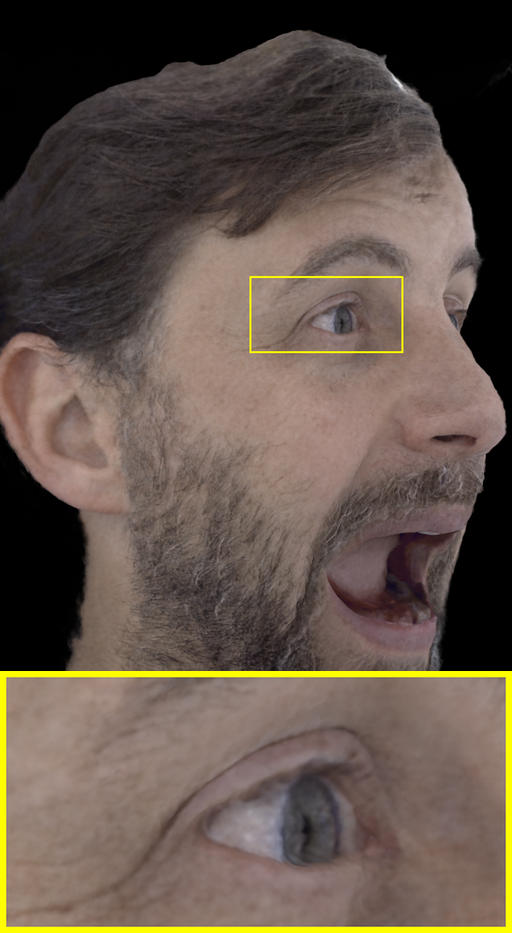}     &
      \includegraphics[width=\width]{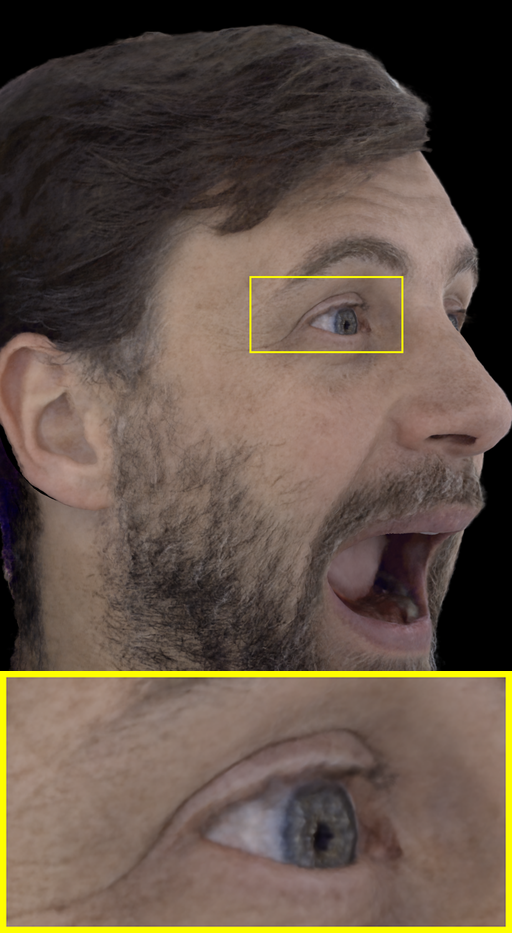}  &
      \includegraphics[width=\width]{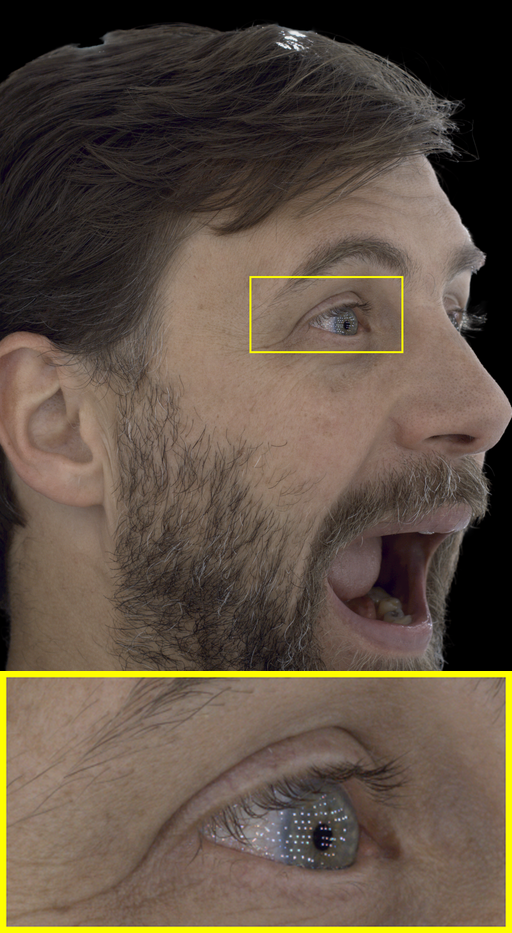}  \\
      \includegraphics[width=\inputwidth]{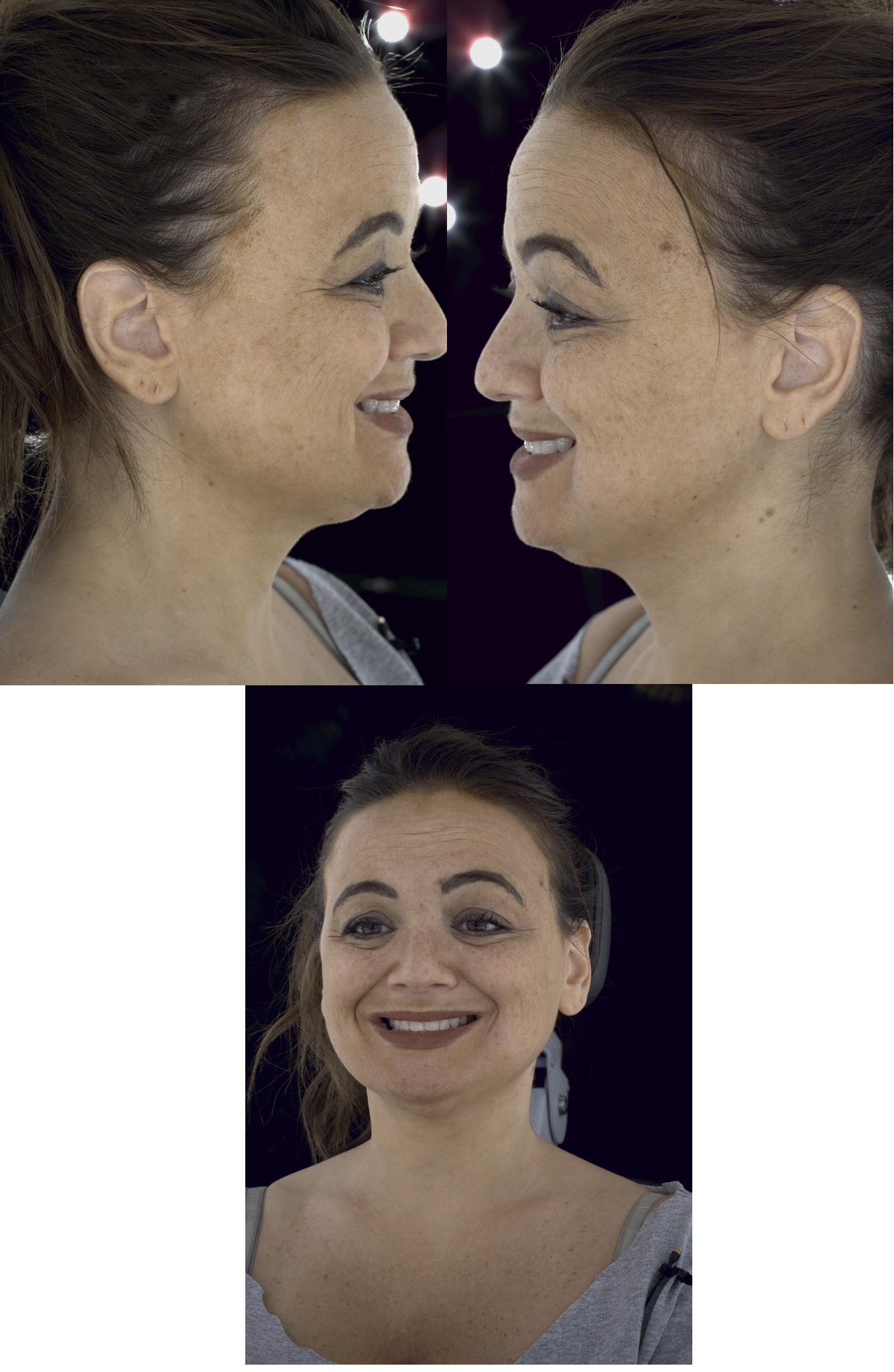} &
      \includegraphics[width=\width]{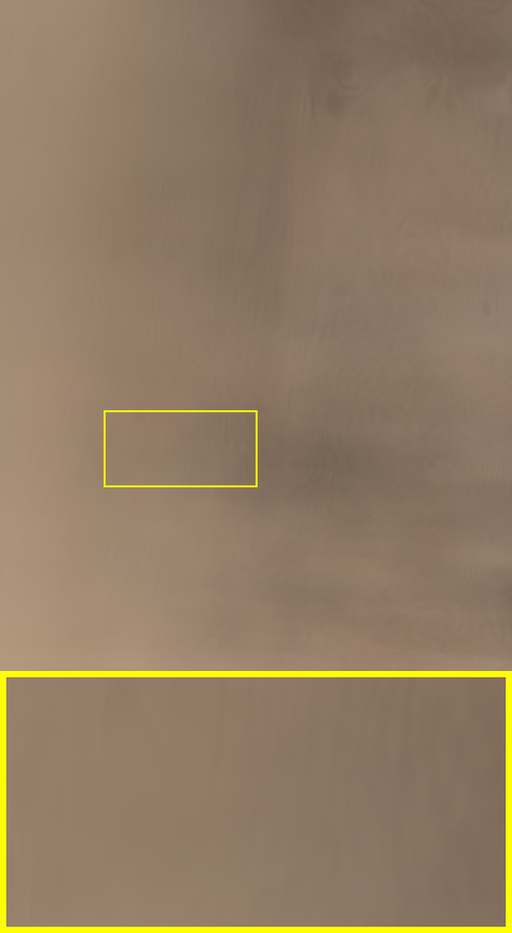}  &
    \includegraphics[width=\width]{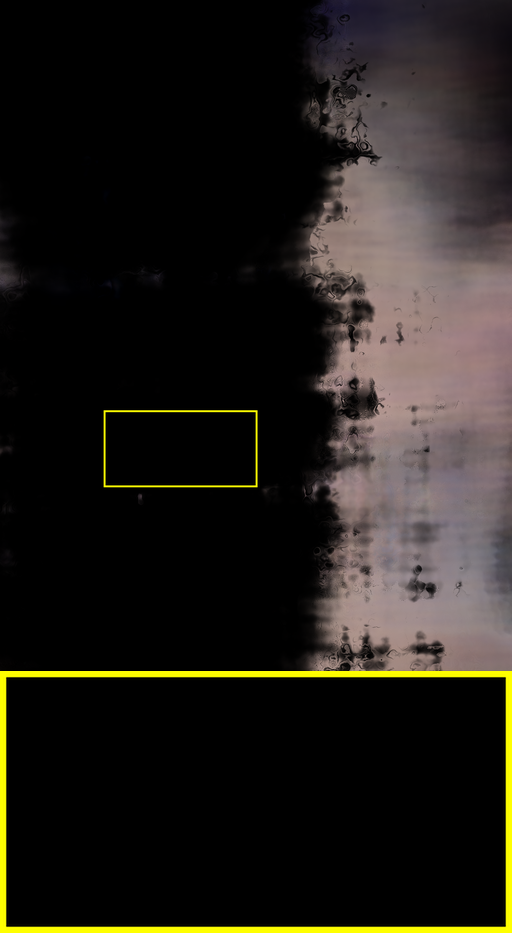}  &
    \includegraphics[width=\width]{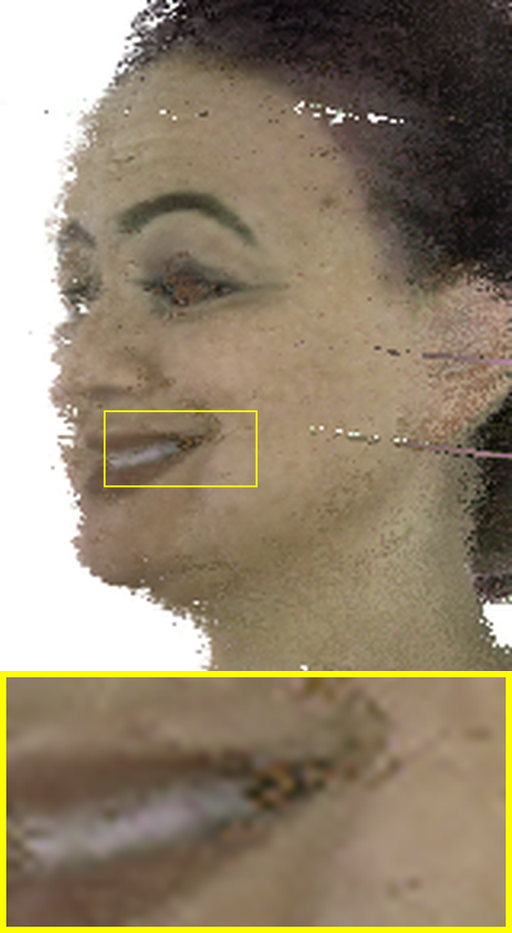}   &
    \includegraphics[width=\width]{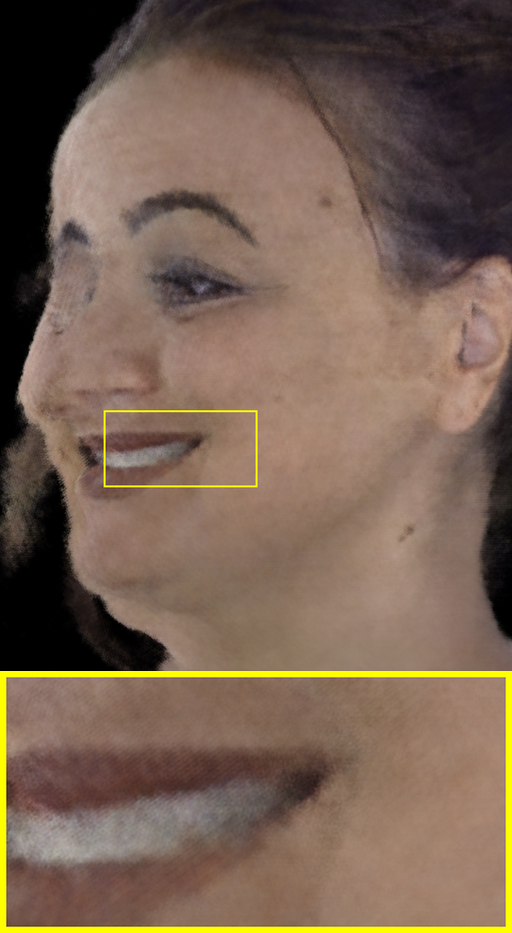}  &
    \includegraphics[width=\width]{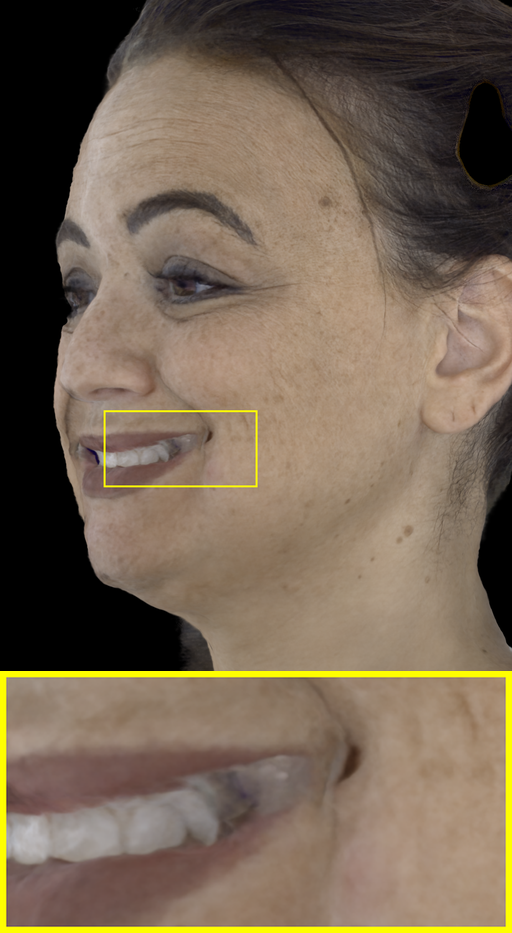}  &
    \includegraphics[width=\width]{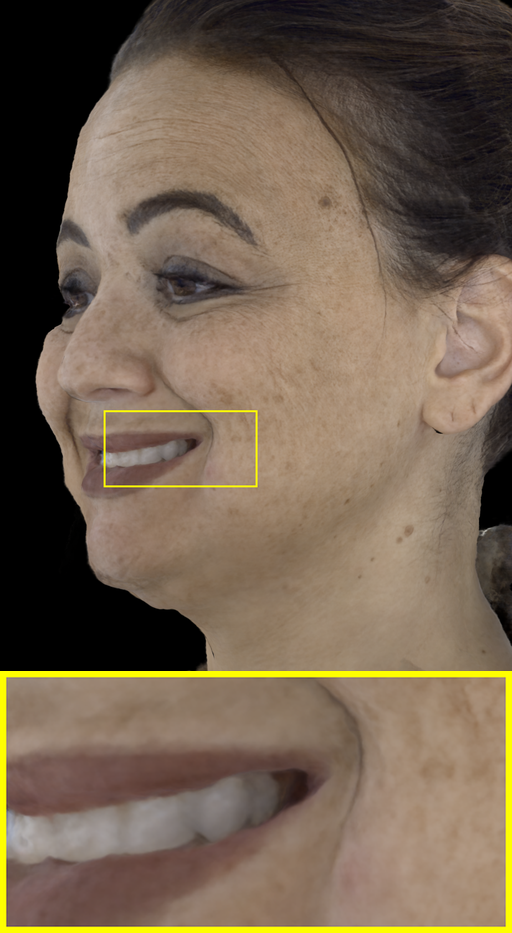}  &
    \includegraphics[width=\width]{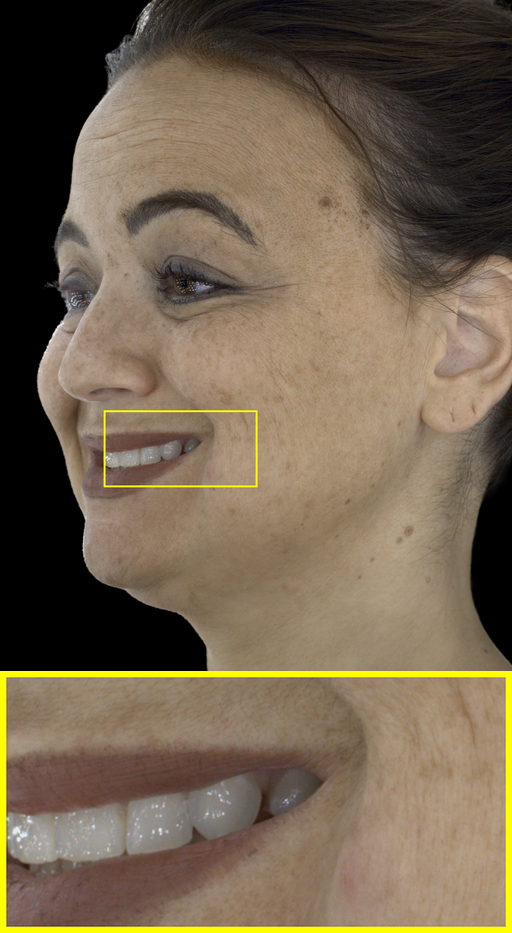}  \\
  \includegraphics[width=\inputwidth]{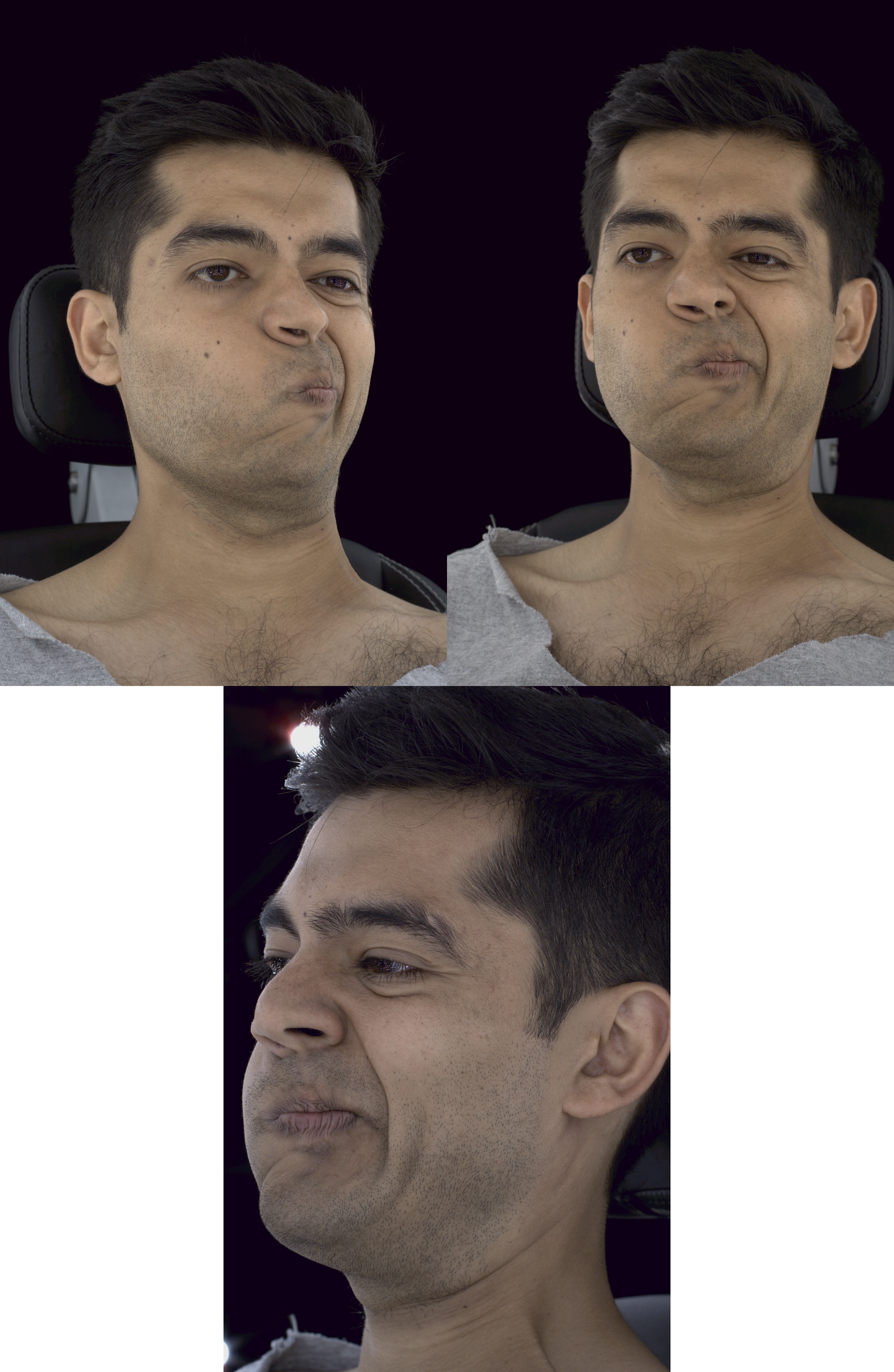} &
      \includegraphics[width=\width]{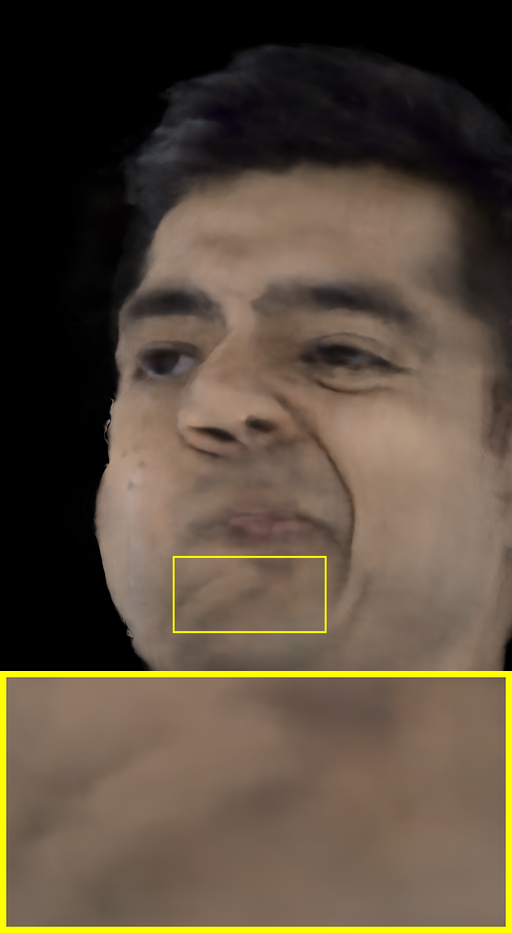}  &
    \includegraphics[width=\width]{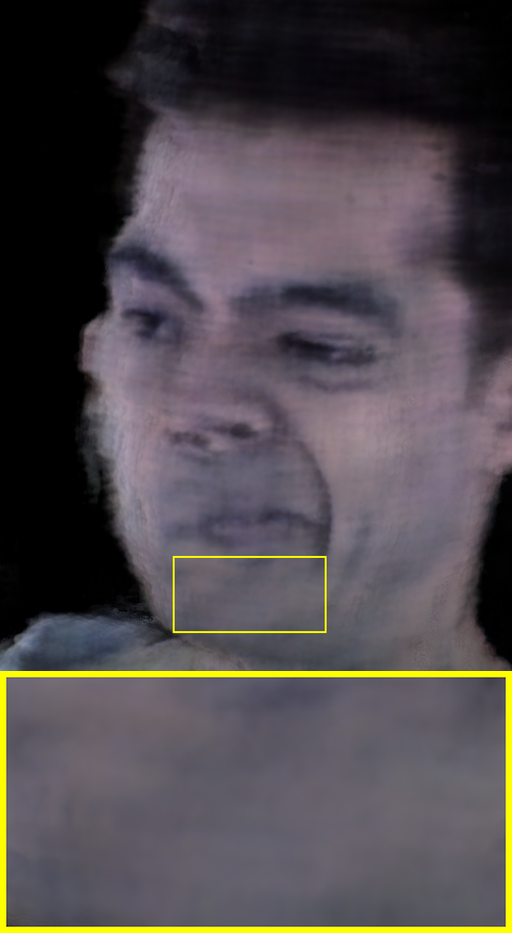}  &
    \includegraphics[width=\width]{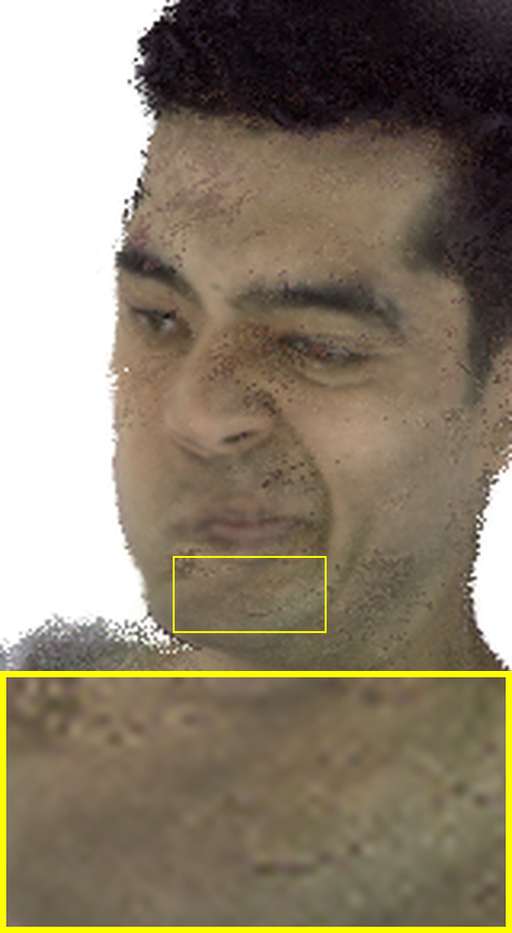}  &
    \includegraphics[width=\width]{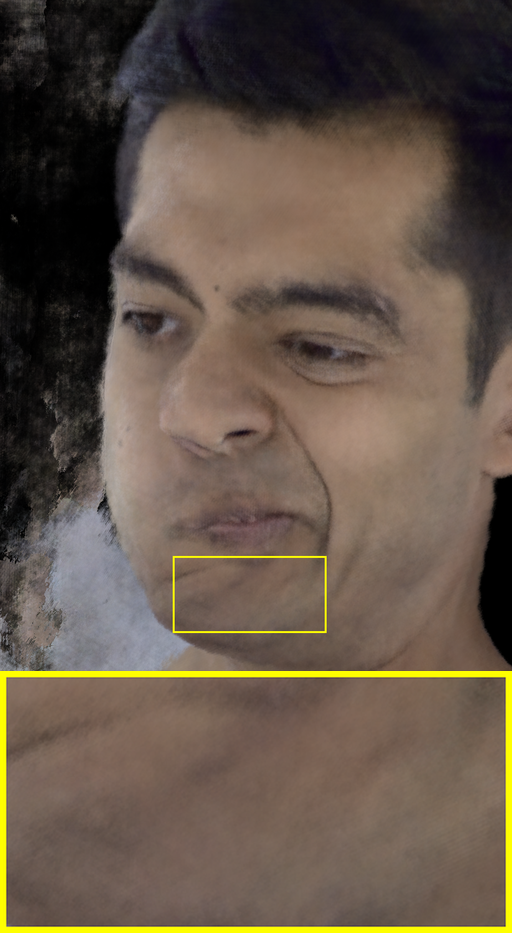}  &
    \includegraphics[width=\width]{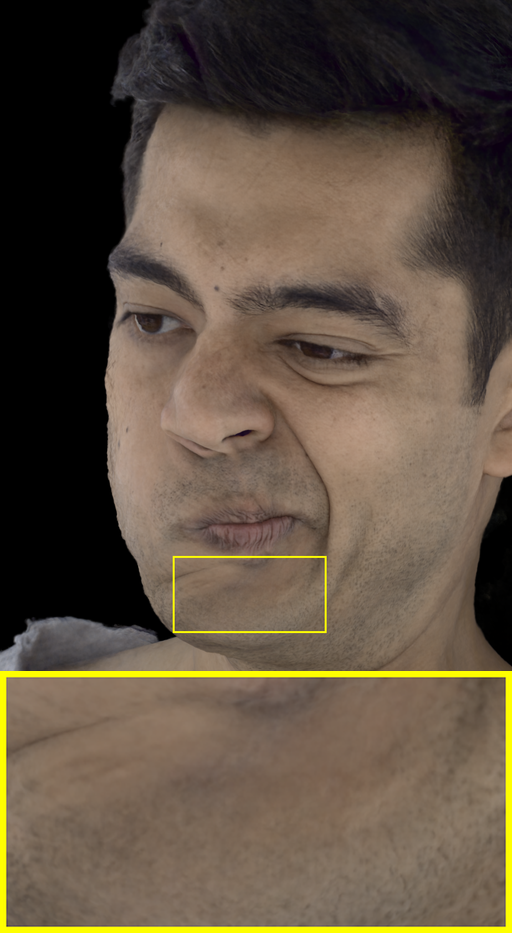}  &
    \includegraphics[width=\width]{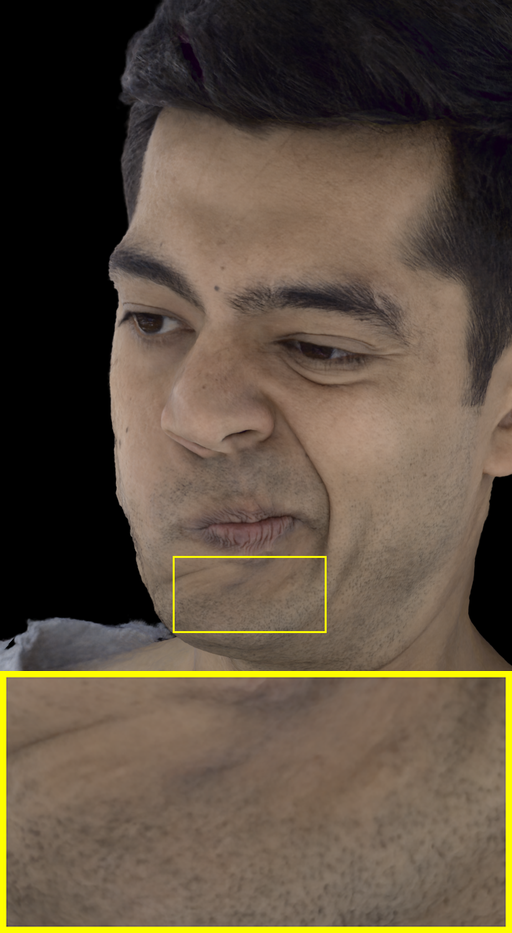}  &
    \includegraphics[width=\width]{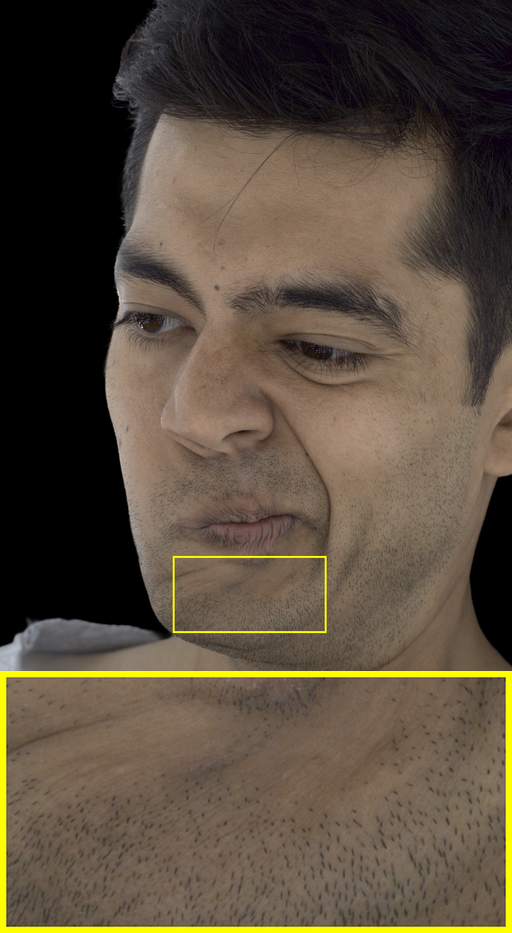}  \\
Input & SparseNeRF & SPARF & Diner$^*$ & FreeNeRF & Preface & \textbf{Ours} & GT
\end{tabular}
\end{center}
\caption{\label{fig:main_comparison}We compare our method with previous work on the Multiface dataset~\cite{wuu2022multiface}. For each expression, we show the novel-view results with three input views. The symbol $^*$ indicates results at a lower resolution. Both visually and quantitatively, our method significantly outperforms SparseNeRF \cite{wang2023sparsenerf}, SPARF~\cite{sparf2023}, Diner~\cite{diner}, and FreeNeRF~\cite{Yang2023FreeNeRF}. Compared to the state-of-the-art Preface~\cite{buhler2023preface}, we see a clear improvement in areas like eye, chin, and face contours, as shown in the zoom-in. For example, our method better reconstructs the teeth on row 2. We also compare with Preface on in-the-wild captures, please see Fig. \ref{fig:comparison_itw}.
}
\end{figure*}

Performance on casual in-the-wild captures is inherently difficult to evaluate quantitatively due to the lack of proper validation views. Often, the captured subject can hardly remain completely still. Therefore, we quantitatively evaluate on the Multiface \cite{wuu2022multiface} studio dataset on nine scenes by randomly selecting three subjects with three expressions per subject. For each test subject, we use one frontal and two side views as input for training, as shown in Fig.~\ref{fig:main_comparison}. We remove the background by multiplying a foreground alpha matte estimated by \cite{pandey2021total}.
We hold out from 26 to 29 validation images per subject, where the camera viewing direction is located in the frontal hemisphere (see details in the supplementary material). As evaluation metrics, we measure photo-metric distance and image similarity via PSNR, SSIM, and LPIPS~\cite{zhang2018perceptual}, as summarized in Tbl.~\ref{tbl:main_comparison}. Note that photometric reconstruction metrics like PSNR and SSIM and perceptual metrics like LPIPS are at odds with each other \cite{blau2018perception}. We handle this tradeoff by optimizing perceptual quality in the warm-up (Eq. \ref{eq:inversion}) and reconstruction quality in the fine-tuning (Eq. \ref{eq:modelfittingloss}).
 
As shown in Tbl.~\ref{tbl:main_comparison}, our new method provides better modeling fidelity across all three metrics over the set of validation views. In particular, compared to the state-of-the-art Preface \cite{buhler2023preface}, we achieve the following improvement on the computed metrics: 6\% PSNR, 2.4\% SSIM, and 5\% LPIPS. Visually, Fig.~\ref{fig:main_comparison} shows that the novel views generated by our fine-tuned model more closely resemble the facial shape and appearance, including eye and mouth details, of the example (ground-truth) validation view.

\subsection{Qualitative Results in the Wild}
\label{sec:qualitative}

To demonstrate the robustness of our method in the wild, we captured subjects with a handheld DSLR camera (Canon EOS 4000D) and mobile phones (Pixel 7 and Pixel 8 Pro). We capture between one and three images per subject in various outdoor and indoor environments, including very challenging lighting conditions. These diverse in-the-wild results are shown in Fig.~\ref{fig:teaser}, Fig. \ref{fig:comparison_itw}, Fig.~\ref{fig:inthewild}, and in the supplementary video and HTML page. The inlays in Fig.~\ref{fig:teaser} show the high level of modeled detail on the lips and on the individual hair strands and eyelashes. Fig. \ref{fig:comparison_itw} compares with the best performing related work \cite{buhler2023preface}, which struggles with the mouth region. Of particular note is also the ``tongue-out'' expression in the third row of Fig.~\ref{fig:inthewild}. Note that our synthetic training data contains tongues but the tongues never stick out of the mouth. 

We also consider the more challenging single input image scenario, see Fig.~\ref{fig:singleimage}. Our method achieves high-fidelity synthesis of the frontal face even for stylized and painted faces (right) but quickly degrades for side views. This behavior is expected, as our model is only trained on low-resolution synthetic images and has never seen high-resolution views from the side of a face. Please see the supp. HTML page for more examples.

Additional qualitative comparisons to previous work are shown in Fig.~\ref{fig:main_comparison}. Our method outperforms the other baseline methods in two main ways. First, our approach captures fine identity-specific details like teeth outlines (middle row) and stubble (bottom row), while previous methods degenerate into blur or noise. Second, the overall face shape is more accurate, as evidenced by the eyeball in the top row and the cheek silhouette in the middle row. Our synthetic face prior encourages the reconstruction to remain faithful to its learned understanding of faces, \eg that corneas should bulge outwards, not be flat. These results and the metrics in Tbl.~\ref{tbl:main_comparison} demonstrate that our new model and fine-tuning method outperforms previous work both qualitatively and quantitatively.

\label{sec:comparisons}

\subsection{Ablation Study}
\label{sec:ablation}
We conduct extensive ablation studies of the prior model. Table ~\ref{tbl:ablation_dataset} lists metrics after fine-tuning to three inputs at 2K resolution. Please see the supp. PDF for more ablations and the HTML page for visuals.

\paragraph{Dataset Size}
We compare prior models without any pre-training (a.i) with models pre-trained on a single subject (a.ii), on 15 subjects (a.iii), and on 1,500 subjects (a.iv). Note that each subject is rendered with 13 expressions and 30 views per expression (as described in Sec. \ref{sec:prior}). Hence, the model trained on 1 subject (a.ii) sees $1 \cdot 13 \cdot 30 = 390$ images in total. Without pre-training (a.i), the reconstruction completely fails. Pre-training on a single subject (a.ii) leads to a noisy face geometry (see the surface normals in the supp. HTML page). The performance improves when more subjects are added (a.iii and a.iv). We conclude that pre-training is necessary and more data improves the reconstruction quality.

\paragraph{Number of Pre-training Steps}
We ablate the performance when pre-training for a fewer number of steps. We observe that pre-training for one day only (105K steps, b.i) already achieves very high photo-metric reconstruction quality (PSNR and SSIM). This shorter training duration offers a more accessible alternative while still delivering high-quality outcomes. While short pre-training already yields good results, longer training is required for the best perceptual quality (LPIPS) (b.ii and b.iii).

\paragraph{Synthetic Data Quality}
We investigate the effects of synthetic data quality in terms of appearance and texture (c.i - c.iv) and geometric diversity (c.v - c.vii). We find that the appearance and texture of the prior model have very little impact on the fine-tuning result but a higher geometry diversity achieves the best results.
Prior models trained on gray-scale and low-quality renderings (c.i and c.ii) perform similarly as training in diverse environments (c.iii) \cite{gardner2017laval,hold2019deep}. Similarly, excluding details like makeup (c.iv) does not deteriorate the performance. However, the diversity in terms of geometry is important for the prior model. When removing hair (c.vi) and other accessories like beards, glasses, and clothing (c.v) from the prior model, the reconstruction still yields a valid face but it may contain some artifacts in non-surface regions like hair. This is notable as a drop in PSNR from 26.54 with all accessories (c.vii) to 26.00 without any accessories (c.v) (please see the supp. mat. for a list of the number of accessories). In summary, we find that the prior acts as a geometric regularization. Including all accessories yields a geometrically diverse prior model with the best results. Rendering even more assets is likely to improve the results for accessories like glasses, earrings, and clothing even further.

\paragraph{Including Real Data}
We study the synthetic vs. real domain gap by pre-training the prior model on real multi-view images (d.i), a mixed dataset with real and synthetic images (d.ii), and synthetic images alone (d.iii). Priors trained on a real and a mixed dataset perform well but synthetic data alone performs best on all metrics.
This behavior might seem non-intuitive. While it is very difficult to precisely determine why synthetic data outperforms real data, we are not the first to find that synthetic data can outperform real data  \cite{wood2021fake,wood20223d,yeh2022learning,sun2021nelf,trevithick2023}. In our experience, real data capture and processing is imperfect compared with synthetic data. Even under controlled conditions, there may be issues like motion blur and imperfect foreground matting.

\subsection{Limitations}
\begin{figure*}[ht]
\begin{center}
\small
\setlength{\tabcolsep}{2pt}

\newcommand{\height}{18mm}
\begin{tabular}{cccc | cccc}
  \includegraphics[height=\height]{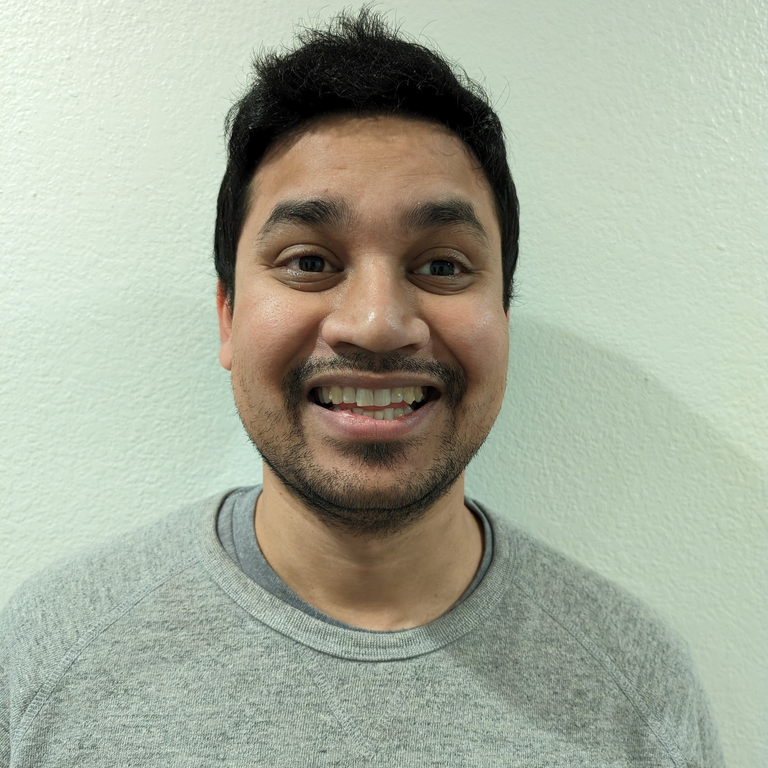} &
      \includegraphics[height=\height]{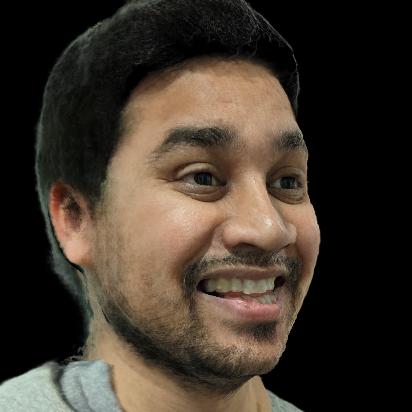} &
      \includegraphics[height=\height]{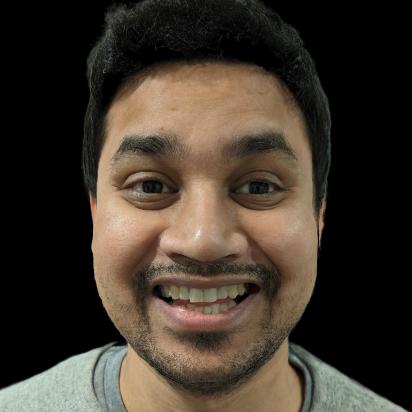} &
      \includegraphics[height=\height]{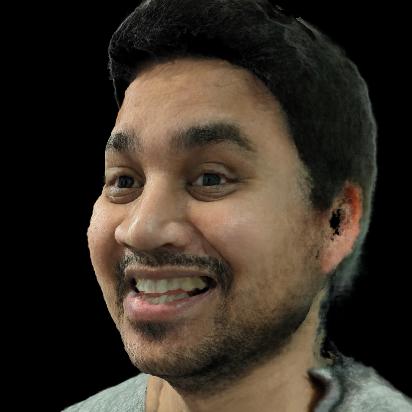}
  & \includegraphics[height=\height]{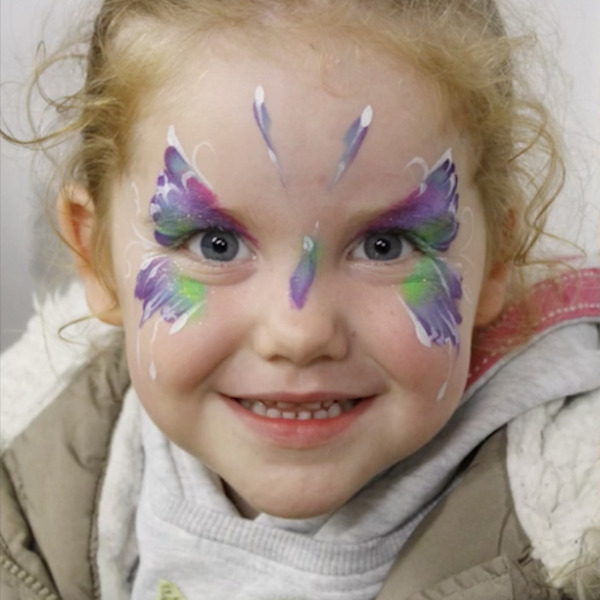} &
        \includegraphics[height=\height]{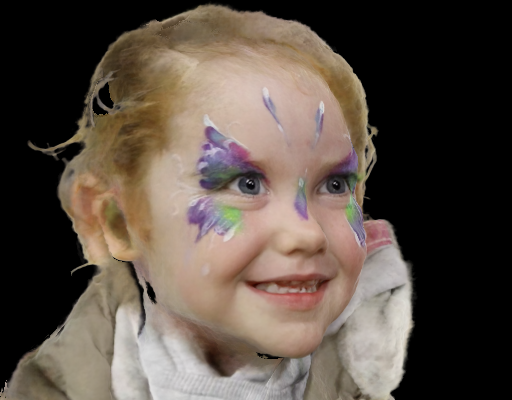} & 
      \includegraphics[height=\height]{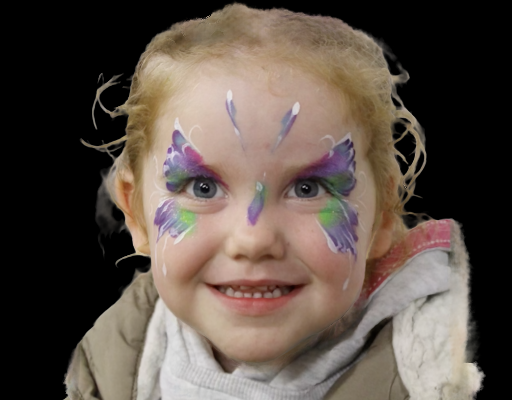} &
      \includegraphics[height=\height]{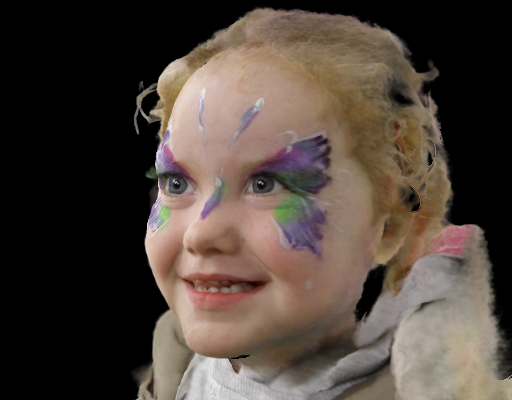}  \\
  \includegraphics[height=\height]{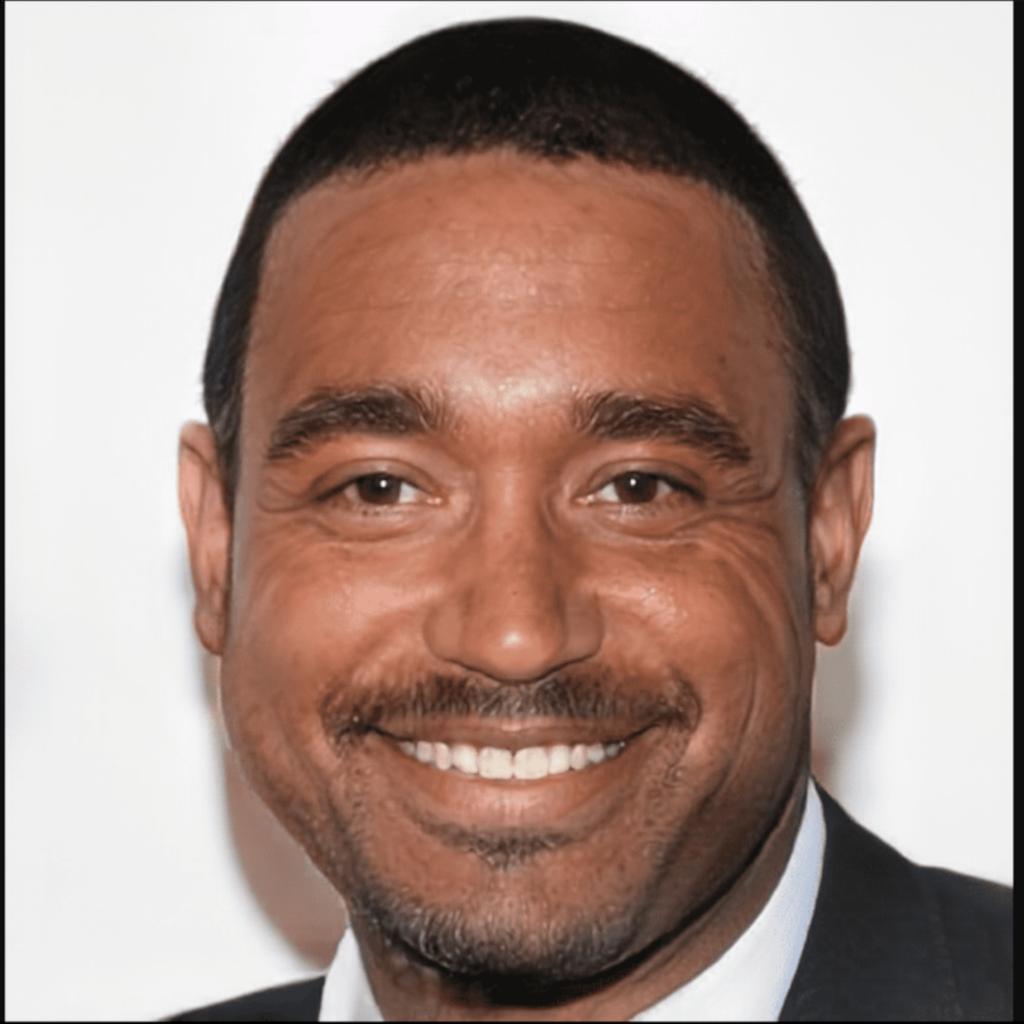} &
  \includegraphics[height=\height]{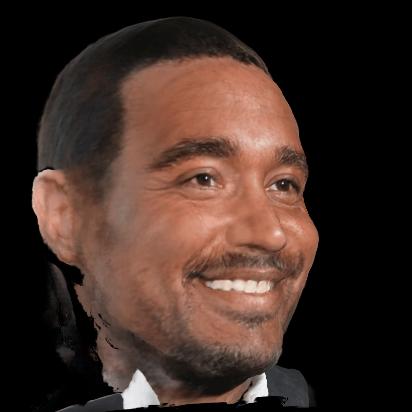} &
  \includegraphics[height=\height]{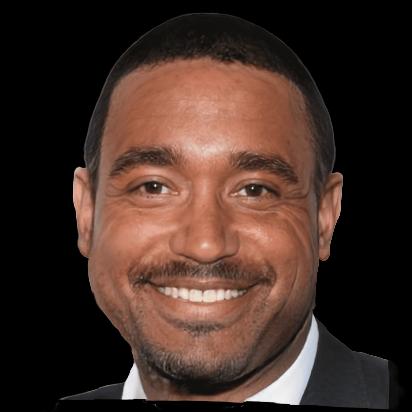} &
  \includegraphics[height=\height]{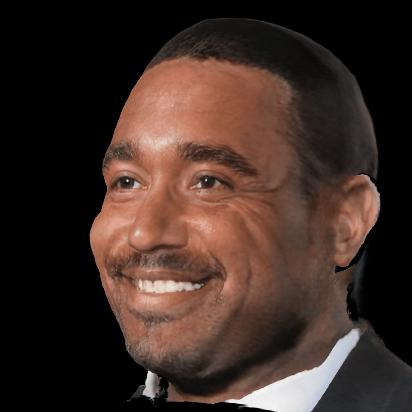} 
& \includegraphics[height=\height]{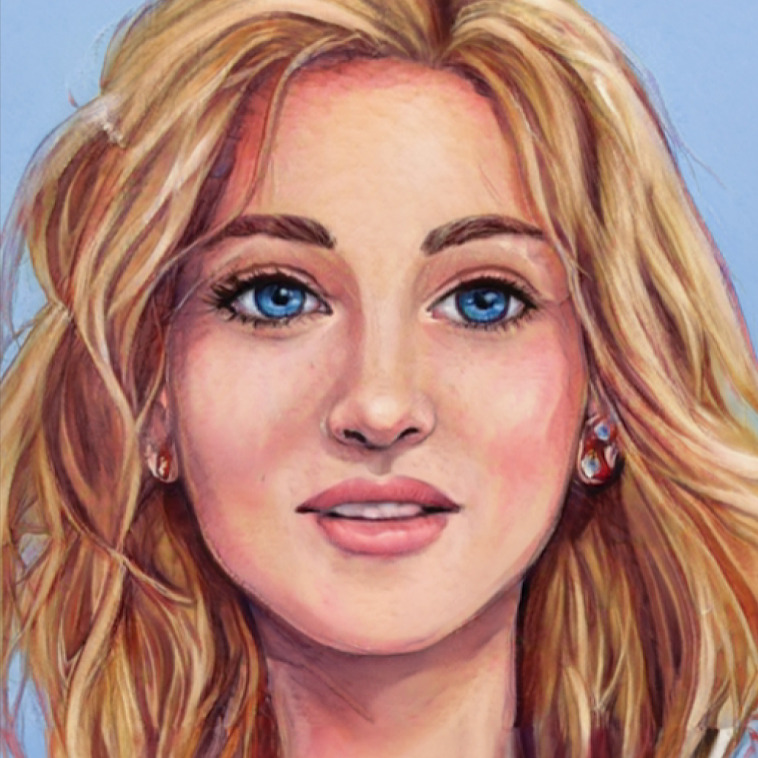} &
        \includegraphics[height=\height]{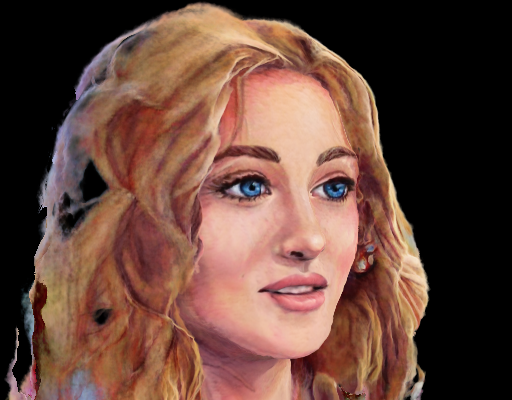}  &
      \includegraphics[height=\height]{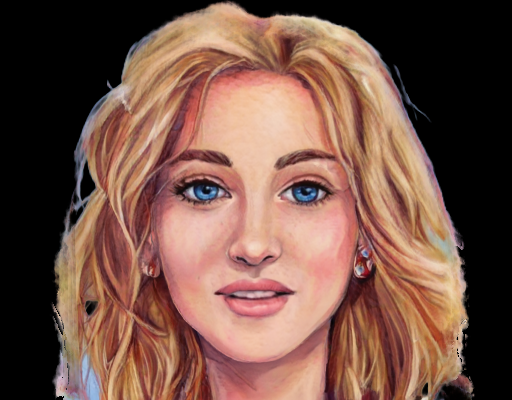} &
      \includegraphics[height=\height]{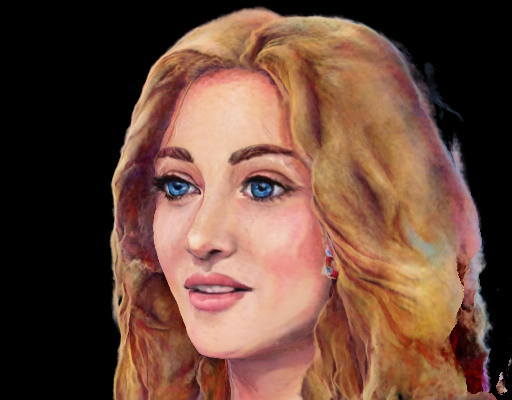}
      \\
  Input & \multicolumn{3}{c}{Novel Views} & Input & \multicolumn{3}{c}{Novel Views} \\
\end{tabular}
\end{center}
\caption{Single Image Results. We push the limits of our method by reconstructing the face with only a single input image. The top left image is a smartphone image, the bottom left an Image from Wang et al. (2023) \cite{wang2023rodin}, and the right two examples are stylized images from \cite{trevithick2023}.
\label{fig:singleimage}}
\end{figure*}

While our method can reconstruct 3D faces from even just one view, we notice that the quality degrades at side views in this extremely challenging case (see Fig.~\ref{fig:singleimage}) due to the lack of observation. Accessories like glasses frames and earrings may not be reconstructed perfectly and for extreme expressions, the mouth interior might not be fully consistent across all viewpoints
(see the supp. HTML page). Furthermore, our method assumes that faces are non-occluded in the input images. This can cause the camera estimation during the 3DMM fitting (Sec. \ref{sec:3dmmfitting}) to fail. These limitations could be potentially addressed by leveraging large generative models to hallucinate unobserved regions, which is an interesting future direction to explore. 
Another direction of future work could explore more efficient backbones to increase pre-training and fine-tuning efficiency, and potentially enable facial animation.

\section {Conclusion}
We present a method for high-fidelity expressive face modeling from as few as three input views captured in the wild. This challenging goal is achieved by leveraging a volumetric face prior and fine-tuning the prior model to sparse observations. Our key insight is that a prior model trained on synthetic data alone can generalize to diverse real-world identities and expressions, bypassing the expensive process of capturing large-scale real-world 3D facial appearances. We experimentally demonstrate that our method can robustly reconstruct expressive faces from sparse or even single-view images with unprecedented fine-scale idiosyncratic details, and achieve superior quality compared to previous state-of-the-art methods for few-shot reconstruction and novel view synthesis.

\bibliographystyle{ACM-Reference-Format}
\bibliography{sample-bibliography}


\begin{thebibliography}{88}


\ifx \showCODEN    \undefined \def \showCODEN     #1{\unskip}     \fi
\ifx \showDOI      \undefined \def \showDOI       #1{#1}\fi
\ifx \showISBNx    \undefined \def \showISBNx     #1{\unskip}     \fi
\ifx \showISBNxiii \undefined \def \showISBNxiii  #1{\unskip}     \fi
\ifx \showISSN     \undefined \def \showISSN      #1{\unskip}     \fi
\ifx \showLCCN     \undefined \def \showLCCN      #1{\unskip}     \fi
\ifx \shownote     \undefined \def \shownote      #1{#1}          \fi
\ifx \showarticletitle \undefined \def \showarticletitle #1{#1}   \fi
\ifx \showURL      \undefined \def \showURL       {\relax}        \fi
\providecommand\bibfield[2]{#2}
\providecommand\bibinfo[2]{#2}
\providecommand\natexlab[1]{#1}
\providecommand\showeprint[2][]{arXiv:#2}

\bibitem[Barron et~al\mbox{.}(2021)]%
        {barron2021mip}
\bibfield{author}{\bibinfo{person}{Jonathan~T Barron}, \bibinfo{person}{Ben Mildenhall}, \bibinfo{person}{Matthew Tancik}, \bibinfo{person}{Peter Hedman}, \bibinfo{person}{Ricardo Martin-Brualla}, {and} \bibinfo{person}{Pratul~P Srinivasan}.} \bibinfo{year}{2021}\natexlab{}.
\newblock \showarticletitle{Mip-nerf: A multiscale representation for anti-aliasing neural radiance fields}. In \bibinfo{booktitle}{\emph{Proceedings of the IEEE/CVF International Conference on Computer Vision}}. \bibinfo{pages}{5855--5864}.
\newblock


\bibitem[Barron et~al\mbox{.}(2022)]%
        {mipnerf360}
\bibfield{author}{\bibinfo{person}{Jonathan~T Barron}, \bibinfo{person}{Ben Mildenhall}, \bibinfo{person}{Dor Verbin}, \bibinfo{person}{Pratul~P Srinivasan}, {and} \bibinfo{person}{Peter Hedman}.} \bibinfo{year}{2022}\natexlab{}.
\newblock \showarticletitle{Mip-nerf 360: Unbounded anti-aliased neural radiance fields}. In \bibinfo{booktitle}{\emph{Proceedings of the IEEE/CVF Conference on Computer Vision and Pattern Recognition}}. \bibinfo{pages}{5470--5479}.
\newblock


\bibitem[Bednarik et~al\mbox{.}(2024)]%
        {10.1111:cgf.15038}
\bibfield{author}{\bibinfo{person}{Jan Bednarik}, \bibinfo{person}{Erroll Wood}, \bibinfo{person}{Vassilis Choutas}, \bibinfo{person}{Timo Bolkart}, \bibinfo{person}{Daoye Wang}, \bibinfo{person}{Chenglei Wu}, {and} \bibinfo{person}{Thabo Beeler}.} \bibinfo{year}{2024}\natexlab{}.
\newblock \showarticletitle{{Learning to Stabilize Faces}}.
\newblock \bibinfo{journal}{\emph{Computer Graphics Forum}} (\bibinfo{year}{2024}).
\newblock
\showISSN{1467-8659}
\urldef\tempurl%
\url{https://doi.org/10.1111/cgf.15038}
\showDOI{\tempurl}


\bibitem[Blanz and Vetter(1999)]%
        {blanz1999morphable}
\bibfield{author}{\bibinfo{person}{Volker Blanz} {and} \bibinfo{person}{Thomas Vetter}.} \bibinfo{year}{1999}\natexlab{}.
\newblock \showarticletitle{A morphable model for the synthesis of 3D faces}. In \bibinfo{booktitle}{\emph{Proceedings of the 26th annual conference on Computer graphics and interactive techniques}}. \bibinfo{pages}{187--194}.
\newblock


\bibitem[Blau and Michaeli(2018)]%
        {blau2018perception}
\bibfield{author}{\bibinfo{person}{Yochai Blau} {and} \bibinfo{person}{Tomer Michaeli}.} \bibinfo{year}{2018}\natexlab{}.
\newblock \showarticletitle{The perception-distortion tradeoff}. In \bibinfo{booktitle}{\emph{Proceedings of the IEEE conference on computer vision and pattern recognition}}. \bibinfo{pages}{6228--6237}.
\newblock


\bibitem[Bojanowski et~al\mbox{.}(2018)]%
        {bojanowski2018optimizing}
\bibfield{author}{\bibinfo{person}{Piotr Bojanowski}, \bibinfo{person}{Armand Joulin}, \bibinfo{person}{David Lopez-Paz}, {and} \bibinfo{person}{Arthur Szlam}.} \bibinfo{year}{2018}\natexlab{}.
\newblock \showarticletitle{Optimizing the latent space of generative networks}. In \bibinfo{booktitle}{\emph{Proceedings of the 35th International Conference on Machine Learning}}. \bibinfo{pages}{2640--3498}.
\newblock


\bibitem[Buehler et~al\mbox{.}(2021)]%
        {varitex}
\bibfield{author}{\bibinfo{person}{Marcel~C. Buehler}, \bibinfo{person}{Abhimitra Meka}, \bibinfo{person}{Gengyan Li}, \bibinfo{person}{Thabo Beeler}, {and} \bibinfo{person}{Otmar Hilliges}.} \bibinfo{year}{2021}\natexlab{}.
\newblock \showarticletitle{VariTex: Variational Neural Face Textures}. In \bibinfo{booktitle}{\emph{Proceedings of the IEEE/CVF International Conference on Computer Vision}}.
\newblock


\bibitem[Buehler et~al\mbox{.}(2023)]%
        {buhler2023preface}
\bibfield{author}{\bibinfo{person}{Marcel~C Buehler}, \bibinfo{person}{Kripasindhu Sarkar}, \bibinfo{person}{Tanmay Shah}, \bibinfo{person}{Gengyan Li}, \bibinfo{person}{Daoye Wang}, \bibinfo{person}{Leonhard Helminger}, \bibinfo{person}{Sergio Orts-Escolano}, \bibinfo{person}{Dmitry Lagun}, \bibinfo{person}{Otmar Hilliges}, \bibinfo{person}{Thabo Beeler}, {et~al\mbox{.}}} \bibinfo{year}{2023}\natexlab{}.
\newblock \showarticletitle{Preface: A Data-driven Volumetric Prior for Few-shot Ultra High-resolution Face Synthesis}. In \bibinfo{booktitle}{\emph{Proceedings of the IEEE/CVF International Conference on Computer Vision}}. \bibinfo{pages}{3402--3413}.
\newblock


\bibitem[Cao et~al\mbox{.}(2022)]%
        {cao2022authentic}
\bibfield{author}{\bibinfo{person}{Chen Cao}, \bibinfo{person}{Tomas Simon}, \bibinfo{person}{Jin~Kyu Kim}, \bibinfo{person}{Gabe Schwartz}, \bibinfo{person}{Michael Zollhoefer}, \bibinfo{person}{Shun-Suke Saito}, \bibinfo{person}{Stephen Lombardi}, \bibinfo{person}{Shih-En Wei}, \bibinfo{person}{Danielle Belko}, \bibinfo{person}{Shoou-I Yu}, {et~al\mbox{.}}} \bibinfo{year}{2022}\natexlab{}.
\newblock \showarticletitle{Authentic volumetric avatars from a phone scan}.
\newblock \bibinfo{journal}{\emph{ACM Transactions on Graphics (TOG)}} \bibinfo{volume}{41}, \bibinfo{number}{4} (\bibinfo{year}{2022}), \bibinfo{pages}{1--19}.
\newblock


\bibitem[Chan et~al\mbox{.}(2022)]%
        {eg3d}
\bibfield{author}{\bibinfo{person}{Eric~R Chan}, \bibinfo{person}{Connor~Z Lin}, \bibinfo{person}{Matthew~A Chan}, \bibinfo{person}{Koki Nagano}, \bibinfo{person}{Boxiao Pan}, \bibinfo{person}{Shalini De~Mello}, \bibinfo{person}{Orazio Gallo}, \bibinfo{person}{Leonidas~J Guibas}, \bibinfo{person}{Jonathan Tremblay}, \bibinfo{person}{Sameh Khamis}, {et~al\mbox{.}}} \bibinfo{year}{2022}\natexlab{}.
\newblock \showarticletitle{Efficient geometry-aware 3D generative adversarial networks}. In \bibinfo{booktitle}{\emph{Proceedings of the IEEE/CVF Conference on Computer Vision and Pattern Recognition}}. \bibinfo{pages}{16123--16133}.
\newblock


\bibitem[Chan et~al\mbox{.}(2021)]%
        {pigan}
\bibfield{author}{\bibinfo{person}{Eric~R Chan}, \bibinfo{person}{Marco Monteiro}, \bibinfo{person}{Petr Kellnhofer}, \bibinfo{person}{Jiajun Wu}, {and} \bibinfo{person}{Gordon Wetzstein}.} \bibinfo{year}{2021}\natexlab{}.
\newblock \showarticletitle{pi-gan: Periodic implicit generative adversarial networks for 3d-aware image synthesis}. In \bibinfo{booktitle}{\emph{Proceedings of the IEEE/CVF Conference on Computer Vision and Pattern Recognition}}. \bibinfo{pages}{5799--5809}.
\newblock


\bibitem[Chen et~al\mbox{.}(2021)]%
        {chen2021mvsnerf}
\bibfield{author}{\bibinfo{person}{Anpei Chen}, \bibinfo{person}{Zexiang Xu}, \bibinfo{person}{Fuqiang Zhao}, \bibinfo{person}{Xiaoshuai Zhang}, \bibinfo{person}{Fanbo Xiang}, \bibinfo{person}{Jingyi Yu}, {and} \bibinfo{person}{Hao Su}.} \bibinfo{year}{2021}\natexlab{}.
\newblock \showarticletitle{Mvsnerf: Fast generalizable radiance field reconstruction from multi-view stereo}. In \bibinfo{booktitle}{\emph{Proceedings of the IEEE/CVF International Conference on Computer Vision}}. \bibinfo{pages}{14124--14133}.
\newblock


\bibitem[Choi et~al\mbox{.}(2020)]%
        {choi2020stargan}
\bibfield{author}{\bibinfo{person}{Yunjey Choi}, \bibinfo{person}{Youngjung Uh}, \bibinfo{person}{Jaejun Yoo}, {and} \bibinfo{person}{Jung-Woo Ha}.} \bibinfo{year}{2020}\natexlab{}.
\newblock \showarticletitle{Stargan v2: Diverse image synthesis for multiple domains}. In \bibinfo{booktitle}{\emph{Proceedings of the IEEE/CVF Conference on Computer Vision and Pattern Recognition}}. \bibinfo{pages}{8188--8197}.
\newblock


\bibitem[Deng et~al\mbox{.}(2022a)]%
        {gram}
\bibfield{author}{\bibinfo{person}{Yu Deng}, \bibinfo{person}{Jiaolong Yang}, \bibinfo{person}{Jianfeng Xiang}, {and} \bibinfo{person}{Xin Tong}.} \bibinfo{year}{2022}\natexlab{a}.
\newblock \showarticletitle{GRAM: Generative Radiance Manifolds for 3D-Aware Image Generation}. In \bibinfo{booktitle}{\emph{IEEE/CVF Conference on Computer Vision and Pattern Recognition}}.
\newblock


\bibitem[Deng et~al\mbox{.}(2022b)]%
        {deng2022gram}
\bibfield{author}{\bibinfo{person}{Yu Deng}, \bibinfo{person}{Jiaolong Yang}, \bibinfo{person}{Jianfeng Xiang}, {and} \bibinfo{person}{Xin Tong}.} \bibinfo{year}{2022}\natexlab{b}.
\newblock \showarticletitle{GRAM: Generative Radiance Manifolds for 3D-Aware Image Generation}. In \bibinfo{booktitle}{\emph{IEEE/CVF Conference on Computer Vision and Pattern Recognition}}.
\newblock


\bibitem[Duan et~al\mbox{.}(2023)]%
        {duan2023bakedavatar}
\bibfield{author}{\bibinfo{person}{Hao-Bin Duan}, \bibinfo{person}{Miao Wang}, \bibinfo{person}{Jin-Chuan Shi}, \bibinfo{person}{Xu-Chuan Chen}, {and} \bibinfo{person}{Yan-Pei Cao}.} \bibinfo{year}{2023}\natexlab{}.
\newblock \showarticletitle{Bakedavatar: Baking neural fields for real-time head avatar synthesis}.
\newblock \bibinfo{journal}{\emph{ACM Transactions on Graphics (TOG)}} \bibinfo{volume}{42}, \bibinfo{number}{6} (\bibinfo{year}{2023}), \bibinfo{pages}{1--17}.
\newblock


\bibitem[Finn et~al\mbox{.}(2017)]%
        {finn2017modelmaml}
\bibfield{author}{\bibinfo{person}{Chelsea Finn}, \bibinfo{person}{Pieter Abbeel}, {and} \bibinfo{person}{Sergey Levine}.} \bibinfo{year}{2017}\natexlab{}.
\newblock \showarticletitle{Model-agnostic meta-learning for fast adaptation of deep networks}. In \bibinfo{booktitle}{\emph{International conference on machine learning}}. PMLR, \bibinfo{pages}{1126--1135}.
\newblock


\bibitem[Gao et~al\mbox{.}(2020)]%
        {gao2020portrait}
\bibfield{author}{\bibinfo{person}{Chen Gao}, \bibinfo{person}{Yichang Shih}, \bibinfo{person}{Wei-Sheng Lai}, \bibinfo{person}{Chia-Kai Liang}, {and} \bibinfo{person}{Jia-Bin Huang}.} \bibinfo{year}{2020}\natexlab{}.
\newblock \showarticletitle{Portrait neural radiance fields from a single image}.
\newblock \bibinfo{journal}{\emph{arXiv preprint arXiv:2012.05903}} (\bibinfo{year}{2020}).
\newblock


\bibitem[Gardner et~al\mbox{.}(2017)]%
        {gardner2017laval}
\bibfield{author}{\bibinfo{person}{Marc-Andr\'{e} Gardner}, \bibinfo{person}{Kalyan Sunkavalli}, \bibinfo{person}{Ersin Yumer}, \bibinfo{person}{Xiaohui Shen}, \bibinfo{person}{Emiliano Gambaretto}, \bibinfo{person}{Christian Gagn\'{e}}, {and} \bibinfo{person}{Jean-Fran\c{c}ois Lalonde}.} \bibinfo{year}{2017}\natexlab{}.
\newblock \showarticletitle{Learning to predict indoor illumination from a single image}.
\newblock \bibinfo{journal}{\emph{ACM Trans. Graph.}} \bibinfo{volume}{36}, \bibinfo{number}{6}, Article \bibinfo{articleno}{176} (\bibinfo{date}{nov} \bibinfo{year}{2017}), \bibinfo{numpages}{14}~pages.
\newblock
\showISSN{0730-0301}
\urldef\tempurl%
\url{https://doi.org/10.1145/3130800.3130891}
\showDOI{\tempurl}


\bibitem[Gu et~al\mbox{.}(2021)]%
        {gu2021stylenerf}
\bibfield{author}{\bibinfo{person}{Jiatao Gu}, \bibinfo{person}{Lingjie Liu}, \bibinfo{person}{Peng Wang}, {and} \bibinfo{person}{Christian Theobalt}.} \bibinfo{year}{2021}\natexlab{}.
\newblock \showarticletitle{Stylenerf: A style-based 3d-aware generator for high-resolution image synthesis}.
\newblock \bibinfo{journal}{\emph{arXiv preprint arXiv:2110.08985}} (\bibinfo{year}{2021}).
\newblock


\bibitem[Guangcong et~al\mbox{.}(2023)]%
        {wang2023sparsenerf}
\bibfield{author}{\bibinfo{person}{Guangcong}, \bibinfo{person}{Zhaoxi Chen}, \bibinfo{person}{Chen~Change Loy}, {and} \bibinfo{person}{Ziwei Liu}.} \bibinfo{year}{2023}\natexlab{}.
\newblock \showarticletitle{SparseNeRF: Distilling Depth Ranking for Few-shot Novel View Synthesis}.
\newblock \bibinfo{journal}{\emph{Technical Report}} (\bibinfo{year}{2023}).
\newblock


\bibitem[He et~al\mbox{.}(2015)]%
        {he2015delving}
\bibfield{author}{\bibinfo{person}{Kaiming He}, \bibinfo{person}{Xiangyu Zhang}, \bibinfo{person}{Shaoqing Ren}, {and} \bibinfo{person}{Jian Sun}.} \bibinfo{year}{2015}\natexlab{}.
\newblock \showarticletitle{Delving deep into rectifiers: Surpassing human-level performance on imagenet classification}. In \bibinfo{booktitle}{\emph{Proceedings of the IEEE international conference on computer vision}}. \bibinfo{pages}{1026--1034}.
\newblock


\bibitem[Hold-Geoffroy et~al\mbox{.}(2019)]%
        {hold2019deep}
\bibfield{author}{\bibinfo{person}{Yannick Hold-Geoffroy}, \bibinfo{person}{Akshaya Athawale}, {and} \bibinfo{person}{Jean-Fran{\c{c}}ois Lalonde}.} \bibinfo{year}{2019}\natexlab{}.
\newblock \showarticletitle{Deep sky modeling for single image outdoor lighting estimation}. In \bibinfo{booktitle}{\emph{Proceedings of the IEEE/CVF conference on computer vision and pattern recognition}}. \bibinfo{pages}{6927--6935}.
\newblock


\bibitem[Hong et~al\mbox{.}(2022)]%
        {headnerf}
\bibfield{author}{\bibinfo{person}{Yang Hong}, \bibinfo{person}{Bo Peng}, \bibinfo{person}{Haiyao Xiao}, \bibinfo{person}{Ligang Liu}, {and} \bibinfo{person}{Juyong Zhang}.} \bibinfo{year}{2022}\natexlab{}.
\newblock \showarticletitle{HeadNeRF: A Real-time NeRF-based Parametric Head Model}. In \bibinfo{booktitle}{\emph{{IEEE/CVF} Conference on Computer Vision and Pattern Recognition (CVPR)}}.
\newblock


\bibitem[Jain et~al\mbox{.}(2021)]%
        {dietnerf}
\bibfield{author}{\bibinfo{person}{Ajay Jain}, \bibinfo{person}{Matthew Tancik}, {and} \bibinfo{person}{Pieter Abbeel}.} \bibinfo{year}{2021}\natexlab{}.
\newblock \showarticletitle{Putting NeRF on a Diet: Semantically Consistent Few-Shot View Synthesis}. In \bibinfo{booktitle}{\emph{Proceedings of the IEEE/CVF International Conference on Computer Vision (ICCV)}}. \bibinfo{pages}{5885--5894}.
\newblock


\bibitem[Karras et~al\mbox{.}(2019)]%
        {karras2019style}
\bibfield{author}{\bibinfo{person}{Tero Karras}, \bibinfo{person}{Samuli Laine}, {and} \bibinfo{person}{Timo Aila}.} \bibinfo{year}{2019}\natexlab{}.
\newblock \showarticletitle{A style-based generator architecture for generative adversarial networks}. In \bibinfo{booktitle}{\emph{Proceedings of the IEEE conference on computer vision and pattern recognition}}. \bibinfo{pages}{4401--4410}.
\newblock


\bibitem[Kingma and Ba(2015)]%
        {KingmB2015}
\bibfield{author}{\bibinfo{person}{Diederik~P. Kingma} {and} \bibinfo{person}{Jimmy Ba}.} \bibinfo{year}{2015}\natexlab{}.
\newblock \showarticletitle{Adam: A Method for Stochastic Optimization}. In \bibinfo{booktitle}{\emph{ICLR}}.
\newblock


\bibitem[Kirschstein et~al\mbox{.}(2023)]%
        {kirschstein2023nersemble}
\bibfield{author}{\bibinfo{person}{Tobias Kirschstein}, \bibinfo{person}{Shenhan Qian}, \bibinfo{person}{Simon Giebenhain}, \bibinfo{person}{Tim Walter}, {and} \bibinfo{person}{Matthias Nie\ss{}ner}.} \bibinfo{year}{2023}\natexlab{}.
\newblock \showarticletitle{NeRSemble: Multi-View Radiance Field Reconstruction of Human Heads}.
\newblock \bibinfo{journal}{\emph{ACM Trans. Graph.}} \bibinfo{volume}{42}, \bibinfo{number}{4}, Article \bibinfo{articleno}{161} (\bibinfo{date}{jul} \bibinfo{year}{2023}), \bibinfo{numpages}{14}~pages.
\newblock
\showISSN{0730-0301}
\urldef\tempurl%
\url{https://doi.org/10.1145/3592455}
\showDOI{\tempurl}


\bibitem[Kret(2015)]%
        {eyeexpressions}
\bibfield{author}{\bibinfo{person}{Mariska~E. Kret}.} \bibinfo{year}{2015}\natexlab{}.
\newblock \showarticletitle{Emotional expressions beyond facial muscle actions. A call for studying autonomic signals and their impact on social perception}.
\newblock \bibinfo{journal}{\emph{Frontiers in Psychology}}  \bibinfo{volume}{6} (\bibinfo{year}{2015}).
\newblock
\showISSN{1664-1078}
\urldef\tempurl%
\url{https://doi.org/10.3389/fpsyg.2015.00711}
\showDOI{\tempurl}


\bibitem[Kundu et~al\mbox{.}(2022)]%
        {kundu2022panoptic}
\bibfield{author}{\bibinfo{person}{Abhijit Kundu}, \bibinfo{person}{Kyle Genova}, \bibinfo{person}{Xiaoqi Yin}, \bibinfo{person}{Alireza Fathi}, \bibinfo{person}{Caroline Pantofaru}, \bibinfo{person}{Leonidas~J Guibas}, \bibinfo{person}{Andrea Tagliasacchi}, \bibinfo{person}{Frank Dellaert}, {and} \bibinfo{person}{Thomas Funkhouser}.} \bibinfo{year}{2022}\natexlab{}.
\newblock \showarticletitle{Panoptic neural fields: A semantic object-aware neural scene representation}. In \bibinfo{booktitle}{\emph{Proceedings of the IEEE/CVF Conference on Computer Vision and Pattern Recognition}}. \bibinfo{pages}{12871--12881}.
\newblock


\bibitem[Lee and Anderson(2017)]%
        {lee2017reading}
\bibfield{author}{\bibinfo{person}{Daniel~H Lee} {and} \bibinfo{person}{Adam~K Anderson}.} \bibinfo{year}{2017}\natexlab{}.
\newblock \showarticletitle{Reading what the mind thinks from how the eye sees}.
\newblock \bibinfo{journal}{\emph{Psychological science}} \bibinfo{volume}{28}, \bibinfo{number}{4} (\bibinfo{year}{2017}), \bibinfo{pages}{494--503}.
\newblock


\bibitem[Li et~al\mbox{.}(2024)]%
        {li2024shellnerf}
\bibfield{author}{\bibinfo{person}{Gengyan Li}, \bibinfo{person}{Kripasindhu Sarkar}, \bibinfo{person}{Abhimitra Meka}, \bibinfo{person}{Marcel Buehler}, \bibinfo{person}{Franziska Mueller}, \bibinfo{person}{Paulo Gotardo}, \bibinfo{person}{Otmar Hilliges}, {and} \bibinfo{person}{Thabo Beeler}.} \bibinfo{year}{2024}\natexlab{}.
\newblock \showarticletitle{ShellNeRF: Learning a Controllable High-resolution Model of the Eye and Periocular Region}. In \bibinfo{booktitle}{\emph{Computer Graphics Forum}}, Vol.~\bibinfo{volume}{43}. Wiley Online Library, \bibinfo{pages}{e15041}.
\newblock


\bibitem[Liu et~al\mbox{.}(2023)]%
        {liu2023zero}
\bibfield{author}{\bibinfo{person}{Ruoshi Liu}, \bibinfo{person}{Rundi Wu}, \bibinfo{person}{Basile Van~Hoorick}, \bibinfo{person}{Pavel Tokmakov}, \bibinfo{person}{Sergey Zakharov}, {and} \bibinfo{person}{Carl Vondrick}.} \bibinfo{year}{2023}\natexlab{}.
\newblock \showarticletitle{Zero-1-to-3: Zero-shot one image to 3d object}. In \bibinfo{booktitle}{\emph{Proceedings of the IEEE/CVF International Conference on Computer Vision}}. \bibinfo{pages}{9298--9309}.
\newblock


\bibitem[Liu et~al\mbox{.}(2018)]%
        {liu2018large}
\bibfield{author}{\bibinfo{person}{Ziwei Liu}, \bibinfo{person}{Ping Luo}, \bibinfo{person}{Xiaogang Wang}, {and} \bibinfo{person}{Xiaoou Tang}.} \bibinfo{year}{2018}\natexlab{}.
\newblock \showarticletitle{Large-scale celebfaces attributes (celeba) dataset}.
\newblock \bibinfo{journal}{\emph{Retrieved August}} \bibinfo{volume}{15}, \bibinfo{number}{2018} (\bibinfo{year}{2018}), \bibinfo{pages}{11}.
\newblock


\bibitem[Lombardi et~al\mbox{.}(2021)]%
        {lombardi2021mvp}
\bibfield{author}{\bibinfo{person}{Stephen Lombardi}, \bibinfo{person}{Tomas Simon}, \bibinfo{person}{Gabriel Schwartz}, \bibinfo{person}{Michael Zollhoefer}, \bibinfo{person}{Yaser Sheikh}, {and} \bibinfo{person}{Jason Saragih}.} \bibinfo{year}{2021}\natexlab{}.
\newblock \showarticletitle{Mixture of Volumetric Primitives for Efficient Neural Rendering}.
\newblock \bibinfo{journal}{\emph{ACM Trans. Graph.}} \bibinfo{volume}{40}, \bibinfo{number}{4}, Article \bibinfo{articleno}{59} (\bibinfo{date}{jul} \bibinfo{year}{2021}), \bibinfo{numpages}{13}~pages.
\newblock
\showISSN{0730-0301}
\urldef\tempurl%
\url{https://doi.org/10.1145/3450626.3459863}
\showDOI{\tempurl}


\bibitem[Massague et~al\mbox{.}(2024)]%
        {comas2024magicmirror}
\bibfield{author}{\bibinfo{person}{Armand~Comas Massague}, \bibinfo{person}{Di Qiu}, \bibinfo{person}{Menglei Chai}, \bibinfo{person}{Marcel~C. Buehler}, \bibinfo{person}{Amit Raj}, \bibinfo{person}{Ruiqi Gao}, \bibinfo{person}{Qiangeng Xu}, \bibinfo{person}{Mark Matthews}, \bibinfo{person}{Paulo F.~U. Gotardo}, \bibinfo{person}{Octavia Camps}, \bibinfo{person}{Sergio Orts}, {and} \bibinfo{person}{Thabo Beeler}.} \bibinfo{year}{2024}\natexlab{}.
\newblock \showarticletitle{MagicMirror: Fast and High-Quality Avatar Generation with a Constrained Search Space}. In \bibinfo{booktitle}{\emph{Proceedings of the European conference on computer vision (ECCV)}}.
\newblock


\bibitem[Melas-Kyriazi et~al\mbox{.}(2023)]%
        {melas2023realfusion}
\bibfield{author}{\bibinfo{person}{Luke Melas-Kyriazi}, \bibinfo{person}{Iro Laina}, \bibinfo{person}{Christian Rupprecht}, {and} \bibinfo{person}{Andrea Vedaldi}.} \bibinfo{year}{2023}\natexlab{}.
\newblock \showarticletitle{Realfusion: 360deg reconstruction of any object from a single image}. In \bibinfo{booktitle}{\emph{Proceedings of the IEEE/CVF conference on computer vision and pattern recognition}}. \bibinfo{pages}{8446--8455}.
\newblock


\bibitem[Mihajlovic et~al\mbox{.}(2022)]%
        {keypointnerf}
\bibfield{author}{\bibinfo{person}{Marko Mihajlovic}, \bibinfo{person}{Aayush Bansal}, \bibinfo{person}{Michael Zollhoefer}, \bibinfo{person}{Siyu Tang}, {and} \bibinfo{person}{Shunsuke Saito}.} \bibinfo{year}{2022}\natexlab{}.
\newblock \showarticletitle{{KeypointNeRF}: Generalizing Image-based Volumetric Avatars using Relative Spatial Encoding of Keypoints}. In \bibinfo{booktitle}{\emph{European conference on computer vision}}.
\newblock


\bibitem[Mildenhall et~al\mbox{.}(2020)]%
        {mildenhall2020nerf}
\bibfield{author}{\bibinfo{person}{Ben Mildenhall}, \bibinfo{person}{Pratul~P Srinivasan}, \bibinfo{person}{Matthew Tancik}, \bibinfo{person}{Jonathan~T Barron}, \bibinfo{person}{Ravi Ramamoorthi}, {and} \bibinfo{person}{Ren Ng}.} \bibinfo{year}{2020}\natexlab{}.
\newblock \showarticletitle{Nerf: Representing scenes as neural radiance fields for view synthesis}. In \bibinfo{booktitle}{\emph{European Conference on Computer Vision}}. Springer, \bibinfo{pages}{405--421}.
\newblock


\bibitem[Nichol et~al\mbox{.}(2018)]%
        {nichol2018first}
\bibfield{author}{\bibinfo{person}{Alex Nichol}, \bibinfo{person}{Joshua Achiam}, {and} \bibinfo{person}{John Schulman}.} \bibinfo{year}{2018}\natexlab{}.
\newblock \showarticletitle{On first-order meta-learning algorithms}.
\newblock \bibinfo{journal}{\emph{arXiv preprint arXiv:1803.02999}} (\bibinfo{year}{2018}).
\newblock


\bibitem[Niemeyer et~al\mbox{.}(2022)]%
        {regnerf}
\bibfield{author}{\bibinfo{person}{Michael Niemeyer}, \bibinfo{person}{Jonathan~T. Barron}, \bibinfo{person}{Ben Mildenhall}, \bibinfo{person}{Mehdi S.~M. Sajjadi}, \bibinfo{person}{Andreas Geiger}, {and} \bibinfo{person}{Noha Radwan}.} \bibinfo{year}{2022}\natexlab{}.
\newblock \showarticletitle{RegNeRF: Regularizing Neural Radiance Fields for View Synthesis from Sparse Inputs}. In \bibinfo{booktitle}{\emph{Proc. IEEE Conf. on Computer Vision and Pattern Recognition (CVPR)}}.
\newblock


\bibitem[Niemeyer and Geiger(2021)]%
        {niemeyer2021giraffe}
\bibfield{author}{\bibinfo{person}{Michael Niemeyer} {and} \bibinfo{person}{Andreas Geiger}.} \bibinfo{year}{2021}\natexlab{}.
\newblock \showarticletitle{Giraffe: Representing scenes as compositional generative neural feature fields}. In \bibinfo{booktitle}{\emph{Proceedings of the IEEE/CVF Conference on Computer Vision and Pattern Recognition}}. \bibinfo{pages}{11453--11464}.
\newblock


\bibitem[Or-El et~al\mbox{.}(2022)]%
        {or2022stylesdf}
\bibfield{author}{\bibinfo{person}{Roy Or-El}, \bibinfo{person}{Xuan Luo}, \bibinfo{person}{Mengyi Shan}, \bibinfo{person}{Eli Shechtman}, \bibinfo{person}{Jeong~Joon Park}, {and} \bibinfo{person}{Ira Kemelmacher-Shlizerman}.} \bibinfo{year}{2022}\natexlab{}.
\newblock \showarticletitle{Stylesdf: High-resolution 3d-consistent image and geometry generation}. In \bibinfo{booktitle}{\emph{Proceedings of the IEEE/CVF Conference on Computer Vision and Pattern Recognition}}. \bibinfo{pages}{13503--13513}.
\newblock


\bibitem[Pandey et~al\mbox{.}(2021)]%
        {pandey2021total}
\bibfield{author}{\bibinfo{person}{Rohit Pandey}, \bibinfo{person}{Sergio~Orts Escolano}, \bibinfo{person}{Chloe Legendre}, \bibinfo{person}{Christian Haene}, \bibinfo{person}{Sofien Bouaziz}, \bibinfo{person}{Christoph Rhemann}, \bibinfo{person}{Paul Debevec}, {and} \bibinfo{person}{Sean Fanello}.} \bibinfo{year}{2021}\natexlab{}.
\newblock \showarticletitle{Total relighting: learning to relight portraits for background replacement}.
\newblock \bibinfo{journal}{\emph{ACM Transactions on Graphics (TOG)}} \bibinfo{volume}{40}, \bibinfo{number}{4} (\bibinfo{year}{2021}), \bibinfo{pages}{1--21}.
\newblock


\bibitem[Papantoniou et~al\mbox{.}(2023)]%
        {papantoniou2023relightify}
\bibfield{author}{\bibinfo{person}{Foivos~Paraperas Papantoniou}, \bibinfo{person}{Alexandros Lattas}, \bibinfo{person}{Stylianos Moschoglou}, {and} \bibinfo{person}{Stefanos Zafeiriou}.} \bibinfo{year}{2023}\natexlab{}.
\newblock \showarticletitle{Relightify: Relightable 3d faces from a single image via diffusion models}. In \bibinfo{booktitle}{\emph{Proceedings of the IEEE/CVF International Conference on Computer Vision}}. \bibinfo{pages}{8806--8817}.
\newblock


\bibitem[Park et~al\mbox{.}(2021a)]%
        {park2021nerfies}
\bibfield{author}{\bibinfo{person}{Keunhong Park}, \bibinfo{person}{Utkarsh Sinha}, \bibinfo{person}{Jonathan~T. Barron}, \bibinfo{person}{Sofien Bouaziz}, \bibinfo{person}{Dan~B Goldman}, \bibinfo{person}{Steven~M. Seitz}, {and} \bibinfo{person}{Ricardo Martin-Brualla}.} \bibinfo{year}{2021}\natexlab{a}.
\newblock \showarticletitle{Nerfies: Deformable Neural Radiance Fields}.
\newblock \bibinfo{journal}{\emph{ICCV}} (\bibinfo{year}{2021}).
\newblock


\bibitem[Park et~al\mbox{.}(2021b)]%
        {park2021hypernerf}
\bibfield{author}{\bibinfo{person}{Keunhong Park}, \bibinfo{person}{Utkarsh Sinha}, \bibinfo{person}{Peter Hedman}, \bibinfo{person}{Jonathan~T. Barron}, \bibinfo{person}{Sofien Bouaziz}, \bibinfo{person}{Dan~B Goldman}, \bibinfo{person}{Ricardo Martin-Brualla}, {and} \bibinfo{person}{Steven~M. Seitz}.} \bibinfo{year}{2021}\natexlab{b}.
\newblock \showarticletitle{HyperNeRF: A Higher-Dimensional Representation for Topologically Varying Neural Radiance Fields}.
\newblock \bibinfo{journal}{\emph{ACM Trans. Graph.}} \bibinfo{volume}{40}, \bibinfo{number}{6}, Article \bibinfo{articleno}{238} (\bibinfo{date}{dec} \bibinfo{year}{2021}).
\newblock


\bibitem[Pascalis and Kelly(2009)]%
        {faceprocess}
\bibfield{author}{\bibinfo{person}{Olivier Pascalis} {and} \bibinfo{person}{David~J Kelly}.} \bibinfo{year}{2009}\natexlab{}.
\newblock \showarticletitle{The origins of face processing in humans: Phylogeny and ontogeny}.
\newblock \bibinfo{journal}{\emph{Perspectives on psychological science}} \bibinfo{volume}{4}, \bibinfo{number}{2} (\bibinfo{year}{2009}), \bibinfo{pages}{200--209}.
\newblock


\bibitem[Poole et~al\mbox{.}(2022)]%
        {poole2022dreamfusion}
\bibfield{author}{\bibinfo{person}{Ben Poole}, \bibinfo{person}{Ajay Jain}, \bibinfo{person}{Jonathan~T. Barron}, {and} \bibinfo{person}{Ben Mildenhall}.} \bibinfo{year}{2022}\natexlab{}.
\newblock \showarticletitle{DreamFusion: Text-to-3D using 2D Diffusion}.
\newblock \bibinfo{journal}{\emph{arXiv}} (\bibinfo{year}{2022}).
\newblock


\bibitem[Prinzler et~al\mbox{.}(2023)]%
        {diner}
\bibfield{author}{\bibinfo{person}{Malte Prinzler}, \bibinfo{person}{Otmar Hilliges}, {and} \bibinfo{person}{Justus Thies}.} \bibinfo{year}{2023}\natexlab{}.
\newblock \showarticletitle{Diner: Depth-aware image-based neural radiance fields}. In \bibinfo{booktitle}{\emph{Proceedings of the IEEE/CVF Conference on Computer Vision and Pattern Recognition}}. \bibinfo{pages}{12449--12459}.
\newblock


\bibitem[Ramon et~al\mbox{.}(2021)]%
        {h3dnet}
\bibfield{author}{\bibinfo{person}{Eduard Ramon}, \bibinfo{person}{Gil Triginer}, \bibinfo{person}{Janna Escur}, \bibinfo{person}{Albert Pumarola}, \bibinfo{person}{Jaime Garcia}, \bibinfo{person}{Xavier Giro-i Nieto}, {and} \bibinfo{person}{Francesc Moreno-Noguer}.} \bibinfo{year}{2021}\natexlab{}.
\newblock \showarticletitle{H3d-net: Few-shot high-fidelity 3d head reconstruction}. In \bibinfo{booktitle}{\emph{Proceedings of the IEEE/CVF International Conference on Computer Vision}}. \bibinfo{pages}{5620--5629}.
\newblock


\bibitem[Rao et~al\mbox{.}(2022)]%
        {prao2022vorf}
\bibfield{author}{\bibinfo{person}{Pramod Rao}, \bibinfo{person}{Mallikarjun B~R}, \bibinfo{person}{Gereon Fox}, \bibinfo{person}{Tim Weyrich}, \bibinfo{person}{Bernd Bickel}, \bibinfo{person}{Hans-Peter Seidel}, \bibinfo{person}{Hanspeter Pfister}, \bibinfo{person}{Wojciech Matusik}, \bibinfo{person}{Ayush Tewari}, \bibinfo{person}{Christian Theobalt}, {and} \bibinfo{person}{Mohamed Elgharib}.} \bibinfo{year}{2022}\natexlab{}.
\newblock \showarticletitle{VoRF: Volumetric Relightable Faces}.
\newblock \bibinfo{journal}{\emph{British Machine Vision Conference (BMVC)}} (\bibinfo{year}{2022}).
\newblock


\bibitem[Rebain et~al\mbox{.}(2022)]%
        {lolnerf}
\bibfield{author}{\bibinfo{person}{Daniel Rebain}, \bibinfo{person}{Mark Matthews}, \bibinfo{person}{Kwang~Moo Yi}, \bibinfo{person}{Dmitry Lagun}, {and} \bibinfo{person}{Andrea Tagliasacchi}.} \bibinfo{year}{2022}\natexlab{}.
\newblock \showarticletitle{Lolnerf: Learn from one look}. In \bibinfo{booktitle}{\emph{Proceedings of the IEEE/CVF Conference on Computer Vision and Pattern Recognition}}. \bibinfo{pages}{1558--1567}.
\newblock


\bibitem[Sarkar et~al\mbox{.}(2023)]%
        {sarkar2023litnerf}
\bibfield{author}{\bibinfo{person}{Kripasindhu Sarkar}, \bibinfo{person}{Marcel~C B{\"u}hler}, \bibinfo{person}{Gengyan Li}, \bibinfo{person}{Daoye Wang}, \bibinfo{person}{Delio Vicini}, \bibinfo{person}{J{\'e}r{\'e}my Riviere}, \bibinfo{person}{Yinda Zhang}, \bibinfo{person}{Sergio Orts-Escolano}, \bibinfo{person}{Paulo Gotardo}, \bibinfo{person}{Thabo Beeler}, {et~al\mbox{.}}} \bibinfo{year}{2023}\natexlab{}.
\newblock \showarticletitle{LitNeRF: Intrinsic Radiance Decomposition for High-Quality View Synthesis and Relighting of Faces}. In \bibinfo{booktitle}{\emph{SIGGRAPH Asia 2023 Conference Papers}}. \bibinfo{pages}{1--11}.
\newblock


\bibitem[Schwarz et~al\mbox{.}(2020)]%
        {schwarz2020graf}
\bibfield{author}{\bibinfo{person}{Katja Schwarz}, \bibinfo{person}{Yiyi Liao}, \bibinfo{person}{Michael Niemeyer}, {and} \bibinfo{person}{Andreas Geiger}.} \bibinfo{year}{2020}\natexlab{}.
\newblock \showarticletitle{Graf: Generative radiance fields for 3d-aware image synthesis}.
\newblock \bibinfo{journal}{\emph{Advances in Neural Information Processing Systems}}  \bibinfo{volume}{33} (\bibinfo{year}{2020}), \bibinfo{pages}{20154--20166}.
\newblock


\bibitem[Simonyan and Zisserman(2015)]%
        {simonyan2014veryvgg}
\bibfield{author}{\bibinfo{person}{Karen Simonyan} {and} \bibinfo{person}{Andrew Zisserman}.} \bibinfo{year}{2015}\natexlab{}.
\newblock \showarticletitle{Very deep convolutional networks for large-scale image recognition}. In \bibinfo{booktitle}{\emph{International Conference on Learning Representations (ICLR)}}.
\newblock


\bibitem[Sinha et~al\mbox{.}(2006)]%
        {sinha06}
\bibfield{author}{\bibinfo{person}{Pawan Sinha}, \bibinfo{person}{Benjamin Balas}, \bibinfo{person}{Yuri Ostrovsky}, {and} \bibinfo{person}{Richard Russell}.} \bibinfo{year}{2006}\natexlab{}.
\newblock \showarticletitle{Face Recognition by Humans: Nineteen Results All Computer Vision Researchers Should Know About}.
\newblock \bibinfo{journal}{\emph{Proc. IEEE}} \bibinfo{volume}{94}, \bibinfo{number}{11} (\bibinfo{year}{2006}), \bibinfo{pages}{1948--1962}.
\newblock
\urldef\tempurl%
\url{https://doi.org/10.1109/JPROC.2006.884093}
\showDOI{\tempurl}


\bibitem[Sitzmann et~al\mbox{.}(2020)]%
        {sitzmann2020metasdf}
\bibfield{author}{\bibinfo{person}{Vincent Sitzmann}, \bibinfo{person}{Eric Chan}, \bibinfo{person}{Richard Tucker}, \bibinfo{person}{Noah Snavely}, {and} \bibinfo{person}{Gordon Wetzstein}.} \bibinfo{year}{2020}\natexlab{}.
\newblock \showarticletitle{Metasdf: Meta-learning signed distance functions}.
\newblock \bibinfo{journal}{\emph{Advances in Neural Information Processing Systems}}  \bibinfo{volume}{33} (\bibinfo{year}{2020}), \bibinfo{pages}{10136--10147}.
\newblock


\bibitem[Sun et~al\mbox{.}(2021)]%
        {sun2021nelf}
\bibfield{author}{\bibinfo{person}{Tiancheng Sun}, \bibinfo{person}{Kai-En Lin}, \bibinfo{person}{Sai Bi}, \bibinfo{person}{Zexiang Xu}, {and} \bibinfo{person}{Ravi Ramamoorthi}.} \bibinfo{year}{2021}\natexlab{}.
\newblock \showarticletitle{NeLF: Neural Light-transport Field for Portrait View Synthesis and Relighting}. In \bibinfo{booktitle}{\emph{Eurographics Symposium on Rendering}}.
\newblock


\bibitem[Tan et~al\mbox{.}(2022)]%
        {voluxgan}
\bibfield{author}{\bibinfo{person}{Feitong Tan}, \bibinfo{person}{Sean Fanello}, \bibinfo{person}{Abhimitra Meka}, \bibinfo{person}{Sergio Orts-Escolano}, \bibinfo{person}{Danhang Tang}, \bibinfo{person}{Rohit Pandey}, \bibinfo{person}{Jonathan Taylor}, \bibinfo{person}{Ping Tan}, {and} \bibinfo{person}{Yinda Zhang}.} \bibinfo{year}{2022}\natexlab{}.
\newblock \showarticletitle{VoLux-GAN: A Generative Model for 3D Face Synthesis with HDRI Relighting}. In \bibinfo{booktitle}{\emph{ACM SIGGRAPH 2022 Conference Proceedings}} (Vancouver, BC, Canada) \emph{(\bibinfo{series}{SIGGRAPH '22})}. \bibinfo{publisher}{Association for Computing Machinery}, \bibinfo{address}{New York, NY, USA}, Article \bibinfo{articleno}{58}, \bibinfo{numpages}{9}~pages.
\newblock
\showISBNx{9781450393379}
\urldef\tempurl%
\url{https://doi.org/10.1145/3528233.3530751}
\showDOI{\tempurl}


\bibitem[Tancik et~al\mbox{.}(2021)]%
        {tancik2021learned}
\bibfield{author}{\bibinfo{person}{Matthew Tancik}, \bibinfo{person}{Ben Mildenhall}, \bibinfo{person}{Terrance Wang}, \bibinfo{person}{Divi Schmidt}, \bibinfo{person}{Pratul~P Srinivasan}, \bibinfo{person}{Jonathan~T Barron}, {and} \bibinfo{person}{Ren Ng}.} \bibinfo{year}{2021}\natexlab{}.
\newblock \showarticletitle{Learned initializations for optimizing coordinate-based neural representations}. In \bibinfo{booktitle}{\emph{Proceedings of the IEEE/CVF Conference on Computer Vision and Pattern Recognition}}. \bibinfo{pages}{2846--2855}.
\newblock


\bibitem[Tang et~al\mbox{.}(2023)]%
        {tang2023make}
\bibfield{author}{\bibinfo{person}{Junshu Tang}, \bibinfo{person}{Tengfei Wang}, \bibinfo{person}{Bo Zhang}, \bibinfo{person}{Ting Zhang}, \bibinfo{person}{Ran Yi}, \bibinfo{person}{Lizhuang Ma}, {and} \bibinfo{person}{Dong Chen}.} \bibinfo{year}{2023}\natexlab{}.
\newblock \showarticletitle{Make-it-3d: High-fidelity 3d creation from a single image with diffusion prior}. In \bibinfo{booktitle}{\emph{Proceedings of the IEEE/CVF International Conference on Computer Vision}}. \bibinfo{pages}{22819--22829}.
\newblock


\bibitem[Trevithick et~al\mbox{.}(2023)]%
        {trevithick2023}
\bibfield{author}{\bibinfo{person}{Alex Trevithick}, \bibinfo{person}{Matthew Chan}, \bibinfo{person}{Michael Stengel}, \bibinfo{person}{Eric~R. Chan}, \bibinfo{person}{Chao Liu}, \bibinfo{person}{Zhiding Yu}, \bibinfo{person}{Sameh Khamis}, \bibinfo{person}{Manmohan Chandraker}, \bibinfo{person}{Ravi Ramamoorthi}, {and} \bibinfo{person}{Koki Nagano}.} \bibinfo{year}{2023}\natexlab{}.
\newblock \showarticletitle{Real-Time Radiance Fields for Single-Image Portrait View Synthesis}. In \bibinfo{booktitle}{\emph{ACM Transactions on Graphics (SIGGRAPH)}}.
\newblock


\bibitem[Truong et~al\mbox{.}(2023)]%
        {sparf2023}
\bibfield{author}{\bibinfo{person}{Prune Truong}, \bibinfo{person}{Marie-Julie Rakotosaona}, \bibinfo{person}{Fabian Manhardt}, {and} \bibinfo{person}{Federico Tombari}.} \bibinfo{year}{2023}\natexlab{}.
\newblock \showarticletitle{SPARF: Neural Radiance Fields from Sparse and Noisy Poses}. \bibinfo{publisher}{{IEEE/CVF} Conference on Computer Vision and Pattern Recognition, {CVPR}}.
\newblock


\bibitem[Vinod et~al\mbox{.}(2024)]%
        {vinod2024teglo}
\bibfield{author}{\bibinfo{person}{Vishal Vinod}, \bibinfo{person}{Tanmay Shah}, {and} \bibinfo{person}{Dmitry Lagun}.} \bibinfo{year}{2024}\natexlab{}.
\newblock \showarticletitle{TEGLO: High Fidelity Canonical Texture Mapping from Single-View Images}. In \bibinfo{booktitle}{\emph{Proceedings of the IEEE/CVF Winter Conference on Applications of Computer Vision}}. \bibinfo{pages}{3585--3595}.
\newblock


\bibitem[Vora et~al\mbox{.}(2021)]%
        {vora2021nesf}
\bibfield{author}{\bibinfo{person}{Suhani Vora}, \bibinfo{person}{Noha Radwan}, \bibinfo{person}{Klaus Greff}, \bibinfo{person}{Henning Meyer}, \bibinfo{person}{Kyle Genova}, \bibinfo{person}{Mehdi~SM Sajjadi}, \bibinfo{person}{Etienne Pot}, \bibinfo{person}{Andrea Tagliasacchi}, {and} \bibinfo{person}{Daniel Duckworth}.} \bibinfo{year}{2021}\natexlab{}.
\newblock \showarticletitle{Nesf: Neural semantic fields for generalizable semantic segmentation of 3d scenes}.
\newblock \bibinfo{journal}{\emph{arXiv preprint arXiv:2111.13260}} (\bibinfo{year}{2021}).
\newblock


\bibitem[Wang et~al\mbox{.}(2022)]%
        {morf}
\bibfield{author}{\bibinfo{person}{Daoye Wang}, \bibinfo{person}{Prashanth Chandran}, \bibinfo{person}{Gaspard Zoss}, \bibinfo{person}{Derek Bradley}, {and} \bibinfo{person}{Paulo Gotardo}.} \bibinfo{year}{2022}\natexlab{}.
\newblock \showarticletitle{MoRF: Morphable Radiance Fields for Multiview Neural Head Modeling}. In \bibinfo{booktitle}{\emph{ACM SIGGRAPH 2022 Conference Proceedings}} (Vancouver, BC, Canada) \emph{(\bibinfo{series}{SIGGRAPH '22})}. \bibinfo{publisher}{Association for Computing Machinery}, \bibinfo{address}{New York, NY, USA}, Article \bibinfo{articleno}{55}, \bibinfo{numpages}{9}~pages.
\newblock
\showISBNx{9781450393379}
\urldef\tempurl%
\url{https://doi.org/10.1145/3528233.3530753}
\showDOI{\tempurl}


\bibitem[Wang et~al\mbox{.}(2021)]%
        {ibrnet}
\bibfield{author}{\bibinfo{person}{Qianqian Wang}, \bibinfo{person}{Zhicheng Wang}, \bibinfo{person}{Kyle Genova}, \bibinfo{person}{Pratul Srinivasan}, \bibinfo{person}{Howard Zhou}, \bibinfo{person}{Jonathan~T. Barron}, \bibinfo{person}{Ricardo Martin-Brualla}, \bibinfo{person}{Noah Snavely}, {and} \bibinfo{person}{Thomas Funkhouser}.} \bibinfo{year}{2021}\natexlab{}.
\newblock \showarticletitle{IBRNet: Learning Multi-View Image-Based Rendering}. In \bibinfo{booktitle}{\emph{CVPR}}.
\newblock


\bibitem[Wang et~al\mbox{.}(2023)]%
        {wang2023rodin}
\bibfield{author}{\bibinfo{person}{Tengfei Wang}, \bibinfo{person}{Bo Zhang}, \bibinfo{person}{Ting Zhang}, \bibinfo{person}{Shuyang Gu}, \bibinfo{person}{Jianmin Bao}, \bibinfo{person}{Tadas Baltrusaitis}, \bibinfo{person}{Jingjing Shen}, \bibinfo{person}{Dong Chen}, \bibinfo{person}{Fang Wen}, \bibinfo{person}{Qifeng Chen}, {et~al\mbox{.}}} \bibinfo{year}{2023}\natexlab{}.
\newblock \showarticletitle{Rodin: A generative model for sculpting 3d digital avatars using diffusion}. In \bibinfo{booktitle}{\emph{Proceedings of the IEEE/CVF Conference on Computer Vision and Pattern Recognition}}. \bibinfo{pages}{4563--4573}.
\newblock


\bibitem[Wang et~al\mbox{.}(2024)]%
        {wang2024prolificdreamer}
\bibfield{author}{\bibinfo{person}{Zhengyi Wang}, \bibinfo{person}{Cheng Lu}, \bibinfo{person}{Yikai Wang}, \bibinfo{person}{Fan Bao}, \bibinfo{person}{Chongxuan Li}, \bibinfo{person}{Hang Su}, {and} \bibinfo{person}{Jun Zhu}.} \bibinfo{year}{2024}\natexlab{}.
\newblock \showarticletitle{Prolificdreamer: High-fidelity and diverse text-to-3d generation with variational score distillation}.
\newblock \bibinfo{journal}{\emph{Advances in Neural Information Processing Systems}}  \bibinfo{volume}{36} (\bibinfo{year}{2024}).
\newblock


\bibitem[Wood et~al\mbox{.}(2021)]%
        {wood2021fake}
\bibfield{author}{\bibinfo{person}{Erroll Wood}, \bibinfo{person}{Tadas Baltru{\v{s}}aitis}, \bibinfo{person}{Charlie Hewitt}, \bibinfo{person}{Sebastian Dziadzio}, \bibinfo{person}{Thomas~J Cashman}, {and} \bibinfo{person}{Jamie Shotton}.} \bibinfo{year}{2021}\natexlab{}.
\newblock \showarticletitle{Fake it till you make it: face analysis in the wild using synthetic data alone}. In \bibinfo{booktitle}{\emph{Proceedings of the IEEE/CVF international conference on computer vision}}. \bibinfo{pages}{3681--3691}.
\newblock


\bibitem[Wood et~al\mbox{.}(2022)]%
        {wood20223d}
\bibfield{author}{\bibinfo{person}{Erroll Wood}, \bibinfo{person}{Tadas Baltru{\v{s}}aitis}, \bibinfo{person}{Charlie Hewitt}, \bibinfo{person}{Matthew Johnson}, \bibinfo{person}{Jingjing Shen}, \bibinfo{person}{Nikola Milosavljevi{\'c}}, \bibinfo{person}{Daniel Wilde}, \bibinfo{person}{Stephan Garbin}, \bibinfo{person}{Toby Sharp}, \bibinfo{person}{Ivan Stojiljkovi{\'c}}, {et~al\mbox{.}}} \bibinfo{year}{2022}\natexlab{}.
\newblock \showarticletitle{3d face reconstruction with dense landmarks}. In \bibinfo{booktitle}{\emph{European Conference on Computer Vision}}. Springer, \bibinfo{pages}{160--177}.
\newblock


\bibitem[Wu et~al\mbox{.}(2024)]%
        {wu2023reconfusion}
\bibfield{author}{\bibinfo{person}{Rundi Wu}, \bibinfo{person}{Ben Mildenhall}, \bibinfo{person}{Philipp Henzler}, \bibinfo{person}{Keunhong Park}, \bibinfo{person}{Ruiqi Gao}, \bibinfo{person}{Daniel Watson}, \bibinfo{person}{Pratul~P Srinivasan}, \bibinfo{person}{Dor Verbin}, \bibinfo{person}{Jonathan~T Barron}, \bibinfo{person}{Ben Poole}, {et~al\mbox{.}}} \bibinfo{year}{2024}\natexlab{}.
\newblock \showarticletitle{Reconfusion: 3d reconstruction with diffusion priors}. In \bibinfo{booktitle}{\emph{Proceedings of the IEEE/CVF Conference on Computer Vision and Pattern Recognition}}. \bibinfo{pages}{21551--21561}.
\newblock


\bibitem[Wuu et~al\mbox{.}(2022)]%
        {wuu2022multiface}
\bibfield{author}{\bibinfo{person}{Cheng-hsin Wuu}, \bibinfo{person}{Ningyuan Zheng}, \bibinfo{person}{Scott Ardisson}, \bibinfo{person}{Rohan Bali}, \bibinfo{person}{Danielle Belko}, \bibinfo{person}{Eric Brockmeyer}, \bibinfo{person}{Lucas Evans}, \bibinfo{person}{Timothy Godisart}, \bibinfo{person}{Hyowon Ha}, \bibinfo{person}{Xuhua Huang}, \bibinfo{person}{Alexander Hypes}, \bibinfo{person}{Taylor Koska}, \bibinfo{person}{Steven Krenn}, \bibinfo{person}{Stephen Lombardi}, \bibinfo{person}{Xiaomin Luo}, \bibinfo{person}{Kevyn McPhail}, \bibinfo{person}{Laura Millerschoen}, \bibinfo{person}{Michal Perdoch}, \bibinfo{person}{Mark Pitts}, \bibinfo{person}{Alexander Richard}, \bibinfo{person}{Jason Saragih}, \bibinfo{person}{Junko Saragih}, \bibinfo{person}{Takaaki Shiratori}, \bibinfo{person}{Tomas Simon}, \bibinfo{person}{Matt Stewart}, \bibinfo{person}{Autumn Trimble}, \bibinfo{person}{Xinshuo Weng}, \bibinfo{person}{David Whitewolf}, \bibinfo{person}{Chenglei Wu}, \bibinfo{person}{Shoou-I Yu}, {and}
  \bibinfo{person}{Yaser Sheikh}.} \bibinfo{year}{2022}\natexlab{}.
\newblock \showarticletitle{Multiface: A Dataset for Neural Face Rendering}. In \bibinfo{booktitle}{\emph{arXiv}}.
\newblock
\urldef\tempurl%
\url{https://doi.org/10.48550/ARXIV.2207.11243}
\showDOI{\tempurl}


\bibitem[Xu et~al\mbox{.}(2022)]%
        {xu2022sinnerf}
\bibfield{author}{\bibinfo{person}{Dejia Xu}, \bibinfo{person}{Yifan Jiang}, \bibinfo{person}{Peihao Wang}, \bibinfo{person}{Zhiwen Fan}, \bibinfo{person}{Humphrey Shi}, {and} \bibinfo{person}{Zhangyang Wang}.} \bibinfo{year}{2022}\natexlab{}.
\newblock \showarticletitle{Sinnerf: Training neural radiance fields on complex scenes from a single image}. In \bibinfo{booktitle}{\emph{Computer Vision--ECCV 2022: 17th European Conference, Tel Aviv, Israel, October 23--27, 2022, Proceedings, Part XXII}}. Springer, \bibinfo{pages}{736--753}.
\newblock


\bibitem[Xu et~al\mbox{.}(2023)]%
        {xu2023latentavatar}
\bibfield{author}{\bibinfo{person}{Yuelang Xu}, \bibinfo{person}{Hongwen Zhang}, \bibinfo{person}{Lizhen Wang}, \bibinfo{person}{Xiaochen Zhao}, \bibinfo{person}{Han Huang}, \bibinfo{person}{Guojun Qi}, {and} \bibinfo{person}{Yebin Liu}.} \bibinfo{year}{2023}\natexlab{}.
\newblock \showarticletitle{Latentavatar: Learning latent expression code for expressive neural head avatar}. In \bibinfo{booktitle}{\emph{ACM SIGGRAPH 2023 Conference Proceedings}}. \bibinfo{pages}{1--10}.
\newblock


\bibitem[Yang et~al\mbox{.}(2023)]%
        {Yang2023FreeNeRF}
\bibfield{author}{\bibinfo{person}{Jiawei Yang}, \bibinfo{person}{Marco Pavone}, {and} \bibinfo{person}{Yue Wang}.} \bibinfo{year}{2023}\natexlab{}.
\newblock \showarticletitle{FreeNeRF: Improving Few-shot Neural Rendering with Free Frequency Regularization}. In \bibinfo{booktitle}{\emph{Proc. IEEE Conf. on Computer Vision and Pattern Recognition (CVPR)}}.
\newblock


\bibitem[Yeh et~al\mbox{.}(2022)]%
        {yeh2022learning}
\bibfield{author}{\bibinfo{person}{Yu-Ying Yeh}, \bibinfo{person}{Koki Nagano}, \bibinfo{person}{Sameh Khamis}, \bibinfo{person}{Jan Kautz}, \bibinfo{person}{Ming-Yu Liu}, {and} \bibinfo{person}{Ting-Chun Wang}.} \bibinfo{year}{2022}\natexlab{}.
\newblock \showarticletitle{Learning to Relight Portrait Images via a Virtual Light Stage and Synthetic-to-Real Adaptation}.
\newblock \bibinfo{journal}{\emph{ACM Transactions on Graphics (TOG)}} (\bibinfo{year}{2022}).
\newblock


\bibitem[Yu et~al\mbox{.}(2021)]%
        {pixelnerf}
\bibfield{author}{\bibinfo{person}{Alex Yu}, \bibinfo{person}{Vickie Ye}, \bibinfo{person}{Matthew Tancik}, {and} \bibinfo{person}{Angjoo Kanazawa}.} \bibinfo{year}{2021}\natexlab{}.
\newblock \showarticletitle{{pixelNeRF}: Neural Radiance Fields from One or Few Images}. In \bibinfo{booktitle}{\emph{CVPR}}.
\newblock


\bibitem[Zakharov et~al\mbox{.}(2019)]%
        {Zakharov_2019_ICCV}
\bibfield{author}{\bibinfo{person}{Egor Zakharov}, \bibinfo{person}{Aliaksandra Shysheya}, \bibinfo{person}{Egor Burkov}, {and} \bibinfo{person}{Victor Lempitsky}.} \bibinfo{year}{2019}\natexlab{}.
\newblock \showarticletitle{Few-Shot Adversarial Learning of Realistic Neural Talking Head Models}. In \bibinfo{booktitle}{\emph{Proceedings of the IEEE/CVF International Conference on Computer Vision (ICCV)}}.
\newblock


\bibitem[Zeng et~al\mbox{.}(2023)]%
        {zeng2023avatarbooth}
\bibfield{author}{\bibinfo{person}{Yifei Zeng}, \bibinfo{person}{Yuanxun Lu}, \bibinfo{person}{Xinya Ji}, \bibinfo{person}{Yao Yao}, \bibinfo{person}{Hao Zhu}, {and} \bibinfo{person}{Xun Cao}.} \bibinfo{year}{2023}\natexlab{}.
\newblock \showarticletitle{Avatarbooth: High-quality and customizable 3d human avatar generation}.
\newblock \bibinfo{journal}{\emph{arXiv preprint arXiv:2306.09864}} (\bibinfo{year}{2023}).
\newblock


\bibitem[Zhang et~al\mbox{.}(2022)]%
        {zhang2022fdnerf}
\bibfield{author}{\bibinfo{person}{Jingbo Zhang}, \bibinfo{person}{Xiaoyu Li}, \bibinfo{person}{Ziyu Wan}, \bibinfo{person}{Can Wang}, {and} \bibinfo{person}{Jing Liao}.} \bibinfo{year}{2022}\natexlab{}.
\newblock \showarticletitle{Fdnerf: Few-shot dynamic neural radiance fields for face reconstruction and expression editing}. In \bibinfo{booktitle}{\emph{SIGGRAPH Asia 2022 Conference Papers}}. \bibinfo{pages}{1--9}.
\newblock


\bibitem[Zhang et~al\mbox{.}(2018)]%
        {zhang2018perceptual}
\bibfield{author}{\bibinfo{person}{Richard Zhang}, \bibinfo{person}{Phillip Isola}, \bibinfo{person}{Alexei~A Efros}, \bibinfo{person}{Eli Shechtman}, {and} \bibinfo{person}{Oliver Wang}.} \bibinfo{year}{2018}\natexlab{}.
\newblock \showarticletitle{The Unreasonable Effectiveness of Deep Features as a Perceptual Metric}. In \bibinfo{booktitle}{\emph{CVPR}}.
\newblock


\bibitem[Zhao et~al\mbox{.}(2023)]%
        {zhao2023havatar}
\bibfield{author}{\bibinfo{person}{Xiaochen Zhao}, \bibinfo{person}{Lizhen Wang}, \bibinfo{person}{Jingxiang Sun}, \bibinfo{person}{Hongwen Zhang}, \bibinfo{person}{Jinli Suo}, {and} \bibinfo{person}{Yebin Liu}.} \bibinfo{year}{2023}\natexlab{}.
\newblock \showarticletitle{Havatar: High-fidelity head avatar via facial model conditioned neural radiance field}.
\newblock \bibinfo{journal}{\emph{ACM Transactions on Graphics}} \bibinfo{volume}{43}, \bibinfo{number}{1} (\bibinfo{year}{2023}), \bibinfo{pages}{1--16}.
\newblock


\bibitem[Zheng et~al\mbox{.}(2022)]%
        {zheng2022imavatar}
\bibfield{author}{\bibinfo{person}{Yufeng Zheng}, \bibinfo{person}{Victoria~Fern{\'a}ndez Abrevaya}, \bibinfo{person}{Marcel~C B{\"u}hler}, \bibinfo{person}{Xu Chen}, \bibinfo{person}{Michael~J Black}, {and} \bibinfo{person}{Otmar Hilliges}.} \bibinfo{year}{2022}\natexlab{}.
\newblock \showarticletitle{Im avatar: Implicit morphable head avatars from videos}. In \bibinfo{booktitle}{\emph{Proceedings of the IEEE/CVF Conference on Computer Vision and Pattern Recognition}}. \bibinfo{pages}{13545--13555}.
\newblock


\bibitem[Zheng et~al\mbox{.}(2023)]%
        {zheng2023pointavatar}
\bibfield{author}{\bibinfo{person}{Yufeng Zheng}, \bibinfo{person}{Wang Yifan}, \bibinfo{person}{Gordon Wetzstein}, \bibinfo{person}{Michael~J Black}, {and} \bibinfo{person}{Otmar Hilliges}.} \bibinfo{year}{2023}\natexlab{}.
\newblock \showarticletitle{Pointavatar: Deformable point-based head avatars from videos}. In \bibinfo{booktitle}{\emph{Proceedings of the IEEE/CVF conference on computer vision and pattern recognition}}. \bibinfo{pages}{21057--21067}.
\newblock


\bibitem[Zhou et~al\mbox{.}(2021)]%
        {zhou2021CIPS3D}
\bibfield{author}{\bibinfo{person}{Peng Zhou}, \bibinfo{person}{Lingxi Xie}, \bibinfo{person}{Bingbing Ni}, {and} \bibinfo{person}{Qi Tian}.} \bibinfo{year}{2021}\natexlab{}.
\newblock \showarticletitle{{{CIPS}}-{{3D}}: A {{3D}}-{{Aware Generator}} of {{GANs Based}} on {{Conditionally}}-{{Independent Pixel Synthesis}}}.
\newblock  (\bibinfo{year}{2021}).
\newblock
\showeprint[arxiv]{2110.09788}


\bibitem[Zhu et~al\mbox{.}(2023)]%
        {facescape}
\bibfield{author}{\bibinfo{person}{Hao Zhu}, \bibinfo{person}{Haotian Yang}, \bibinfo{person}{Longwei Guo}, \bibinfo{person}{Yidi Zhang}, \bibinfo{person}{Yanru Wang}, \bibinfo{person}{Mingkai Huang}, \bibinfo{person}{Menghua Wu}, \bibinfo{person}{Qiu Shen}, \bibinfo{person}{Ruigang Yang}, {and} \bibinfo{person}{Xun Cao}.} \bibinfo{year}{2023}\natexlab{}.
\newblock \showarticletitle{FaceScape: 3D Facial Dataset and Benchmark for Single-View 3D Face Reconstruction}.
\newblock \bibinfo{journal}{\emph{IEEE Transactions on Pattern Analysis and Machine Intelligence (TPAMI)}} (\bibinfo{year}{2023}).
\newblock


\end{thebibliography}

\begin{figure*}[ht]
\begin{center}
\small
\setlength{\tabcolsep}{2pt}
\newcommand{\width}{5cm}
\newcommand{\inputwidth}{1.5cm}
\begin{tabular}{ccc}
  \includegraphics[width=\inputwidth]{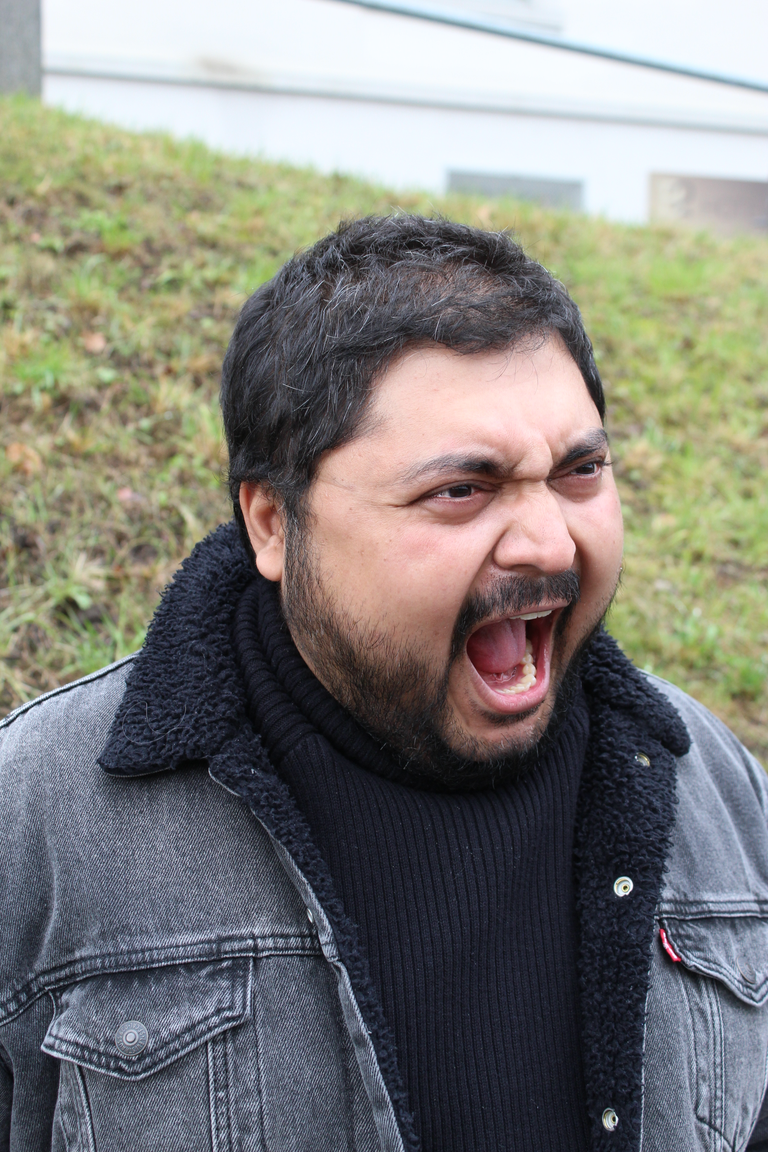}
      \includegraphics[width=\inputwidth]{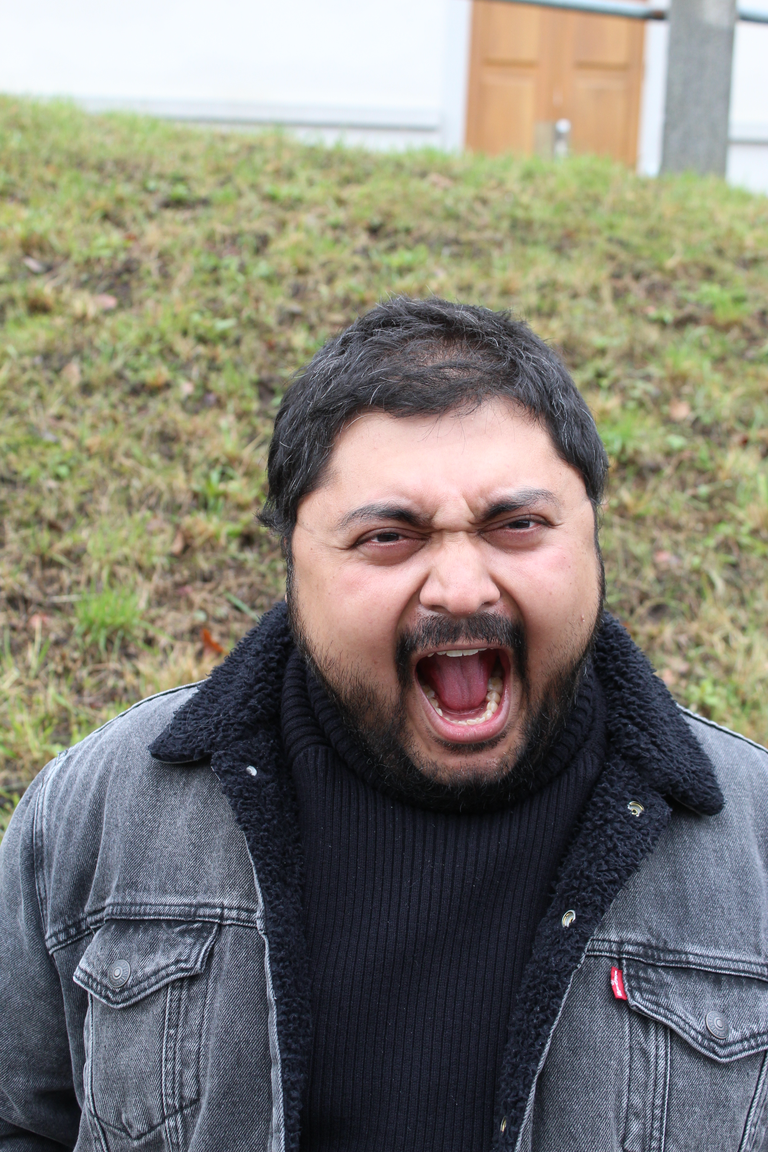}
      \includegraphics[width=\inputwidth]{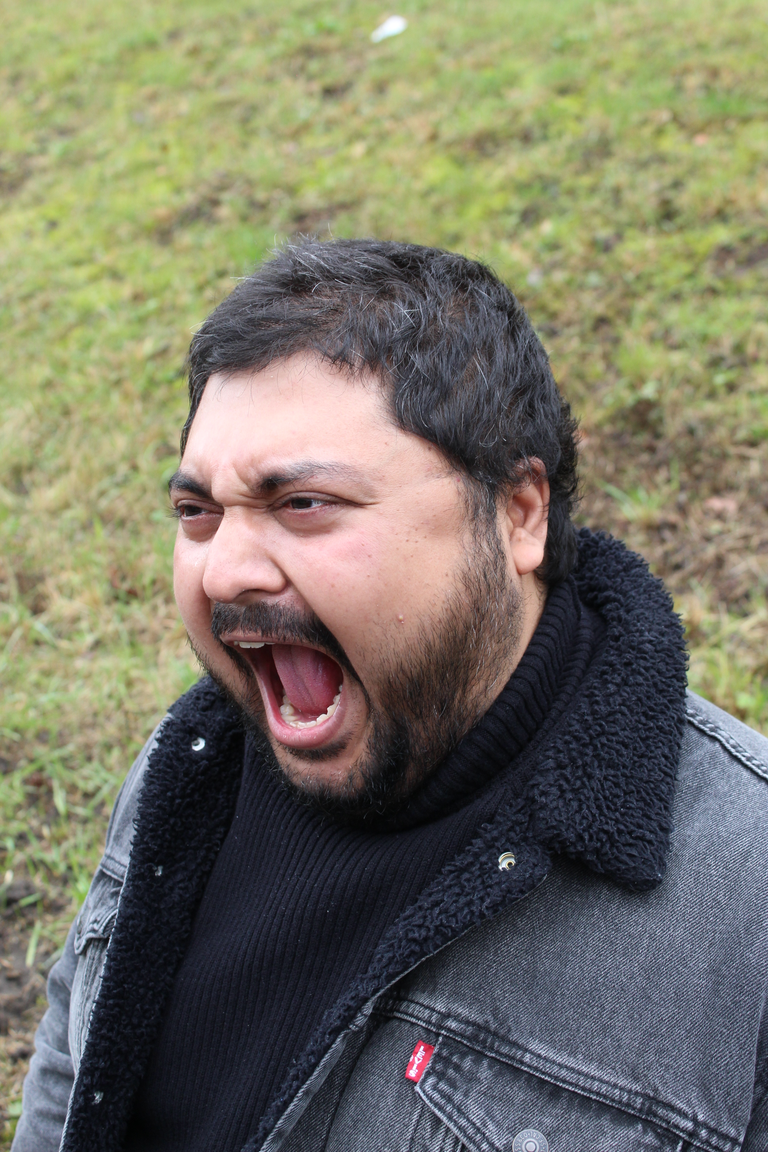}&
      \includegraphics[width=\width]{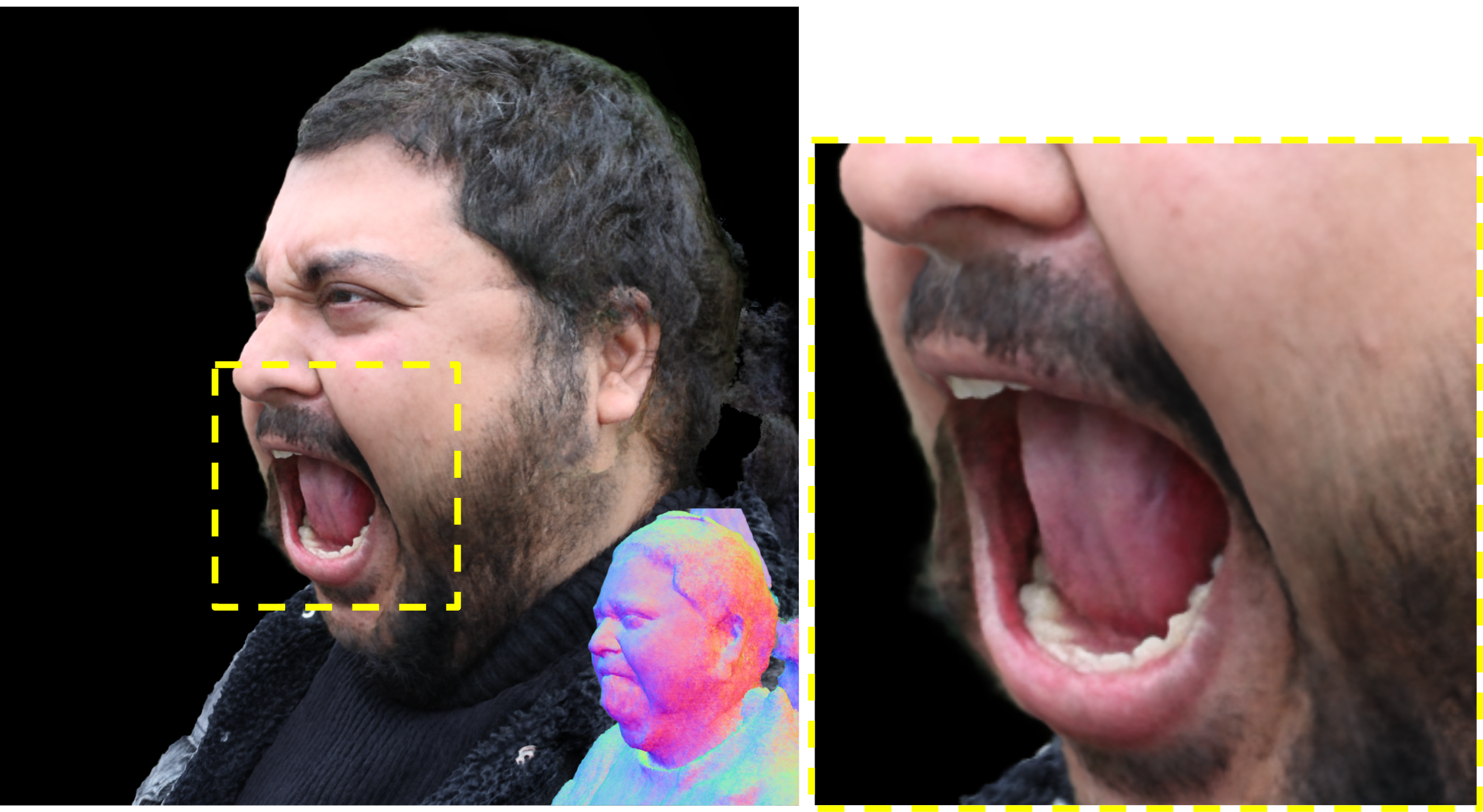}  &
      \includegraphics[width=\width]{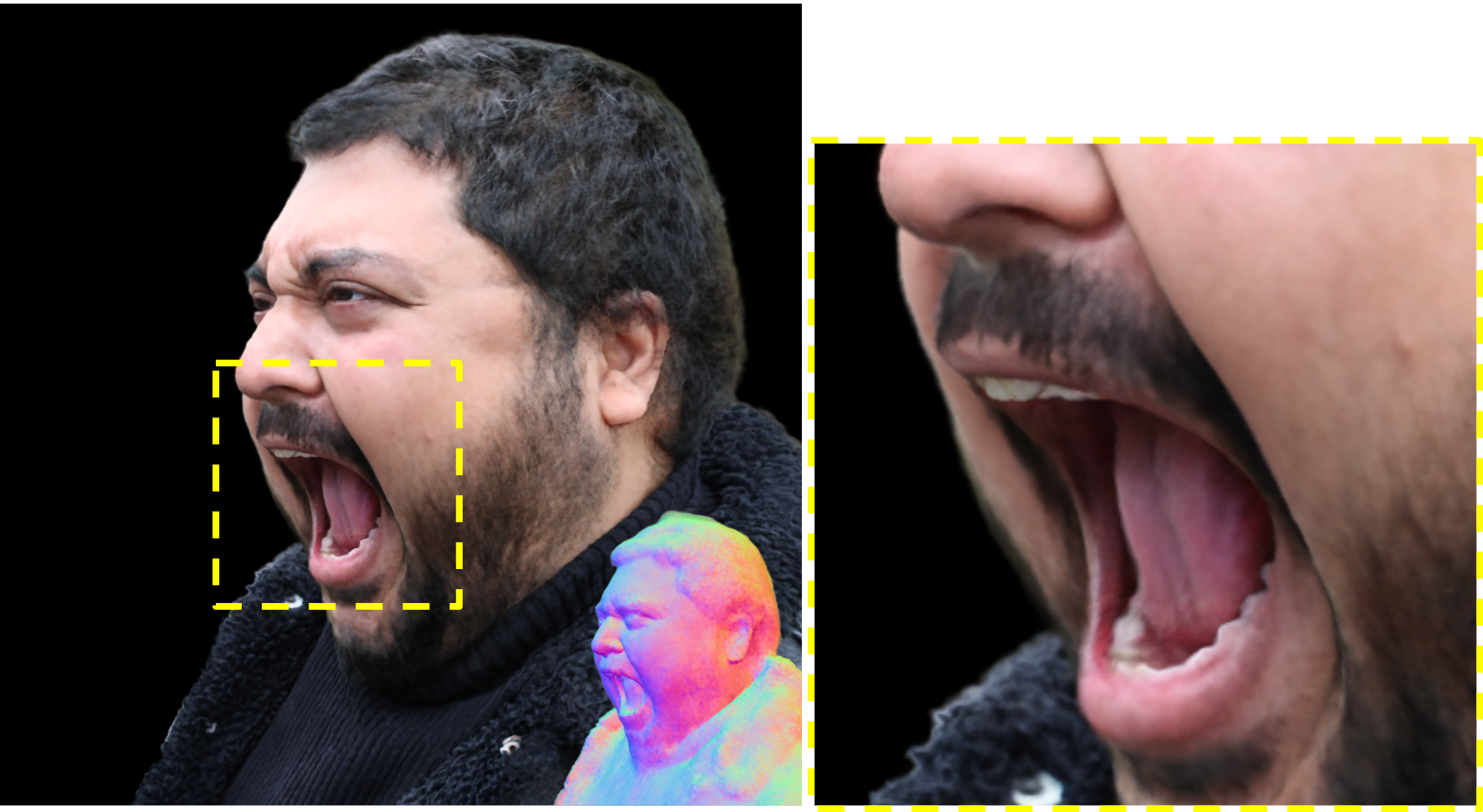} \\
  \includegraphics[width=\inputwidth]{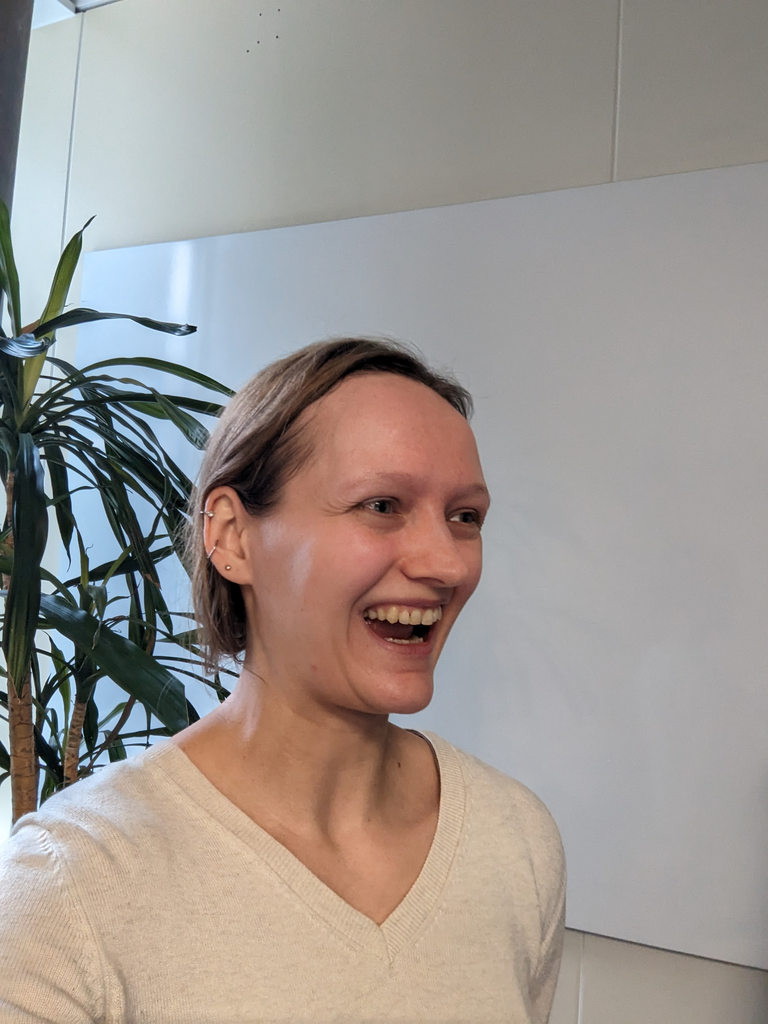}  
  \includegraphics[width=\inputwidth]{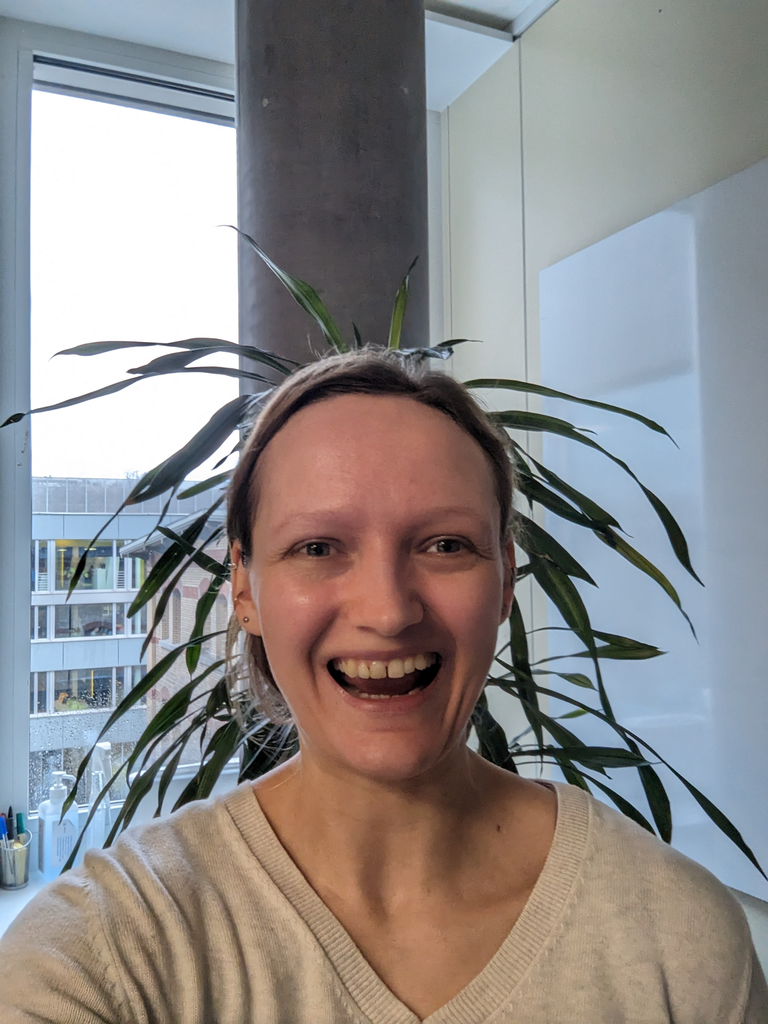}  
  \includegraphics[width=\inputwidth]{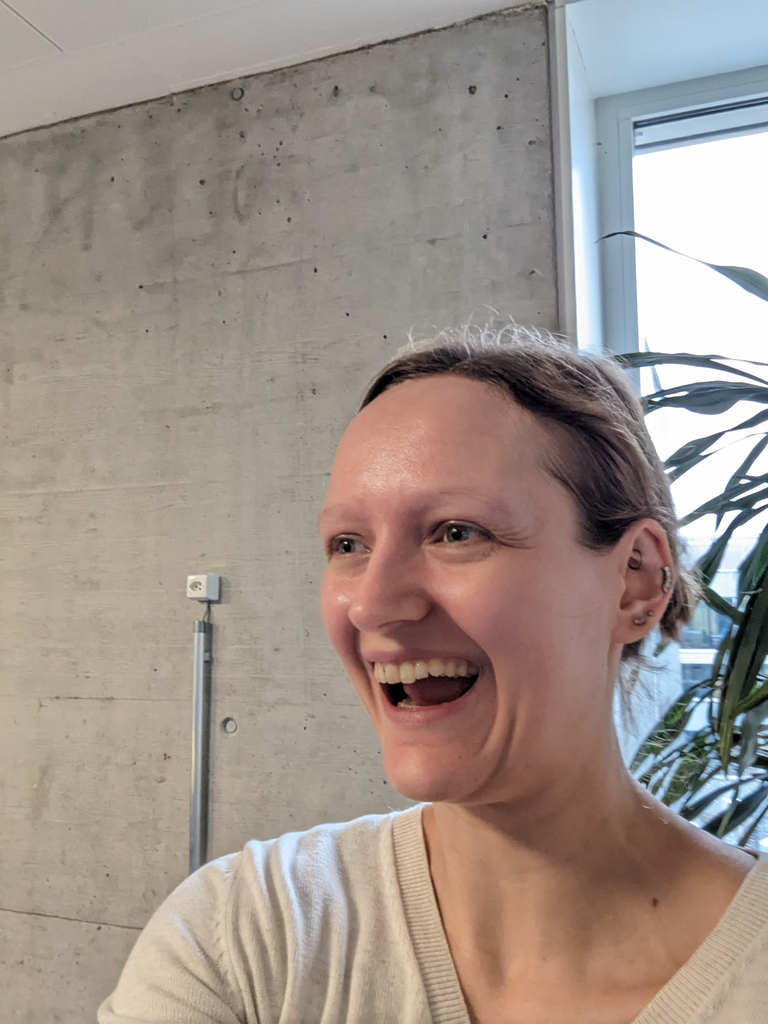}  &
      \includegraphics[width=\width]{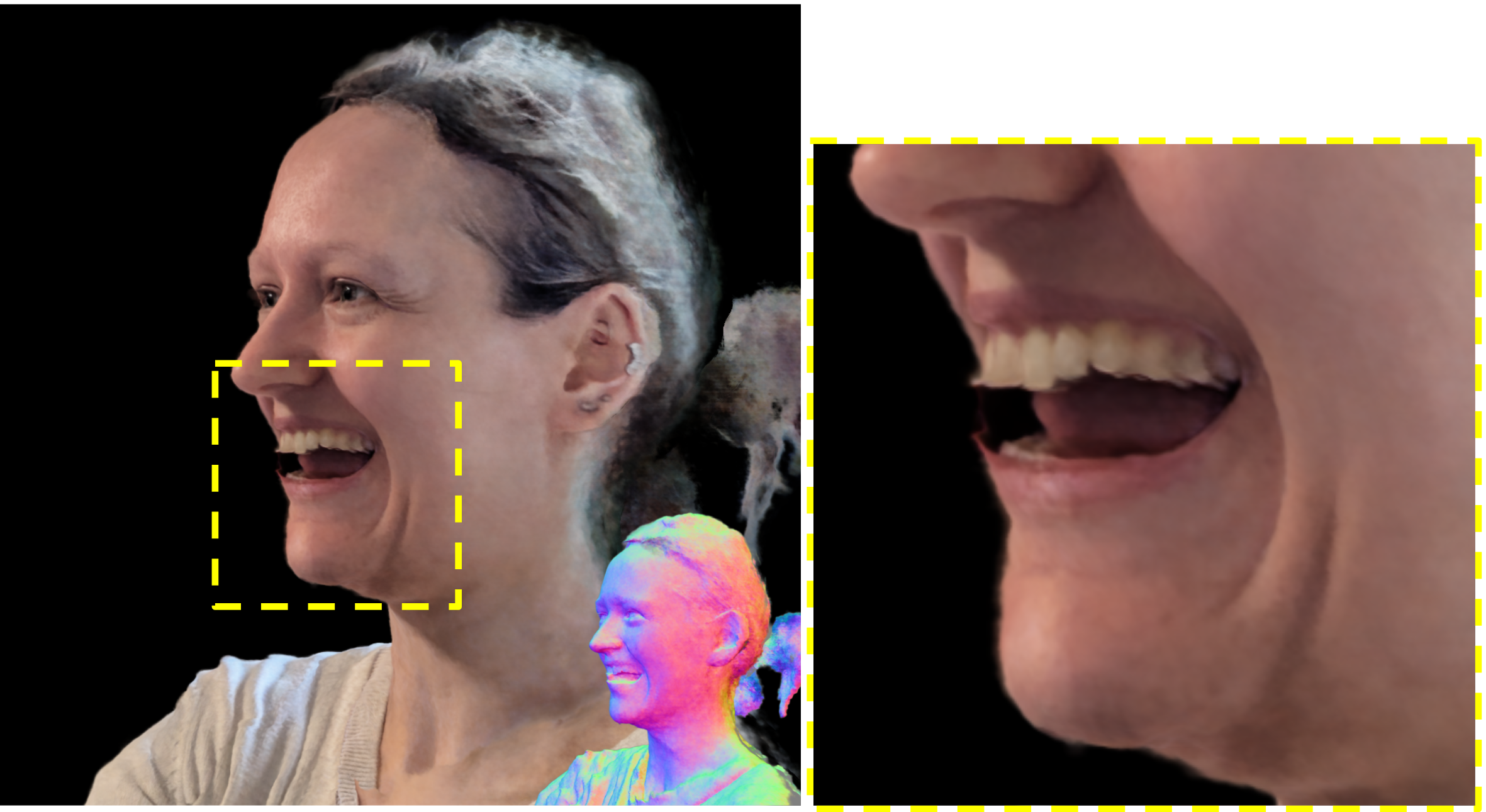}  &
      \includegraphics[width=\width]{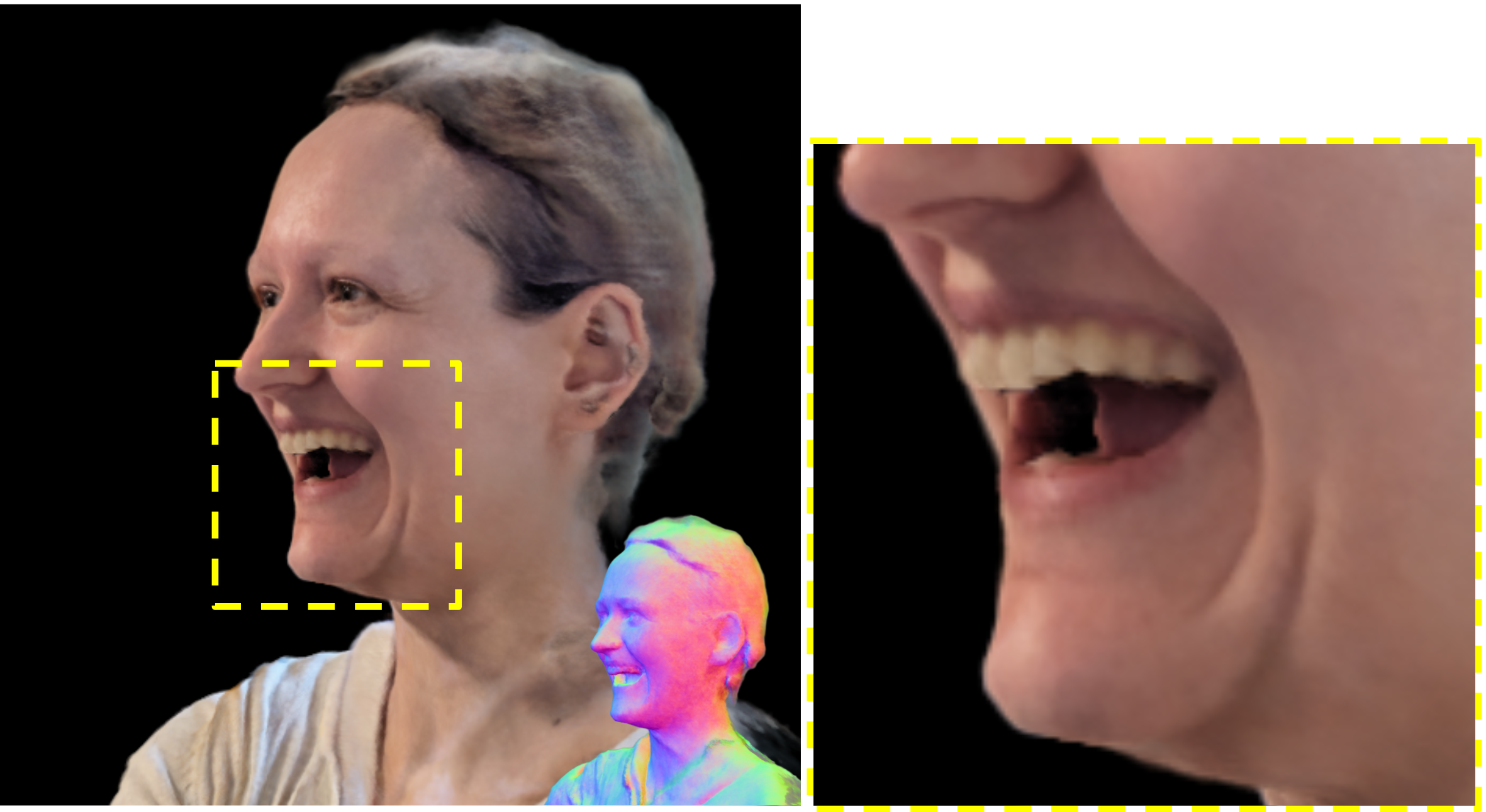} \\
Input & Preface & \textbf{Ours}
\end{tabular}
\end{center}
\caption{\label{fig:comparison_itw}We highlight a typical failure case for the best-performing related work. Preface struggles \cite{buhler2023preface} with strong facial expressions, in particular in the mouth and chin region. Please see the supplementary HTML page for video results.
}
\end{figure*}
\begin{figure*}[ht]

\begin{center}
\small
\setlength{\tabcolsep}{2pt}

\newcommand{\width}{2.8cm}
\newcommand{\gtwidth}{1cm}
\begin{tabular}{ccccccc}
  \includegraphics[width=\gtwidth]{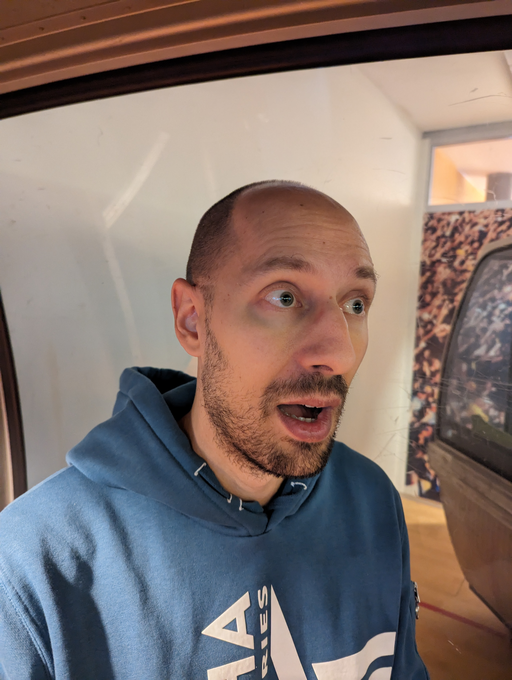}
      \includegraphics[width=\gtwidth]{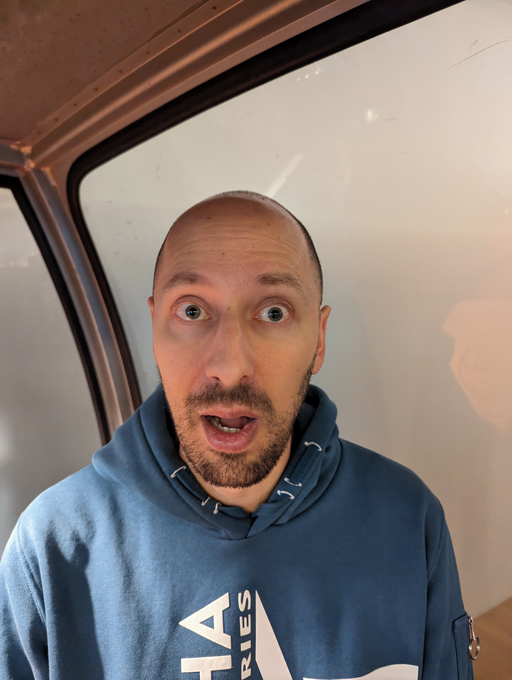}
      \includegraphics[width=\gtwidth]{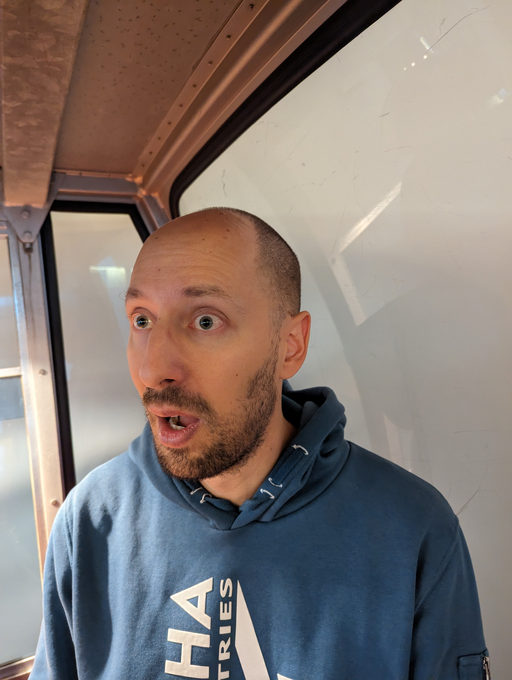}&
  \includegraphics[width=\width]        {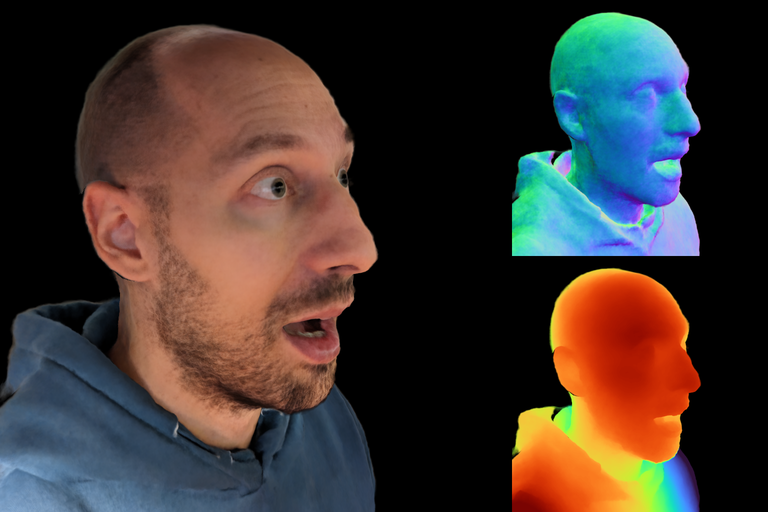}  &
  \includegraphics[width=\width]{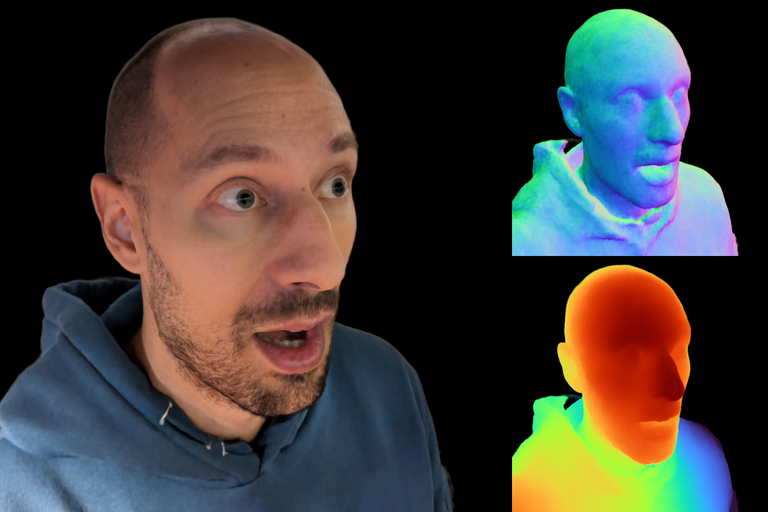} &
  \includegraphics[width=\width]{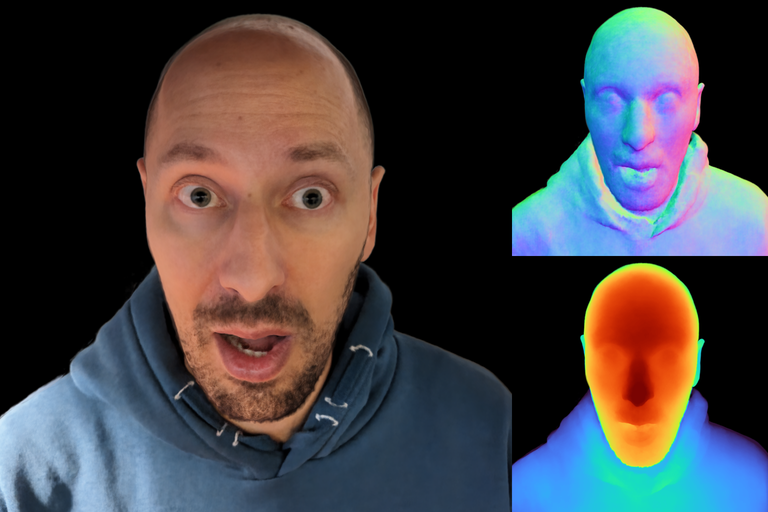}  &
  \includegraphics[width=\width]{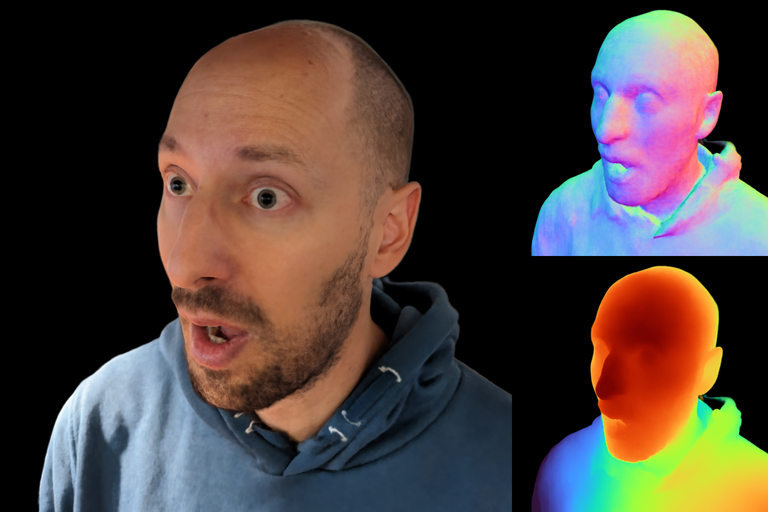}  &\includegraphics[width=\width]{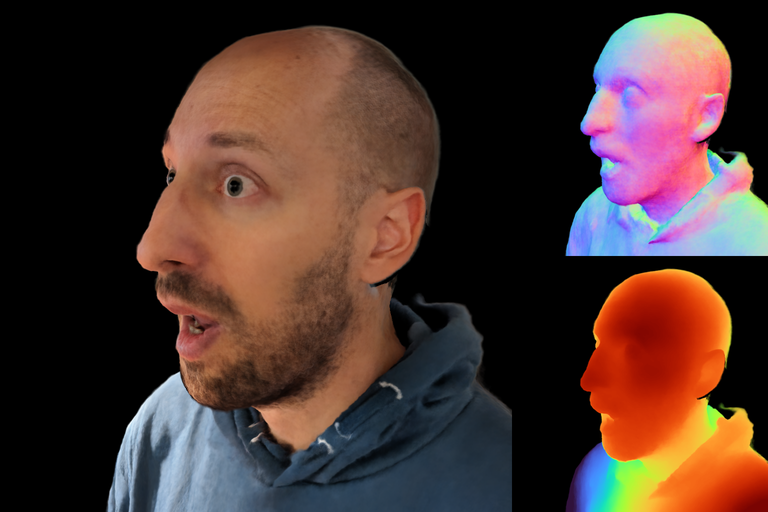}   \\
  \includegraphics[width=\gtwidth]{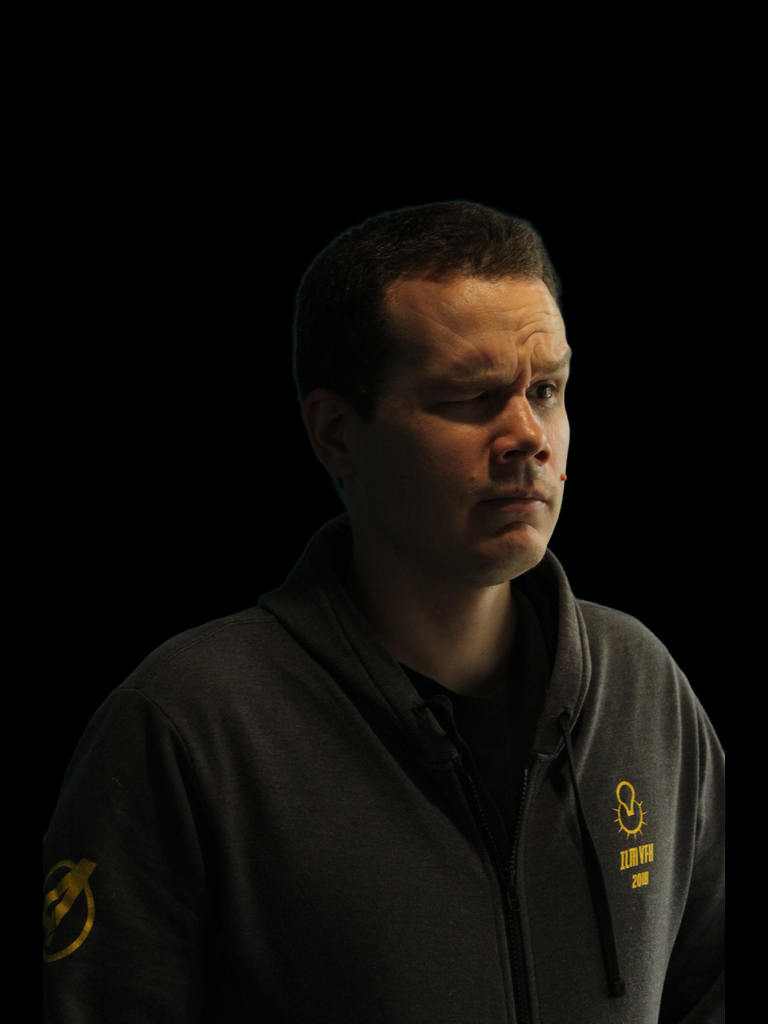}
      \includegraphics[width=\gtwidth]{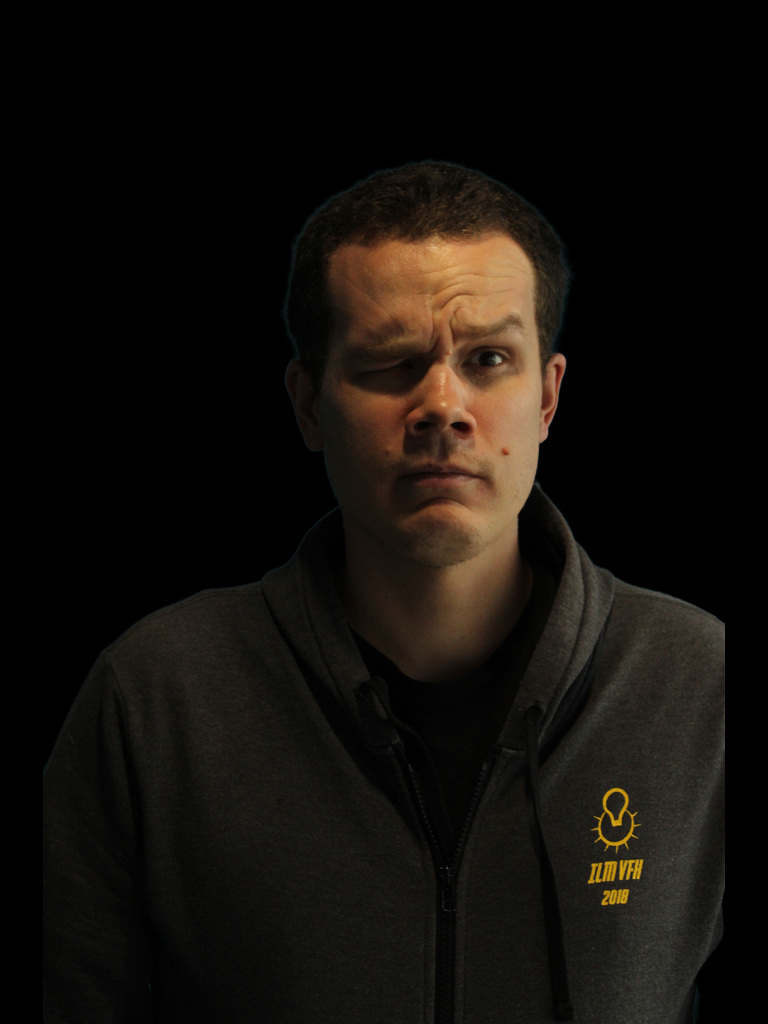}
      \includegraphics[width=\gtwidth]{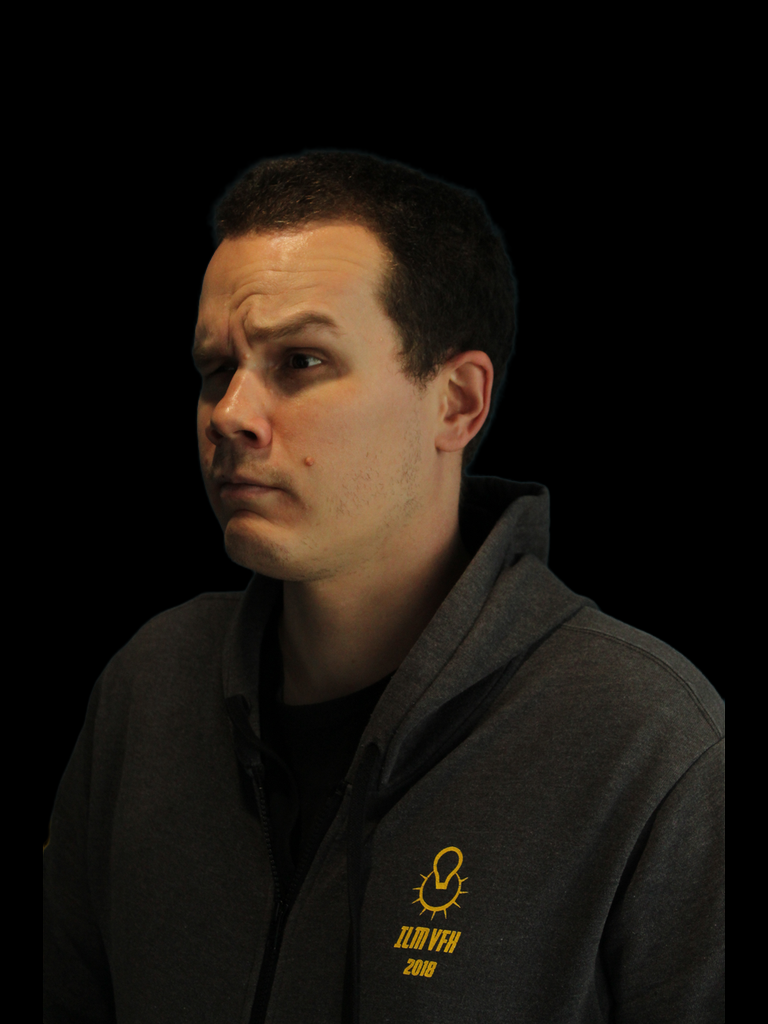}&
  \includegraphics[width=\width]        {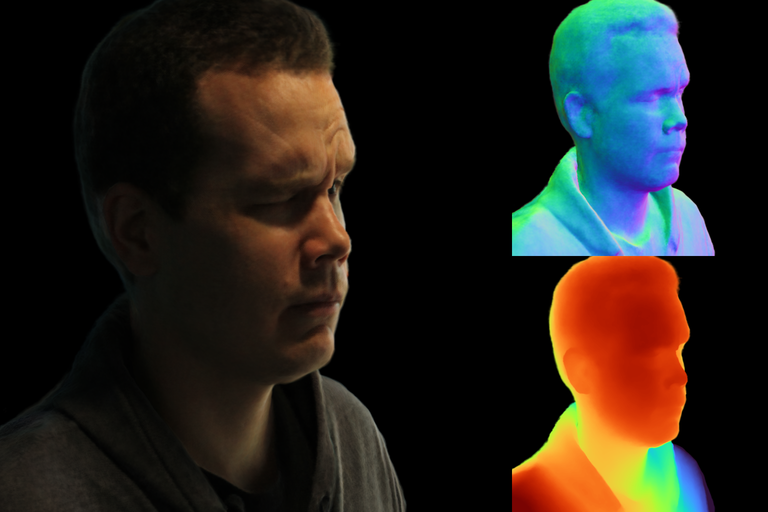}  &
  \includegraphics[width=\width]{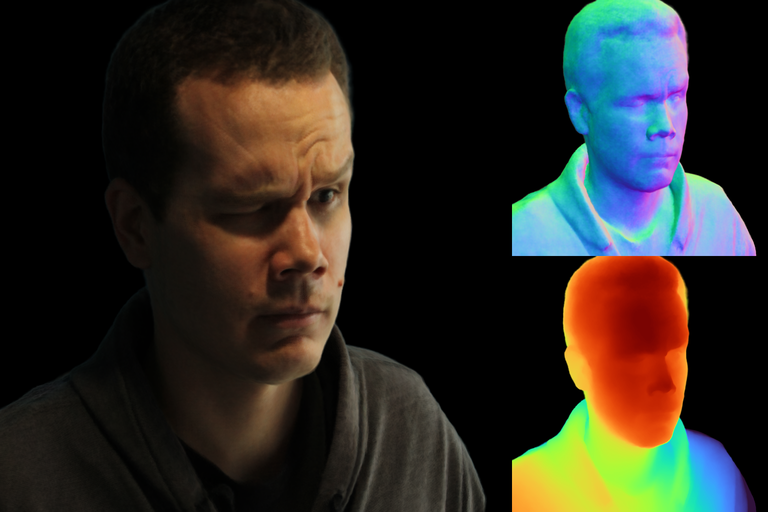} &
  \includegraphics[width=\width]{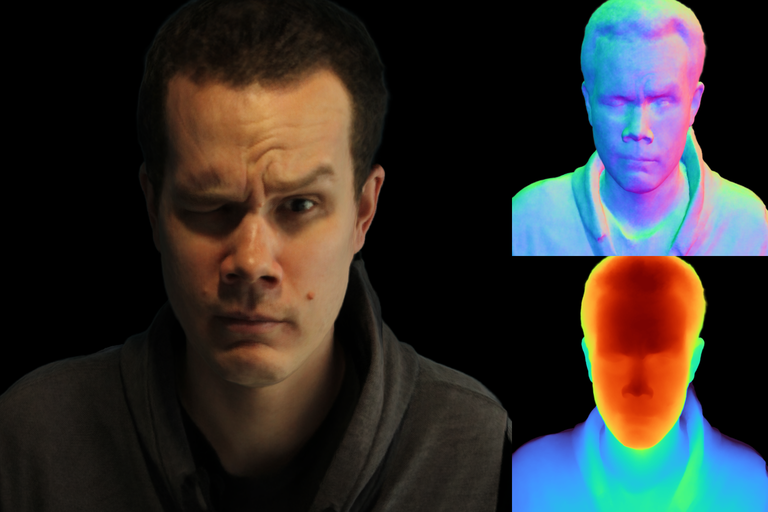}  &
  \includegraphics[width=\width]{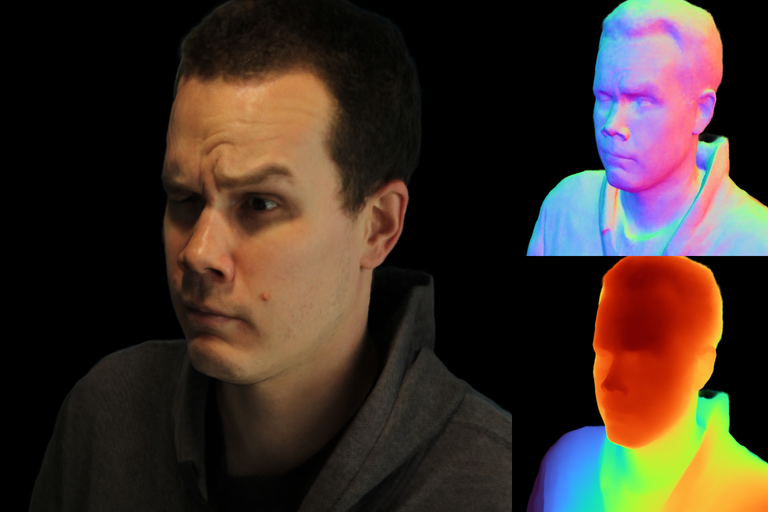}  &\includegraphics[width=\width]{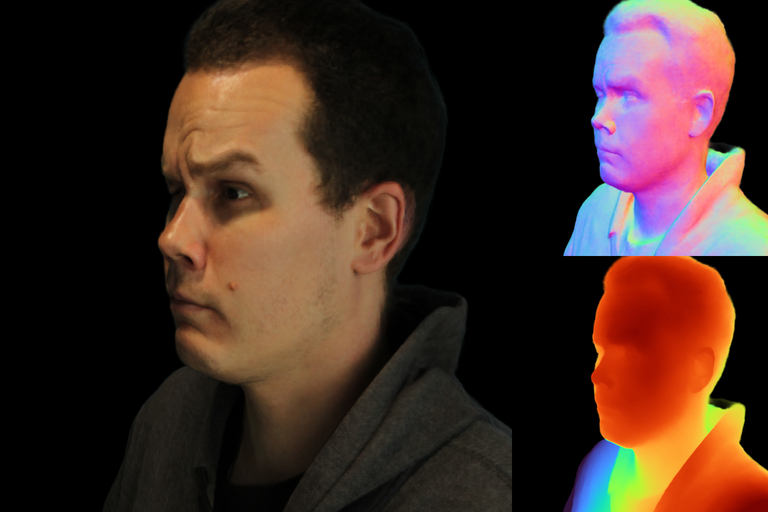}   \\
  \includegraphics[width=\gtwidth]{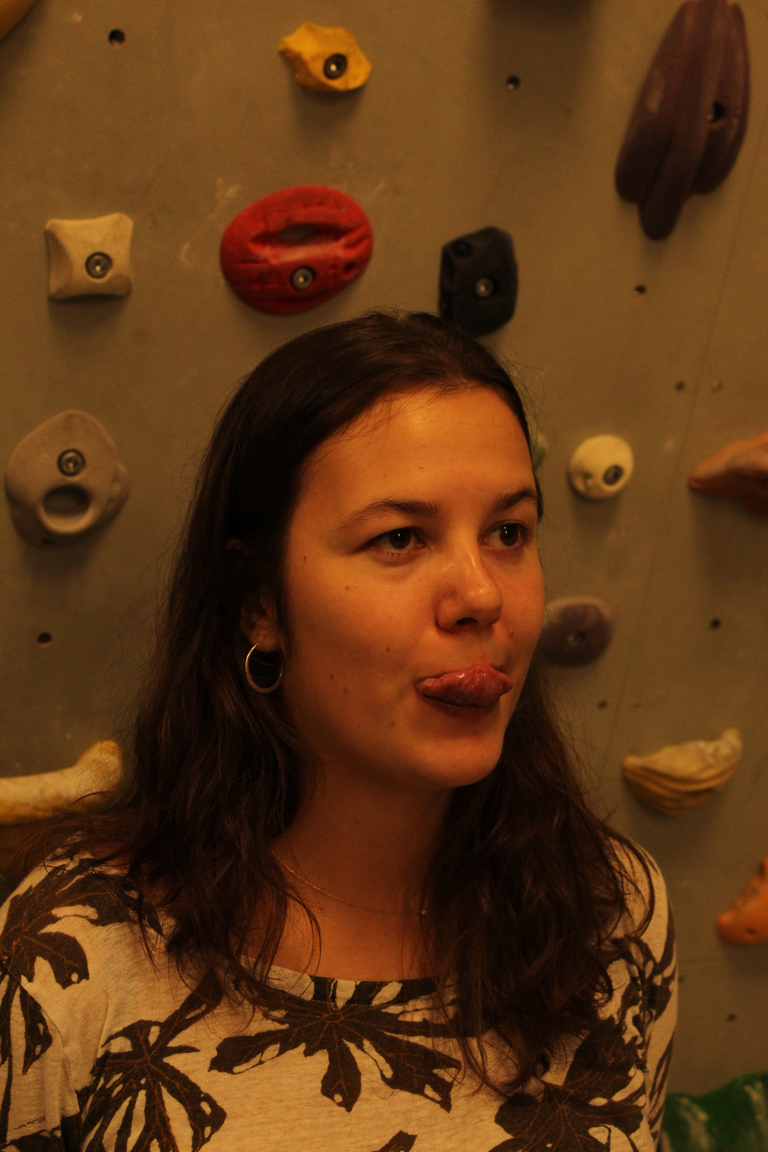}
      \includegraphics[width=\gtwidth]{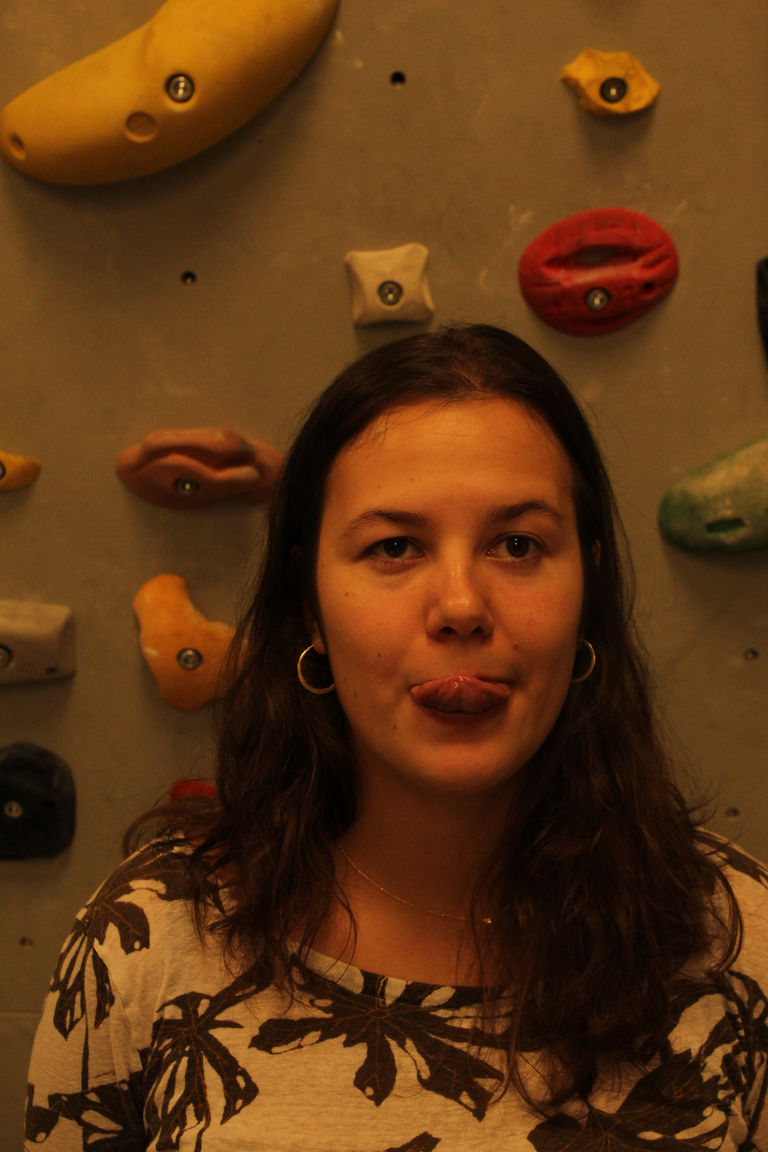}
      \includegraphics[width=\gtwidth]{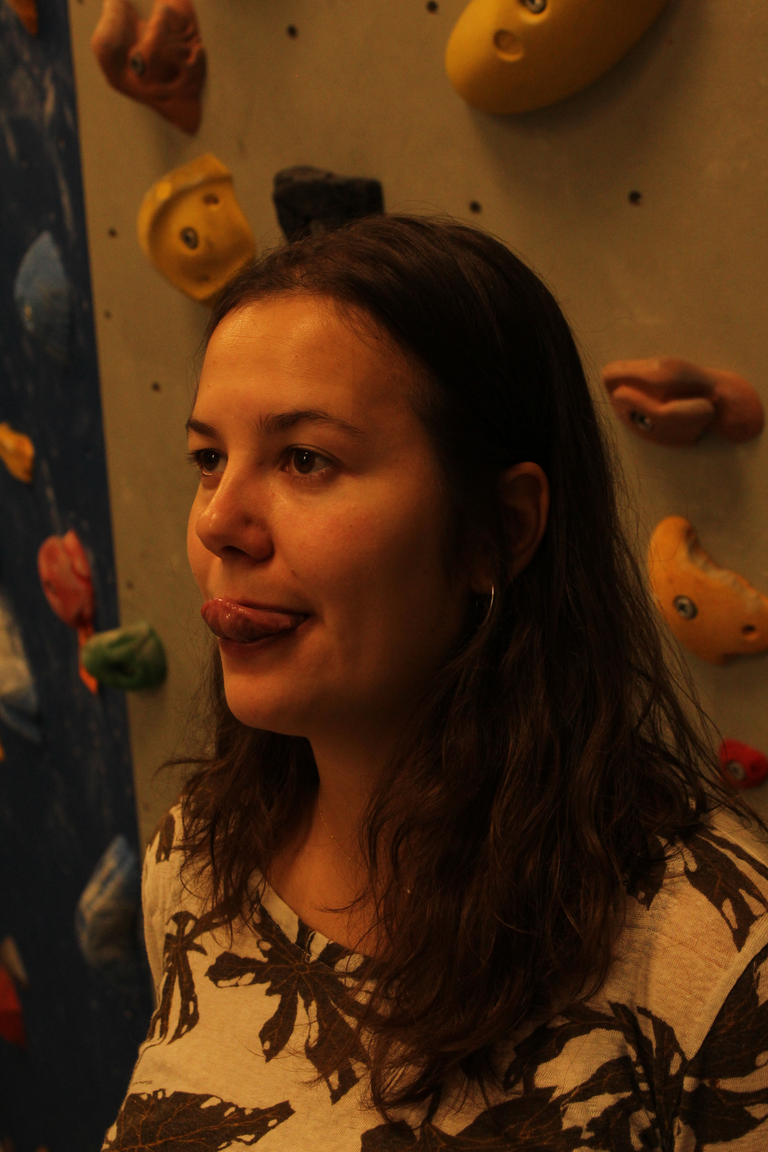}&
  \includegraphics[width=\width]        {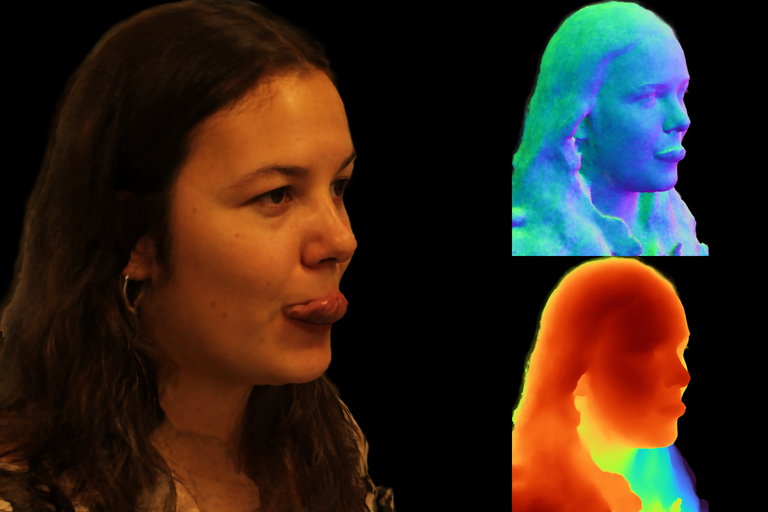}  &
  \includegraphics[width=\width]{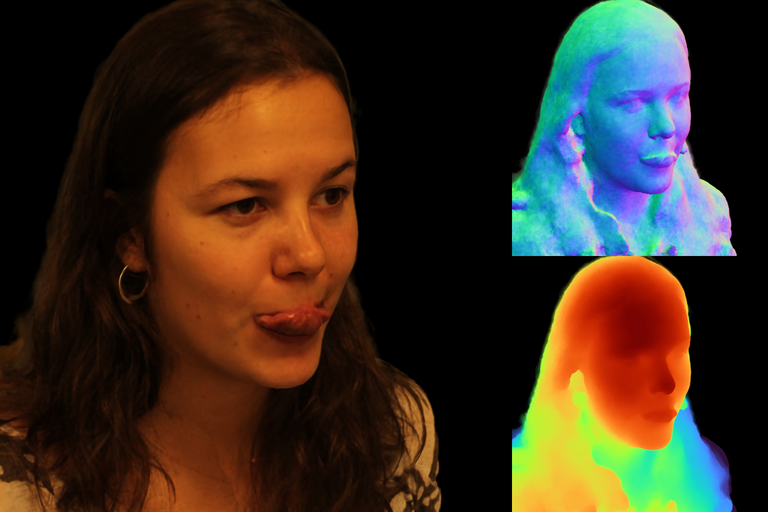} &
  \includegraphics[width=\width]{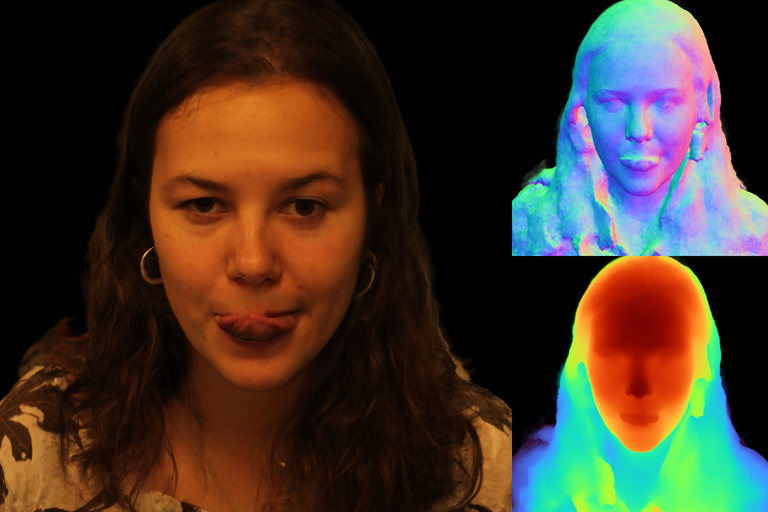}  &
  \includegraphics[width=\width]{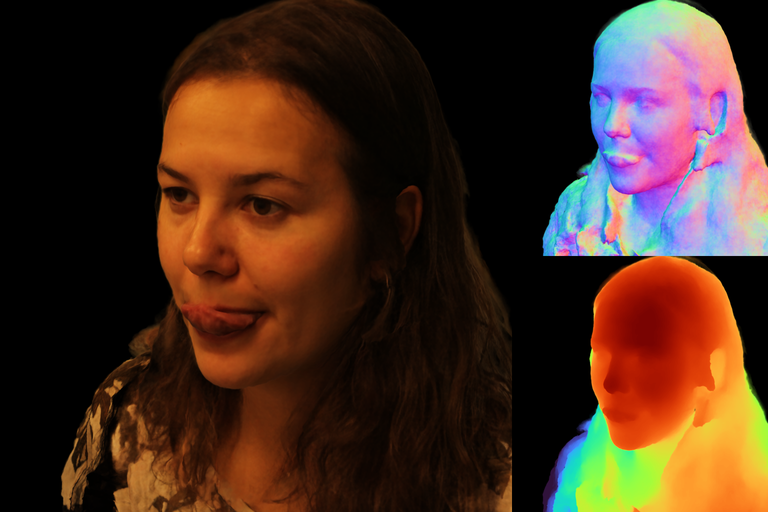}  &\includegraphics[width=\width]{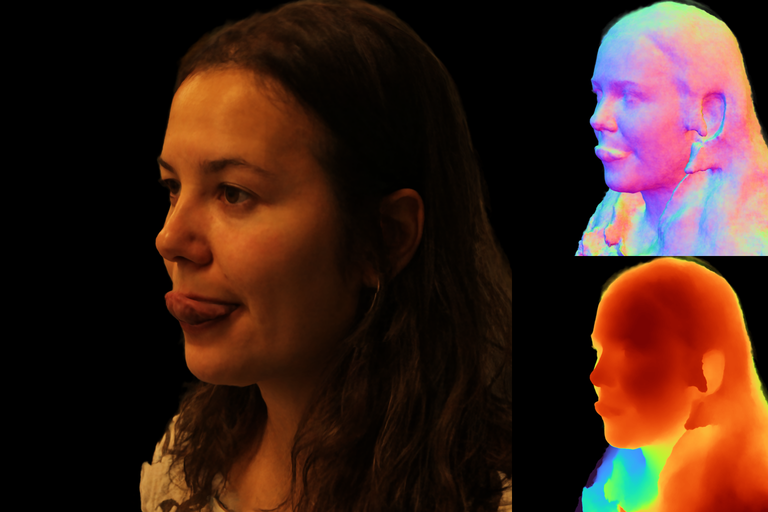}   \\
  \includegraphics[width=\gtwidth]{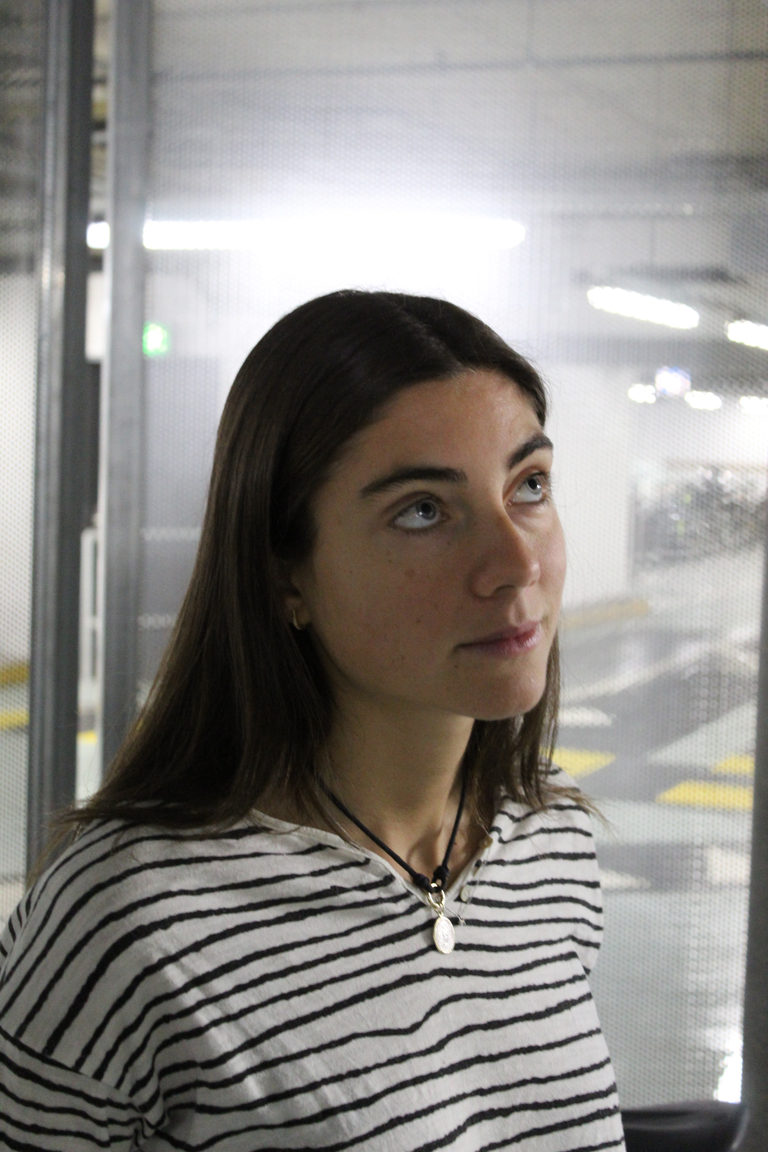}
      \includegraphics[width=\gtwidth]{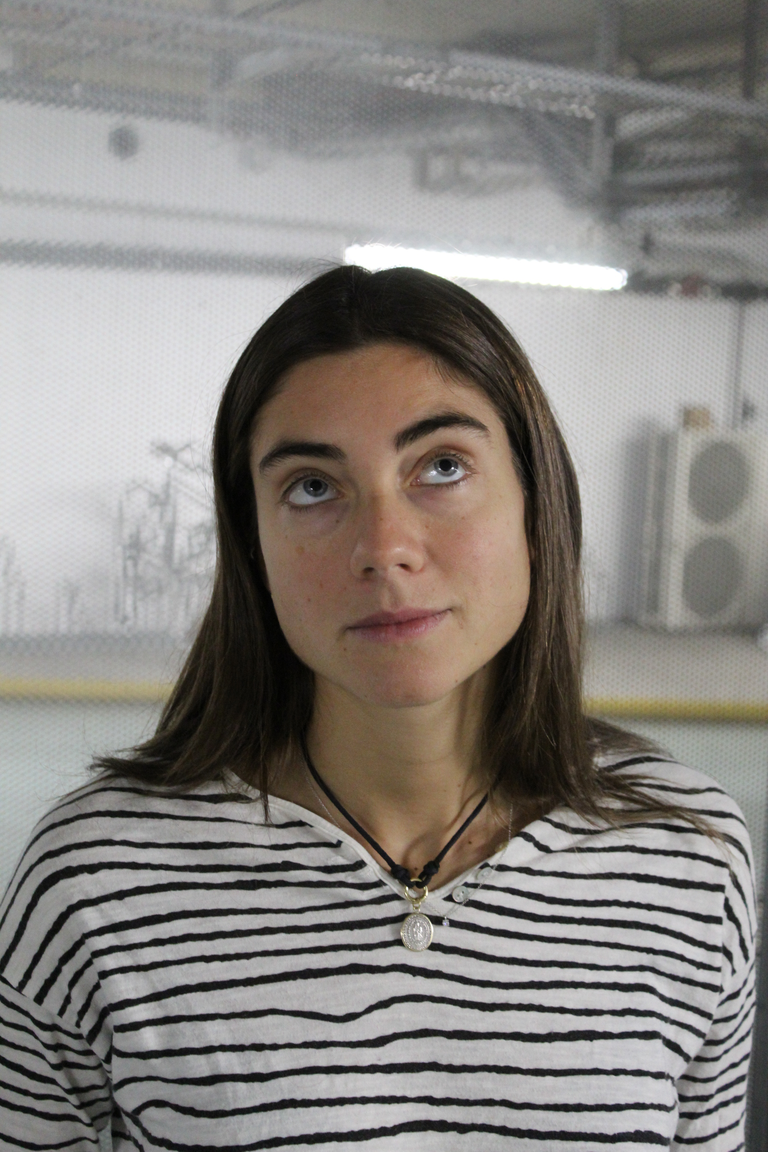}
      \includegraphics[width=\gtwidth]{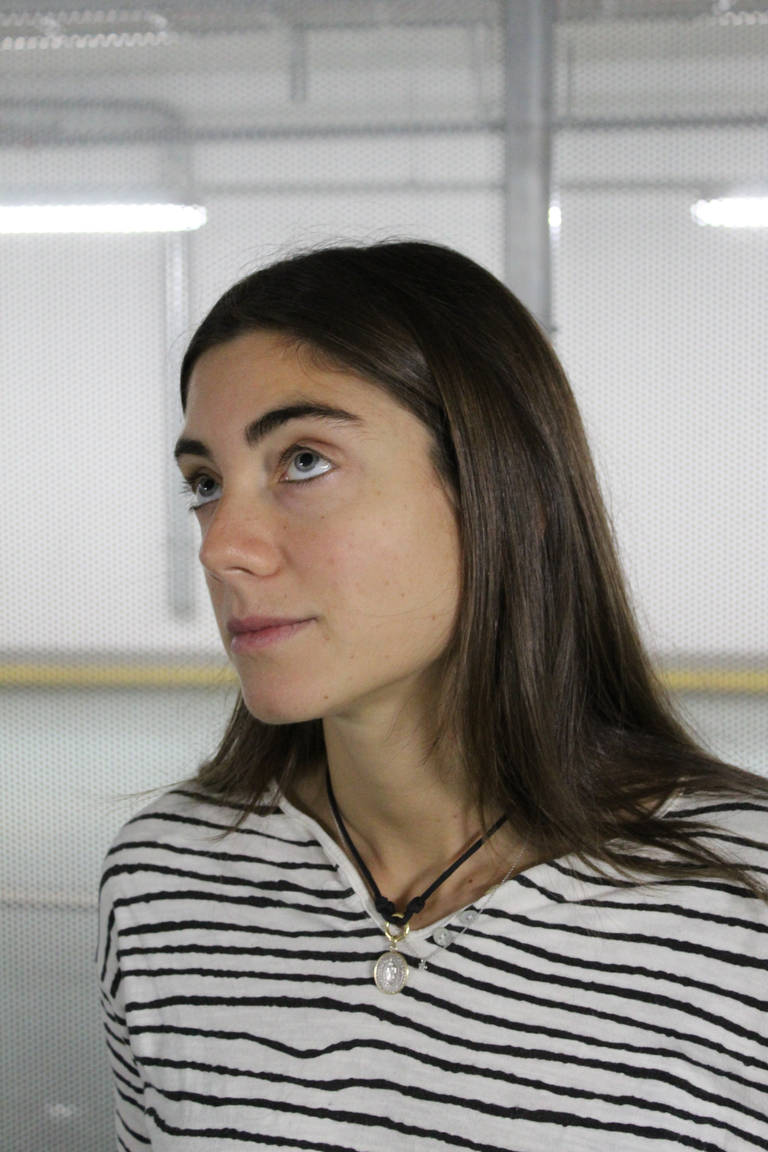}&
  \includegraphics[width=\width]        {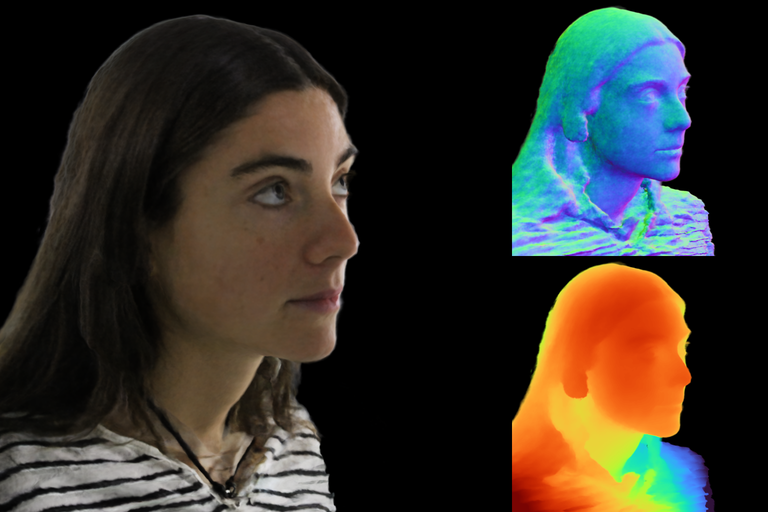}  &
  \includegraphics[width=\width]{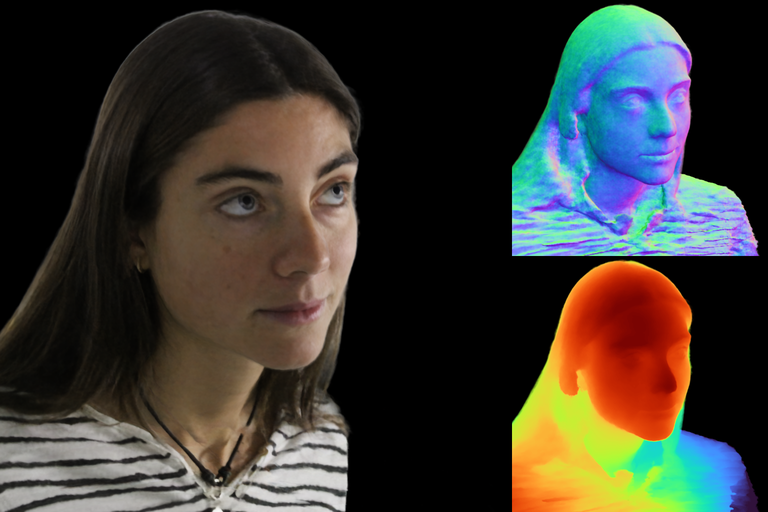} &
  \includegraphics[width=\width]{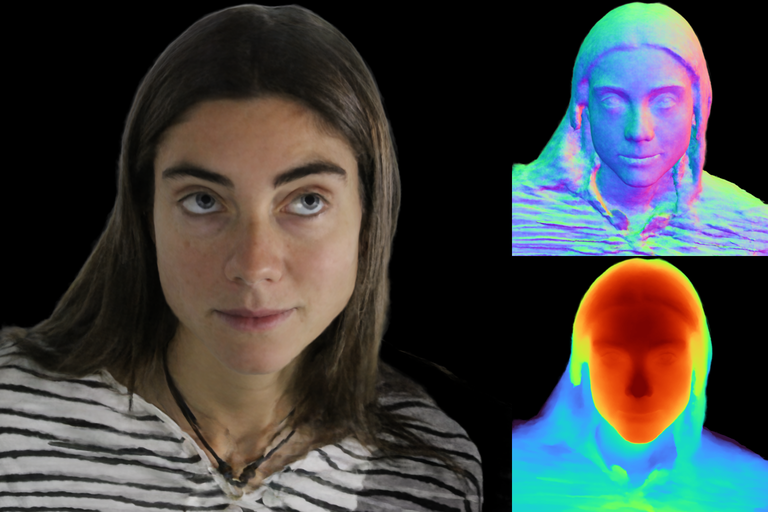}  &
  \includegraphics[width=\width]{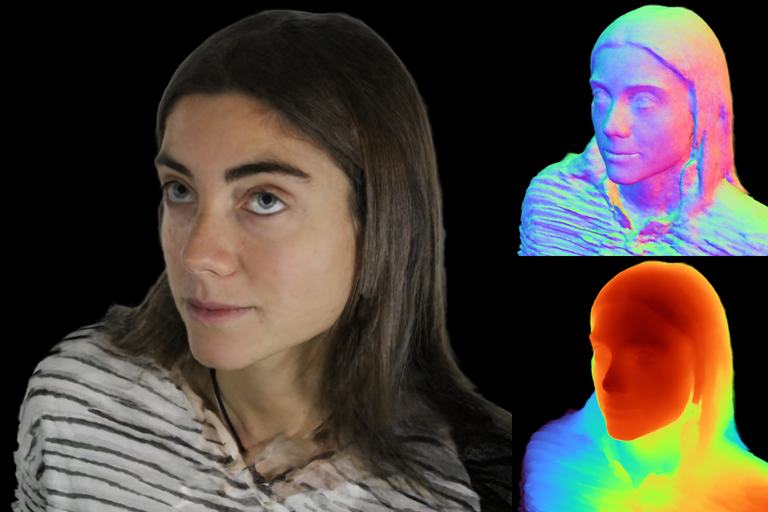}  &\includegraphics[width=\width]{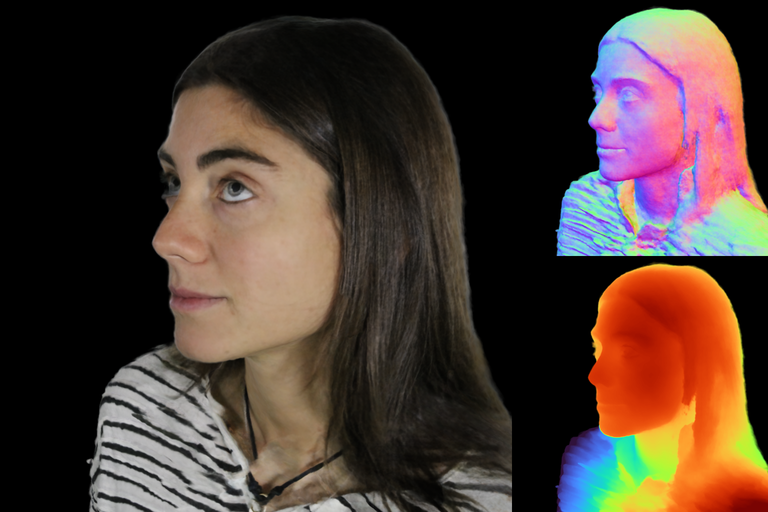}   \\
  \includegraphics[width=\gtwidth]{figures/inthewild/gt_240117_000_000_C01.jpg}
      \includegraphics[width=\gtwidth]{figures/inthewild/gt_240117_000_000_C00.jpg}
      \includegraphics[width=\gtwidth]{figures/inthewild/gt_240117_000_000_C02.jpg}&
  \includegraphics[width=\width]        {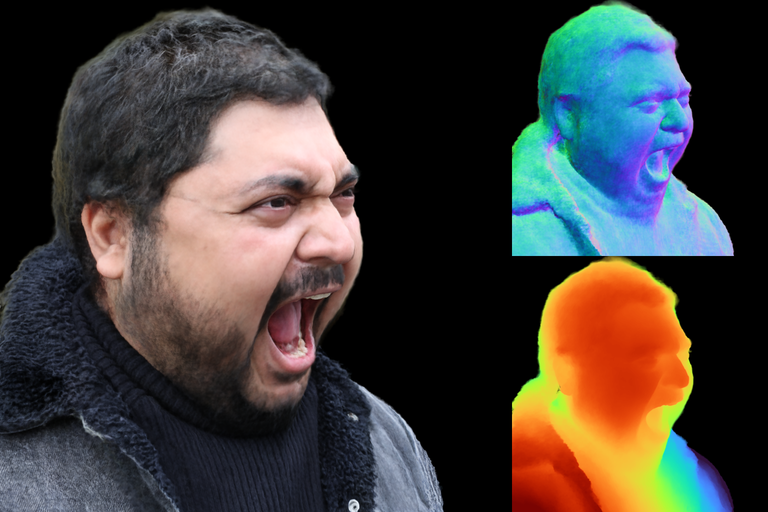}  &
  \includegraphics[width=\width]{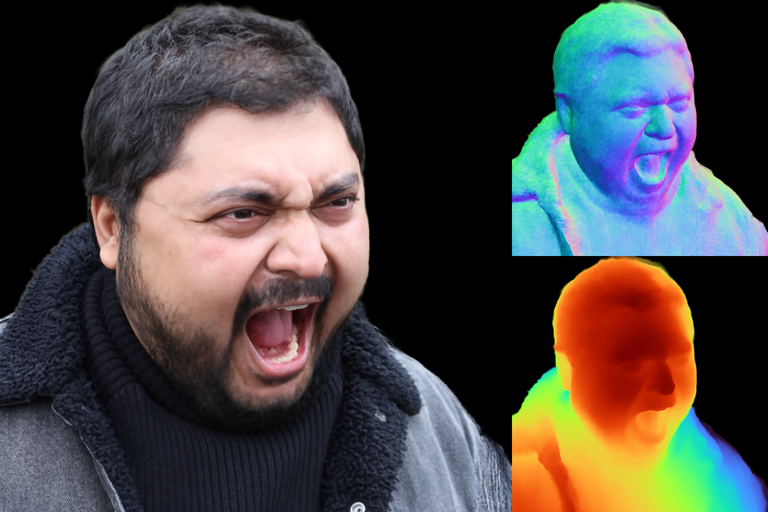} &
  \includegraphics[width=\width]{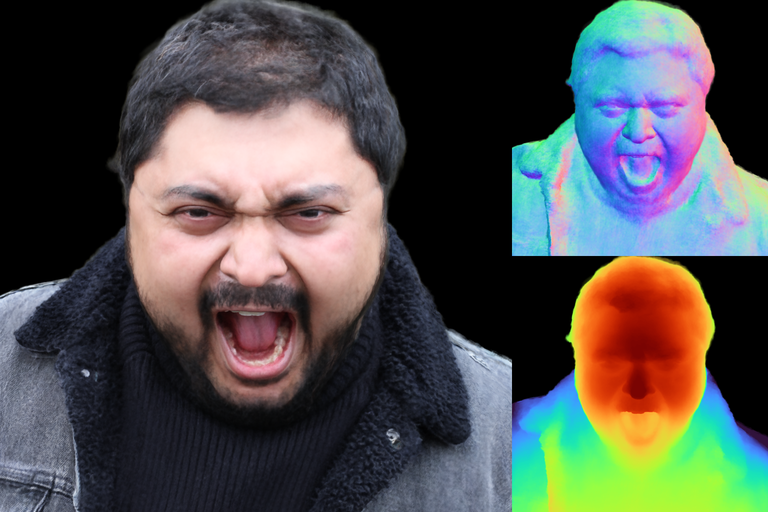}  &
  \includegraphics[width=\width]{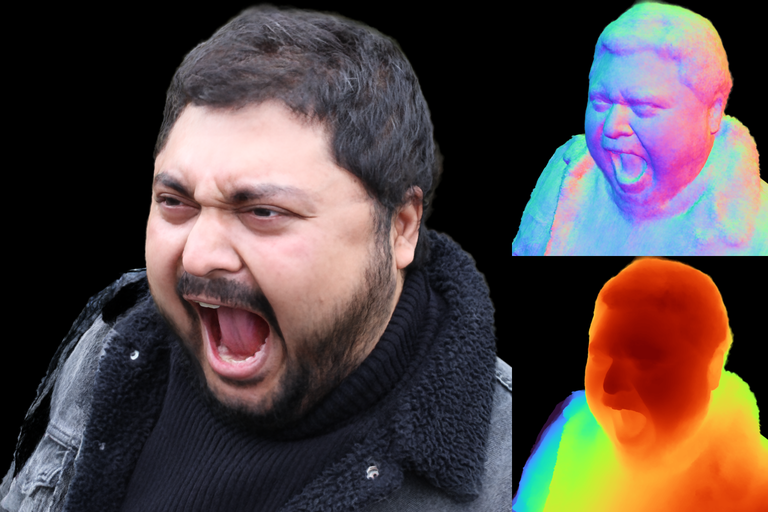}  &\includegraphics[width=\width]{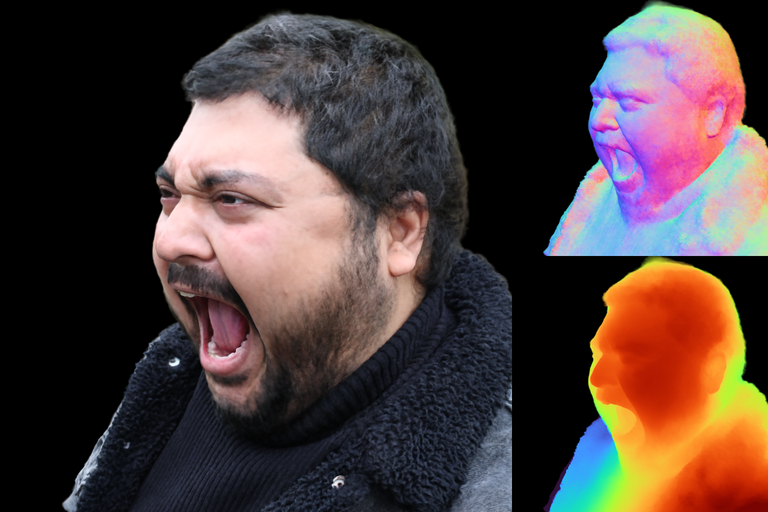}   \\
  
Input & \multicolumn{5}{c}{Novel views with world-space normals (top) and depth (bottom)}\\
\end{tabular}
\end{center}
\caption{\label{fig:inthewild} In-the-wild results. Given three in-the-wild images of a subject, our method reconstructs the subject and renders novel views with high resolution. The figure shows the input images (which can be taken sequentially), high-resolution rendering results in novel view, and also the reconstructed normal and depth maps. Our model generalizes to different challenging in-the-wild lighting. These results include indoor, outdoor, and dim scenes. Our method also captures strong idiosyncratic facial expressions such as the tongue-out case in row 3. This is an expression not included in the synthetic training data.}
\end{figure*}

\clearpage
\appendix
\section{Supplementary}
This supplementary document provides more details about the synthetic dataset (Sec. \ref{sec:supp_dataset}), the models and method (Sec. \ref{sec:supp_method}), experiments (Sec. \ref{sec:supp_exp}), and supplementary results (Sec. \ref{sec:supp_results}).

\subsection{Dataset Details}
\label{sec:supp_dataset}
Our synthetic dataset is generated in a similar fashion to previous work~\cite{wood2021fake}.
We use Blender to both generate and render realistic and diverse 3D scenes containing a face.
We sample faces from a large dataset of $50,000$ 3D scans that have been registered with a common template mesh~\cite{10.1111:cgf.15038}.
Then, we make each face mesh look realistic by applying a physically-based skin material from a high-resolution skin texture collection.
Next, we ``dress up'' our face by procedurally attaching traditional CG assets\footnote{\url{https://www.3dscanstore.com/}}.
Our total asset library contains the following items: $23$ upper-body garments (jackets, sweaters, suits, uniforms, coats, t-shirts, scarfs), $8$ types of eye-glasses, $157$ eyebrows, $6$ types of makeup, $2$ sets of eyelashes (with and without mascara), $1,679$ head textures, $2$ headwear items, $125$ strand-based hair / beard styles, and $6$ eye textures.
Each asset is authored as a rig using Blender Geometry Nodes, and each class of item executes class-specific rig logic to robustly attach itself to arbitrary 3D face meshes, regardless of identity or expression.
For example, upper-body clothing items non-rigidly deform themselves to fit around the neck of a target face, but eyeglasses are posed using inverse kinematics to rest on the nose-bridge and ears.

Finally, we render the scenes with Cycles: a physically-based ray-tracing renderer.
We render all faces in the same environment: a uniform well-lit environment.
Each face is rendered from $30$ distinct viewpoints, generated by placing the camera at random points around the head and pointing it at the face.
We choose random camera positions by sampling spherical coordinates: azimuthal angle, elevation angle, and radius.
We avoid overly similar viewpoints by discarding those with viewing directions closer than $25$ degrees to a previous one and re-sampling.
We found it helpful to sample random viewpoints for each subject and expression rather than sampling the same viewpoints across the entire dataset. The latter led to more floaters when training the prior model.

For the ablation with diverse environments, we sample random environment maps from the Laval indoor dataset \cite{hold2019deep,gardner2017laval}.

\subsection{Method Details}
\label{sec:supp_method}
\subsubsection{Prior Model Details}
\paragraph{Architecture}
The prior model architecture largely follows Preface \cite{buhler2023preface}. The prior model has a \emph{proposal} MLP predicting density and normals, and a \emph{NeRF} MLP predicting density and color \cite{mipnerf360}. The proposal MLP has depth $4$ and layers width $(256 + |\boldsymbol{\beta}| + |\boldsymbol{\psi}| + |\boldsymbol{w}|) \times 256$ parameters, where the $\boldsymbol{\beta} \in \mathbb{R}^{48}$ are the 3DMM identity coefficients, $\boldsymbol{\psi} \in \mathbb{R}^{157}$ are the 3DMM expression coefficients, and $\boldsymbol{w} \in \mathbb{R}^{64}$ is the optimizable latent identity code. The NeRF MLP trunk has depth $8$ and layers width $(1024 + |\boldsymbol{\beta}| + |\boldsymbol{\psi}| + |\boldsymbol{w}|) \times 1024$. The NeRF MLP trunk features are projected to 3 dimensions and normalized to predict negative normals. The color is predicted from the NeRF MLP trunk features after a bottleneck with width 256 and a single view-conditioned layer with width 128. We optimize one $\boldsymbol{w}$ code per identity, which yields a codebook of size $1500 \times 64$. The total parameter count is 32 Mio.
The weights are initialized with He Uniform Variance Scaling \cite{he2015delving}.

\paragraph{Training} We train the prior model on images of 1,500 synthetic identities, rendering each with 13 expressions and 30 views. In total, we train the full prior model for 1 Mio. steps on 64 TPUs with a batch size of
$131,072$ rays per step ($256$ identities $\times$ $8$ view $\times$ $64$ pixels), which takes about 10 days. However, we observe that the model already reaches near convergence after 105,000 steps. In particular, fine-tuning that uses the model trained for 105,000 steps achieves very similar photo-metric quality as fine-tuning the model trained for 1 Mio. steps, please see our ablation study in Sec. \ref{sec:ablation} and the supp. HTML page for visuals.

We train our model with 128 samples for the proposal and 128 samples for the NeRF MLP. The proposal MLP is sampled twice and the NeRF MLP is sampled once. Both the proposal and NeRF MLP use the same positional encoding for the inputs. The xyz inputs use twelve levels; the view direction four levels and appends the view direction without positional encoding.

\paragraph{Losses}
The terms in the loss function $\mathcal{L}_\text{prior} = \mathcal{L}_\text{recon} + \lambda_\text{prop} \mathcal{L}_\text{prop}$ are a combination of the mean absolute photo-metric error of the ground truth color $\boldsymbol{c}$ versus the predicted color $\boldsymbol{\hat c}$: $\mathcal{L}_\text{recon} = \lVert \boldsymbol{c} - \boldsymbol{\hat c} \rVert_1$, and the MipNeRF360 proposal loss $\mathcal{L}_\text{prop} = \sum_i \frac{1}{w_i} \max (0, w_i - \text{bound}(\hat t, \boldsymbol{\hat w}, T_i))^2$. Please see Eq. 13 in the original paper for more details \cite{mipnerf360}).

The optimization employs Adam~\cite{KingmB2015} with $\beta_1 = 0.9, \beta_2=0.999$. The learning rate decays exponentially from $0.002$ to $0.00002$. We clip gradients with norms larger than 0.001.

We also experimented with the supervision of accumulation and depth but didn't find an improvement.

\subsection{Inference Details}

\paragraph{3DMM Fitting}
We fit the 3DMM \cite{blanz1999morphable} from 599 2D probabilistic landmarks, where each landmark is the projection of one vertex of the 3DMM mesh with an uncertainty $\sigma$.
The landmark fitting follows \cite{wood20223d} and minimizes the energy function 
\[ E(\boldsymbol{\Phi}; L) = E_\text{landmarks} + E_\text{identity} + E_\text{expression} + E_\text{joints} + E_\text{intersect}\text{,} \] where $L$ denotes the 599 probabilistic landmarks and $\boldsymbol{\Phi}$ all the optimized 3DMM parameters including identity, expressions, joint rotations, and global translation, and intrinsic and extrinsic camera parameters, if unknown.
The landmark term \[ E_\text{landmarks} = \sum_{j,k}^{C,|L|} \frac{\lVert \boldsymbol x_{jk} - 
\boldsymbol \mu_{jk}\rVert^2}{2\sigma_{jk}^2} \] minimizes the distance between the projected landmarks from the 3DMM $\boldsymbol x_{jk}$ and the estimated probabilistic landmarks with mean $\boldsymbol{\mu}_{jk}$ and variance $\sigma_{jk}^2$ for all available camera views $C$.
The identity $E_\text{identity} = -\log(p(\boldsymbol{\beta}))$ is a regularizer that encourages plausible face shapes. It is computed as the negative log-likelihood given a fitted Gaussian Mixture Model of 3D head scans \cite{wood2021fake}.
The expression and joints terms $E_\text{expression} = \lVert \boldsymbol{\psi} \rVert^2$ and $E_\text{joints} = \lVert  \boldsymbol{J}_i \rVert^2$ are regularization terms to encourage small values in the expression code $\boldsymbol{\psi}$ and joint rotations $\boldsymbol{J}_i$ for the neck and the two eyeball joints. Note that the regularization does not apply to the global (head) joint. $E_\text{intersect}$ discourages intersecting vertices. Please see \cite{wood20223d} for more details.

\paragraph{Warm-up by Inversion}
We sample random patches of size $32 \times 32$. Our batch size is $4096$ rays so we use $4$ patches in each batch. We run 1,500 inversion steps, which takes about 10 minutes on four TPUs.

\paragraph{Optimization}
The model fitting uses the same sampling strategy as prior model training (2x proposal and 1x NeRF sampling) and employs the Adam optimizer with the same parameters.
We sample $4096$ random rays across all available views in each step. We optimize for 50,000 steps, which takes about 3.5 hours.

\paragraph{Losses}
The loss terms follow related works. In the following, the variable $w$ corresponds to the NeRF sample weight defined in Eq. 5 of the original NeRF paper \cite{mildenhall2020nerf}.
The normal consistency loss is defined as $\mathcal{L}_\text{normal} = \sum_i w_i \cdot (1 - \boldsymbol {\text{n}}^\top \boldsymbol{\hat {\text{n}}})$, where $\boldsymbol {\text{n}}$ are the analytical normals and $\boldsymbol{\hat {\text{n}}}$ are the predicted normals.
The regularization of the view direction weights is defined as 
$\mathcal{L}_{d} = \lVert \theta_v \rVert^2$, where $\theta_d$ are the model parameters that process the positionally encoded view directions. The distortion loss is defined as 
$\mathcal{L}_\text{dist} = \sum_{i,j} w_i w_j \lVert \frac{s_i + s_{i + 1}}{2} - \frac{s_j + s_{j + 1}}{2}\rVert + \frac{1}{3} \sum_i w_i^2 (s_{i+1} - s_i)$, where $s$ is the normalized ray distance. For details about the LPIPS loss $\mathcal{L}_\text{LPIPS}$, please refer to Eq. 1 in \cite{zhang2018perceptual}. 
We find that sampling individual random rays during model fitting yields better results than sampling patches. Hence, we only employ the perceptual loss $\mathcal{L}_\text{LPIPS}$ during warm-up but not during fine-tuning. 

\paragraph{Rendering Time}
We render on 4 TPUs, where rendering takes about 20.5 seconds per $1024\times1024$ frame.

\subsection{Experimental Details}
\label{sec:supp_exp}
\subsubsection{Multiface Dataset}
\begin{figure}[ht]
    \centering
  \includegraphics[width=0.7\columnwidth]{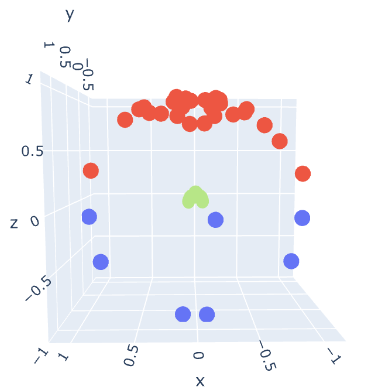}
    \caption{Multiface cameras used for evaluation are highlighted in red. Blue cameras are discarded. The face (green) is looking along the z axis.\label{fig:multiface_cameras}}
\end{figure}

We quantitatively compare and ablate on a subset of the Multiface dataset \cite{wuu2022multiface}. Metrics are computed on three expressions from three identities---nine scenes in total. For each scene, we select three views for training: One frontal and two side views. For evaluation, we compute metrics on all holdout views where the cameras are not located on the back of the head. Please see Fig. \ref{fig:multiface_cameras} for a visualization.

\subsubsection{Preface Dataset}
 Our ablation study that trains on real data uses the Preface dataset \cite{buhler2023preface}. It contains multi-view captures of 1,500 identities. Each identity is captured for 13 facial expressions and from 12 views. Please see \cite{buhler2023preface} for more details and a breakdown of demographics.

\subsection{Supplementary Results}
\label{sec:supp_results}
\subsubsection{In-the-wild Results}
We provide extensive results for high-resolution in-the-wild captures on the supplementary HTML page and video.

\subsubsection{Supplementary Ablations and Comparisons}

\begin{table}
\small 
\begin{center}
\begin{tabular}{c | ccc}
\textbf{\# Views} & \multicolumn{1}{c}{\textbf{PSNR} $\uparrow$} &
    \multicolumn{1}{c}{\textbf{SSIM} $\uparrow$} & 
    \multicolumn{1}{c}{\textbf{LPIPS} $\downarrow$}\\
\hline
1 & 23.79 & 0.7323 & 0.3139\\
2 & 24.29 & 0.7409 & 0.3184 \\
3 & 26.54 & 0.7750 & 0.3144
\end{tabular}
\end{center}
\caption{Number of Input Views. Our method can produce good-quality frontal views from a single frontal image. However, the quality suffers from side views, where the input image doesn't provide any information. Please see the supplementary HTML page for visuals.
\label{tbl:ablation_supp_view}}
\end{table}

\begin{table}[ht]
\caption{
\label{tbl:ablation_supp}Additional ablations. 
Training a prior model with a background leads to more floating artifacts during fine-tuning.
}
\small 
\begin{center}
\begin{tabular}{lccc|ccc}
\hline
\textbf{Variant} &
    \multicolumn{1}{c}{\textbf{PSNR} $\uparrow$} &
    \multicolumn{1}{c}{\textbf{SSIM} $\uparrow$} & 
    \multicolumn{1}{c}{\textbf{LPIPS} $\downarrow$}\\
\hline
Prior with background & 24.51 & 0.7313 & 0.3244 \\
Prior without background & 26.54 & 0.7750 & 0.3144
\end{tabular}
\end{center}
\end{table}

We extend our ablations from the main paper with metrics computed on different variants of our prior model in Tables \ref{tbl:ablation_supp_view} and \ref{tbl:ablation_supp}. The metrics are computed on the Multiface dataset \cite{wuu2022multiface}, as described in the main paper.

We ablate fine-tuning results when a different number of views are available in Tbl. \ref{tbl:ablation_supp_view}. We find that our method produces pleasing front-view faces even for a single input view. However, the quality quickly degrades for side views, where the input views do not provide any signal. 
Table \ref{tbl:ablation_supp} ablates a prior model trained without background removal. Keeping the background (i) yields more floaters during model fitting.

For visuals and additional results for single image inputs, please see the main paper and the supplementary HTML page.

\subsection{Ethics}
The content of this paper follows the SIGGRAPH Asia policies for data privacy. In particular, all in-the-wild subjects have signed an agreement to be captured and reconstructed for research purposes. This approach inhibits the same risks and dangers as other face reconstruction methods.

\end{document}